\documentclass[11pt]{article}
\usepackage[utf8]{inputenc}
\usepackage[margin=1in]{geometry}

\usepackage{amsmath, amssymb, parskip,mathtools,amsthm}
\usepackage{graphicx,subcaption,multirow,booktabs}
\usepackage{enumerate}
\usepackage{enumitem}
\usepackage{outlines}

\usepackage{natbib}
\usepackage{url} %
\usepackage{shortcut}
\usepackage{hyperref}
\usepackage[capitalise]{cleveref}
\Crefname{assumption}{Assumption}{Assumptions}
\Crefname{enumerate}{condition}{conditions}
\usepackage{colortbl}
\usepackage{tikz}
\usepackage{diagbox}
\usetikzlibrary{positioning}

\def\spacingset#1{\renewcommand{\baselinestretch}%
{#1}\small\normalsize} \spacingset{1}

\usepackage{natbib}
 \bibpunct[, ]{(}{)}{,}{a}{}{,}%

\newcommand{\expect}{\mathbb{E}}

\newcommand{\ind}{\mathbb{I}}

\newcommand{\tnua}{\eta_1^*}
\newcommand{\tnub}{\eta_2^*}
\newcommand{\nua}{\eta_1}
\newcommand{\nub}{\eta_2}

\newcommand{\tth}{\theta^*}
\newcommand{\hth}{\hat{\theta}}
\newcommand{\hthinit}{\hat{\theta}_{1, \text{init}}}

\newcommand{\hthall}{\hat{\theta}_{\text{init}}}

\newcommand{\hnua}{\hat{\eta}_1}
\newcommand{\hnub}{\hat{\eta}_2}

\newcommand{\tpib}{\pi^*}
\newcommand{\hpib}{\hat{\pi}}
\newcommand{\tq}{\mu^*}
\newcommand{\hq}{\hat{\mu}}

\newcommand{\calI}{\mathcal{I}}
\newcommand{\calF}{\mathcal{F}}
\newcommand{\hG}{\mathbb{G}}

\newcommand{\diag}{\operatorname{diag}}
\newcommand{\calT}{\mathcal{T}}
\newcommand{\pk}{^{(k)}}

\newcommand{\mb}[1]{\mathbf{#1}}
\newcommand{\Dcal}{\mathcal{D}}

\newcommand{\hP}{\mathbb{P}}
\newcommand{\Pcal}{\mathcal{P}}
\newcommand{\Fcal}{\mathcal{F}}

\newcount\Comments  %
\newcommand{\kibitz}[2]{\ifnum\Comments=1\textcolor{#1}{#2}\fi}

\newtheorem{theorem}{Theorem}

\newtheorem{proposition}{Proposition}
\newtheorem{assumption}{Assumption}
\newtheorem{definition}{Definition}
\theoremstyle{definition}

\newtheorem{remark}{Remark}

\newcommand*\samethanks[1][\value{footnote}]{\footnotemark[#1]}

\title{Localized Debiased Machine Learning: Efficient Inference on Quantile Treatment Effects and Beyond}
\author{Nathan Kallus$^1$\thanks{Alphabetical order.}, 
~Xiaojie Mao$^2$\samethanks,
~Masatoshi Uehara$^1$\samethanks}
\date{
$^1$Cornell University; ~~ 
$^2$Tsinghua University.
}

\begin{document}
\maketitle

\begin{abstract}
We consider estimating a low-dimensional parameter in an estimating equation involving high-dimensional nuisances that depend on the parameter. A central example is the efficient estimating equation for the (local) quantile treatment effect ((L)QTE) in causal inference, which involves as a nuisance the covariate-conditional cumulative distribution function evaluated at the quantile to be estimated. Debiased machine learning (DML) is a data-splitting approach to estimating high-dimensional nuisances using flexible machine learning methods, but applying it to problems with parameter-dependent nuisances is impractical. For (L)QTE, DML requires we learn the \emph{whole} covariate-conditional cumulative distribution function. We instead propose \emph{localized} debiased machine learning (LDML), which avoids this burdensome step and needs only estimate nuisances at a \emph{single} initial rough guess for the parameter. For (L)QTE, LDML involves learning just two regression functions, a standard task for machine learning methods. We prove that under lax rate conditions our estimator has the same favorable asymptotic behavior as the infeasible estimator that uses the unknown true nuisances. Thus, LDML notably enables practically-feasible and theoretically-grounded efficient estimation of important quantities in causal inference such as (L)QTEs when we must control for many covariates and/or flexible relationships, as we demonstrate in empirical studies.   
\end{abstract}

\section{Introduction}
\label{sec:intro}
In this paper, we consider estimating parameters 
$\tth = (\theta_1^*,\, \theta_2^*)\in\Theta = \Theta_1 \times \Theta_2 \subseteq \R{d}$ 
defined as the (unique) solution to the following $d$-dimensional estimating equation:
\begin{align}\label{eq: est-equation}
  \hP \psi(Z; \theta, \tnua(Z; \theta_1), \tnub(Z)) = \mb{0},
\end{align}
where $Z\in \mathcal{Z}$ are observed random variables with distribution $\hP$, $\tnua(Z; \theta_1)$  and $\tnub(Z)$ are two unknown nuisance functions, and $\mb{0}$ is the zero vector in $\R d$.
 We hope to estimate $\tth$ based on $\prns{Z_1, \dots, Z_N}$, $N$ independent and identically distributed (i.i.d) draws from the distribution $\hP$.
As we will show, estimating equations of the form above are prevalent in efficient estimation in causal inference and missing data problems, with quantile treatment effect (QTE) estimation (\cref{sec: ex qte}) as a prominent example, among many others (\cref{sec: estimating equation example}). 

{ One important feature of \cref{eq: est-equation} is that the nuisance $\tnua(Z; \theta_1)$ depends on the parameters to be estimated, which raises several challenges and causes existing methods to be unstable and computationally burdensome. Specifically, we could potentially use the observed data to estimate the nuisances $\tnua(Z; \theta_1)$ and $\tnub(Z)$ and then solve a sample analogue of \cref{eq: est-equation} based on the estimated nuisances in order to estimate $\tth$, possibly using cross-fitting \citep{ChernozhukovVictor2018Dmlf}. 
However, this requires estimating the nuisance $\tnua(Z; \theta_1)$ for \emph{all} possible $\theta_1$, \ie, learning \emph{infinitely} many functions of $Z$, and then solving for the root of an estimated function. 
For example, when estimating QTE (see \cref{sec: ex qte}), 
this involves estimating a \emph{whole} conditional cumulative distribution function, or equivalently, \emph{infinitely} many binary probability regressions (one for each threshold). 
This can be very unstable, especially in causal inference with observational data where typically a large number of covariates need to be conditioned on to remove confounding. 
Although one may discretize the space of $\Theta_1$ and estimate $\tnua(Z; \theta_1)$ only for finitely many $\theta_1$, this can still be computationally burdensome when the discrete grid is large, and the resulting estimator can be sensitive to the discretization scheme. }

In this paper, we propose a localized debiased machine learning (LDML) approach that only requires estimating $\eta_1^*(Z;\theta_1)$ at a \emph{single} $\theta_1$ value, without estimating it for all possible values or ad-hoc discretized values of $\theta_1$. 
Importantly, our estimator is asymptotically equivalent to an oracle estimator that knows the whole continuum of nuisance function $\eta_1^*(Z;\theta_1)$ for all $\theta_1 \in \Theta_1$.  
In other words, asymptotically, our method does not incur any loss even though it only estimates the nuisance function at  a single  $\theta_1$ value.
Moreover, estimating this far simpler nuisance reduces to standard classification and regression tasks, \ie, fitting conditional expectations (regression) and conditional binary probabilities (classification), for which 
many machine learning methods exist.
 In particular, our approach will be shown to be largely insensitive to \emph{how} these conditional expectation functions are estimated, so we may directly use off-the-shelf machine learning methods and treat them as black-box regression or classification algorithms (\eg, random forests, gradient boosting, neural networks).
  Therefore, our proposed method notably enables practical and efficient estimation using time-tested machine learning methods to solve \cref{eq: est-equation}.
  
{ 
In comparison,
existing approaches for debiased and efficient estimation with black-box nuisance estimators either focus on settings where nuisances do not depend on target parameters or treat nuisances as abstract objects so that one must estimate a continuum of nuisances when applying to \cref{eq: est-equation} thus precluding the use of standard machine-learning algorithms for regression and classification
\citep{robins2008higher,zheng2011cross,robins2013new,ChernozhukovVictor2018Dmlf,bravo2020two}.
Similarly, existing works specifically on the efficient estimation of QTEs either apply similar debiased approaches using a continuum of nuisances \citep{belloni2017program,diaz2017efficient} or use specific non-black-box nuisance estimators like polynomial sieves and local polynomial kernel regression and make explicit smoothness restrictions \citep{FirpoSergio2007ESEo,frolich2007unconditional}. 
(We provide an extensive literature review in \cref{sec:related}.)
Compared to these works, our proposal is fully generic, flexible, and machine-learning driven in that it handles many important examples that fit into \cref{eq: est-equation}, as we review in the next two sections; it permits the use of flexible black-box nuisance estimators, since we only require lax rate conditions that are in fact more lax than some previous results; and these black boxes may be standard machine-learning methods for regression and classification, since whenever $\eta_1^*(Z;\theta_1)$ for any single $\theta_1\in\Theta_1$ is a conditional expectation, such as in all of the examples we review in the next section, our method only ever has to fit very few conditional expectations.}

 \subsection{Motivating Example: Quantile Treatment Effects}\label{sec: ex qte}

A primary motivation of considering the setting of \cref{eq: est-equation} is the estimation of QTE. 
In this case, we consider a population $\overline\hP$ of units, each associated with some baseline covariates $X \in \mathcal{X}$, two potential outcomes $Y(0),Y(1)\in\Rl$ for each of two possible treatments, and a treatment indicator $T\in\{0,1\}$. Since both potential outcomes are included in this description, we refer to $\overline\hP$ as the complete-data distribution. We are interested in the $\gamma$-quantile of $Y(1)$: the $\theta^*_1$ such that $\overline\hP(Y(1)\leq \theta_1^*) = \gamma$ (assuming existence and uniqueness) for $\gamma\in(0,1)$. And, similarly, we are interested in the quantile of $Y(0)$ and in the difference of the quantiles, known as the quantile treatment effect (QTE), but these estimation questions are analogous so for brevity we focus just on $\theta^*_1$, the $\gamma$-quantile of $Y(1)$ (see also \cref{remark: effect variance}). 
Compared to the average outcome and the average treatment effect (ATE), the quantile of outcomes and the QTE provide a more robust assessment of the effects of treatment and are very important quantities in program evaluation.

We do not observe the potential outcomes but instead only the realized factual outcome corresponding to the assigned treatment, $Y=Y(T)$. Therefore, we only observe $Z=(X,T,Y)$, whose distribution $\hP$ is given by coarsening $\overline\hP$ via $Y=Y(T)$. Ignorable treatment assignment with respect to $X$ assumes that $Y(1)\independent T\mid X$ (\ie, no unobserved confounders) and overlap assumes that $\hP(T=1\mid X)>0$, and these together ensure that $\theta^*_1$ is identifiable from observations of $Z$.
Specifically, a straightforward identification is given by the so-called inverse propensity weighting (IPW) equation:
\begin{align}\label{eq: IPW}
\hP&\psi^\text{IPW}(Z;\theta^*_1,\eta_2^*(Z))=0,\\
\text{where}\quad &\psi^\text{IPW}(Z;\theta_1,\eta_2(Z))=\indic{T=1}\indic{Y\leq \theta_1}/\eta_2(Z)-\gamma, ~~ \eta_2^*(Z)=\Prb{T=1\mid X} \nonumber.
\end{align}
Here estimating the propensity score function $\eta_2^*$ amounts to learning a conditional probability function from a binary response, for which many standard machine learning methods exist. 
Once we construct an estimator $\hat\eta_2$, we can obtain the 
standard IPW estimator $\hat\theta_1^\text{IPW}$ by  solving $\frac1N\sum_{i=1}^N\psi^\text{IPW}(Z_i;\theta_1,\hat\eta_2(Z_i))=0$.
Generally, the error of the IPW estimator
can heavily depend on the particular method used to construct $\hat\eta_2$ and its convergence rate
can be slowed down by that of $\hat\eta_2$, prohibiting the use of general nonparametric machine learning methods and potentially leading to unstable estimates.

Instead, one can alternatively obtain the following estimating equation from the efficient influence function for $\theta^*$ \citep[\eg, ][]{tsi}:
\begin{align}\label{eq: quantile-only}
\hP&\psi(Z;\theta^*_1,\eta_1^*(Z;\theta^*_1),\eta_2^*(Z))=0,\\\nonumber
\text{where}\quad&\psi(Z;\theta_1,\eta_1(Z;\theta_1),\eta_2(Z))=\indic{T = 1}\prns{\indic{Y\leq \theta_1}-\eta_1(Z;\theta_1)}/\eta_2(Z)+\eta_1(Z;\theta_1)-\gamma,\\\nonumber
&\eta_1^*(Z;\theta_1)=\Prb{Y\leq \theta_1\mid X,T=1}.
\end{align}
An important feature of the above is that it satisfies a property known as \emph{Neyman orthogonality}: the moment $\hP\psi(Z;\theta_1,\eta_1(Z;\theta_1),\eta_2(Z))$ has \emph{zero} derivatives with respect to the nuisances at $\theta_1^*,\eta_1^*,\eta_2^*$.
This means that the estimating equation is robust to small perturbations in the nuisances so that estimation errors therein contribute only to higher-order error terms in the final estimate of $\tth_1$.
In particular, \citet{ChernozhukovVictor2018Dmlf} recently proposed to leverage Neyman orthogonality to enable the use of plug-in machine learning estimates of the nuisances. Their proposal, called debiased machine learning (DML), is as follows:
split the data randomly into $K$ folds, $\mathcal D_1, \dots, \mathcal D_K$, and then for each $k = 1, \dots, K$, use all but the $k^{\text{th}}$ fold to construct nuisance estimates $\hat\eta_1^{(k)},\,\hat\eta_2^{(k)}$, and finally solve the empirical estimating equation $\frac{1}{N}\sum_{k = 1}^K\sum_{i\in\mathcal D_k}\psi(Z_i;\theta_1,\hat\eta_1^{(k)}(Z_i;\theta_1),\hat\eta_2^{(k)}(Z_i))=0$ to obtain the estimator $\hat \theta$. 
They prove that as long as the estimates $\hat\eta_1^{(k)},\,\hat\eta_2^{(k)}$ converge to $\tnua,\,\tnub$ faster than $N^{-1/4}$, the estimate $\hat\theta_1$ will have similar behavior to the \emph{oracle} estimate that solves $\frac1N\sum_{i=1}^N\psi(Z_i;\theta_1,\eta_1^*(Z_i;\theta_1),\eta_2^*(Z_i))=0$, \ie, solving the empirical estimating equation using the \emph{true} nuisance functions.
As a result, the estimate $\hat \theta_1$ is asymptotically normal and semiparametrically \emph{efficient}. 
Since, apart from the mild rate requirement on $\hat\eta_1^{(k)},\,\hat\eta_2^{(k)}$, no metric entropy conditions are assumed, this allows one to successfully use machine learning methods to learn nuisances and achieve asymptotically normal and efficient estimation.

The problem with this approach for estimating quantiles of outcomes (similarly, QTEs), however, is that it requires the estimation of a very complex nuisance function: $\eta_1^*(Z;\theta_1)$ is the \emph{whole} conditional cumulative distribution function of a real-valued outcome, potentially conditioned on high-dimensional covariates.
While certainly nonparametric methods for estimating conditional distributions exist such as kernel estimators, this learning problem is \emph{much harder} to do in a flexible, blackbox, machine-learning manner, compared to just estimating a single regression function. 
This indeed stands in stark contrast to the estimation of ATEs, where applying DML requires a far simpler nuisance function given by the regression of outcome on covariates and treatment, $\Eb{Y\mid X,T}$, for which a long list of practice-proven machine learning methods can be directly and successfully applied. The key difference is that the nuisance function in ATE estimation does \emph{not} depend on the estimand and can therefore be estimated in an independent manner whereas the nuisance function in QTE estimation \emph{does} depend on the estimand. This issue makes DML, despite its theoretical benefits, untenable in practice for the important task of QTE estimation.

The primary goal of this paper can be understood as extending DML to 
effectively tackle the case where nuisances depend on the estimand by alleviating this dependence via localization.
In particular, this will enable efficient estimation of important 
quantities such as QTEs in the presence of 
high-dimensional nuisances by using and debiasing black-box 
machine learning methods for the standard regression task.

The basic idea as it applies to the estimation of the quantile of outcomes, which we will generalize and analyze thoroughly, is as follows. While perhaps inefficient, $\hat\theta_1^\text{IPW}$ relies only on estimating a binary regression $\eta_1^*$. This is amenable to machine learning approaches but may have a slow convergence rate in general.
Despite its slow rate, this rough initial guess can sufficiently localize our nuisance estimation and it may suffice to only estimate $\eta_1^*(Z;\hat\theta_1^\text{IPW})$, \ie, the nuisance evaluated at just a \emph{single} initial estimate of $\theta_1$.
 Then we use this estimated nuisance at this initial estimate of $\theta_1^*$ in place of $\eta_1^*(Z;\theta_1)$ when solving the empirical estimating equation for $\theta_1$. For estimating the quantiles, this means we only have to regress the binary response $\mathbb I[{Y\leq \hat\theta_1^\text{IPW}}]$ on $X$, treating $\hat\theta_1^\text{IPW}$ as fixed.
In particular, we propose a special three-way data splitting procedure that debiases such plug-in nuisance estimates in order to obtain an estimate for $\tth$ with near-oracle performance.

\subsection{Estimating Equations with Incomplete Data under Ignorable Treatment Assignment or Using Instrumental Variables}\label{sec: estimating equation example}
More generally, we can consider parameters $(\theta_1^*,\theta_2^*)$ defined as the solution to the following estimating equation on the (unavailable) complete data:
\begin{equation}\label{eq: est-eq-causal-complete-case}
\overline\hP[U(Y(1);\theta_1)+V(\theta_2)]=0,
\end{equation}
for some given functions $U(y;\theta_1)$ and $V(\theta_2)$. 
Quantile is one example of this. Another example is conditional value at risk (CVaR) of outcomes: $\theta_2^*=\overline\hP[Y(1) \indic{F_1(Y(1))\geq \gamma}]/(1-\gamma)$, where $F_1$ is the cumulative distribution function of $Y(1)$, that is, the expectation of $Y(1)$ conditioned on being above the $\gamma$-quantile (again, assuming uniqueness). 
CVaR is also known as expected shortfall, a popular risk measure widely used in risk management and optimization 
\citep{rockafellar2002conditional}.
Again, we may consider the CVaR of $Y(0)$ and the differences of CVaRs analogously. Letting
\begin{align}\label{eq: est-eq-cvar-original}
U(y;\theta_1)=\prns{\indic{y\leq\theta_1},~\max\{\theta_1,\,\prns{1-\gamma}^{-1}(y-\gamma\theta_1)\}},
~~
V(\theta_2)=\prns{-\gamma,~-\theta_2},
\end{align}
\cref{eq: est-eq-causal-complete-case} defines $(\theta_1^*,\theta_2^*)$ as the quantile and CVaR of $Y(1)$.

Yet another example is the expectile, a measure for asymmetric risk  
\citep{newey1987asymmetric}. 
The $\gamma$-expectile of $Y(1)$ is  defined by the following  asymmetric least squares problem:
\[
\theta_1^* = \underset{\theta_1 \in \mathbb{R}}{\operatorname{argmin}}~\overline\hP\left[\left|\gamma - \ind\left(Y(1) - \theta_1 \le 0\right)\right|(Y(1) - \theta_1)^2\right],
\]
whose first-order condition gives an estimating equation for \emph{complete} data:
\begin{align}\label{eq: expectile-complete}
\overline\hP\left[U(Y(1); \theta_1)\right] = \hP\left[\left(1 - \gamma\right)\left(Y(1)  - \theta_1\right) - (1 - 2\gamma)\max\left(Y(1) - \theta_1 , 0\right)\right] = 0.
\end{align}
Under ignorable treatment assignment and overlap,
a general-purpose Neyman-orthogonal estimating equation 
for the estimand $(\theta_1^*,\theta_2^*)$ defined by \cref{eq: est-eq-causal-complete-case} is given by
\begin{align}
&\psi(Z;\theta,\eta^*_1(Z;\theta_1),\eta^*_2(Z))=\frac{\indic{T=1}}{\eta_2^*(Z)}
\bigg(U(Y; \theta_1) - \eta^*_1(Z;\theta_1)\bigg) + \eta^*_1(Z;\theta_1) + V(\theta_2), \label{eq: est-eq-causal}\\
&\text{where}\quad\eta^*_1(Z;\theta_1)=\Eb{U(Y;\theta_1)\mid X,T=1}, \quad
\eta_2^*(Z)=\Prb{T=1\mid X}. \nonumber
\end{align}
Alternatively, instead of assuming ignorable treatment assignment, we may have access to an instrumental variable (IV). We considier a binary IV  denoted as $W \in \braces{0, 1}$ and assume that it satisfies identification conditions in \citet{imbens1994identification} (namely, for potential treatments $T(w)$ and potential outcomes $Y(t,w)$, we have exclusion $Y(t):=Y(t,w)=Y(t,1-w)$, exogeneity $(Y(t), T(w)) \perp W \mid X$, overlap $\hP(W=1\mid X)\in(0,1)$, relevance $\overline\hP(T(1)=1)>\overline\hP(T(0)=1)$, and monotonicity $T(1)\geq T(0)$). We seek to use observations of $Z=(X,W,T,Y)$ to estimate \emph{local} parameters defined by the following estimating equation conditionally on the subpopulation of compliers (\ie, $T(1)> T(0)$): 
\begin{equation}\label{eq: est-eq-causal-complete-case-local}
\overline\hP[U(Y(1);\theta_1)+V(\theta_2)\mid T(1)> T(0)]=0.
\end{equation}
For example, specializing \cref{eq: est-eq-causal-complete-case-local} to the functions $U\prns{y; \theta_1}, V\prns{\theta_2}$ in  
\cref{eq: est-eq-cvar-original} gives the \emph{local} quantile and CVaR, which in turn gives the \emph{local} QTE (LQTE).
In \cref{sec: IV}, we present the efficient estimating equations for these local parameters and show they also satisfy Neyman orthogonality and involve some estimand-dependent nuisance functions $\eta^*_1(Z;\theta_1)$.

In all examples above, the nuisance $\eta^*_1(Z;\theta_1)$ depends on the estimand. 
This occurs whether estimating quantiles, CVaR, or expectiles (more generally, whenever $U(y;\theta_1)$ is not linear in $\theta_1$) and whether the identification is via ignorable treatment assignment, ignorable coarsening, or valid IV.
And, in such cases, learning $\eta^*_1(Z;\theta_1)$ for \emph{all} $\theta_1$ is practically difficult, which may involve learning a whole conditional distribution function or a whole continuum of conditional expectation functions given potentially high-dimensional covariates. 

\paragraph{Notation.} We let $d_1,d_2$ be the dimensions of $\theta_1^*,\theta_2^*$, respectively, where $d_1+d_2=d$.
For $f:\R{d}\to\R{m}$, $\partial_{\theta^\top}f(\theta)$ is the $m\times d$-matrix-valued function with entry $\frac{\partial f_i(\theta)}{\partial\theta_j}$ in position $(i,j)$ and $\partial_{\theta^\top}f(\theta)|_{\theta=\theta_0}$ is its evaluation at $\theta_0$.
For $g: \R{d} \to \mathbb{R}$, $\partial_{\theta}\partial_{\theta^\top} g\prns{\theta}$ is the $d\times d$-matrix-valued function with entry $\frac{\partial g(\theta)}{\partial\theta_i\partial\theta_j}$ in position $(i,j)$. 
We use $\sigma_{\max}\prns{\partial_{\theta}\partial_{\theta^\top} g\prns{\theta}}$ to denote its largest singular value. 
 We let $\Prb{Z\in A}$ and $\Eb{Z\mid Z\in A}$ for measurable sets $A$ denote probabilities and expectations with respect to $\hP$.
We let $\hP f(Z)=\int f \diff \hP$ for measurable functions $f$ denote expectations with respect to $Z$ alone, while we let $\E f(Z;Z_1,\dots,Z_n)$ denote expectations with respect to $Z$ \emph{and} the data. Thus, if $\hat\varphi$ depends on the data, $\hP f(Z;\hat\varphi)$ remains a function of the data  while $\E f(Z;\hat\varphi)$ is a number.
We let $\hP_{N}$ denote the empirical expectation: $\hP_{N}f(Z) = \frac{1}{N}\sum_{i= 1}^N f(Z_i)$ for any measurable function $f$. 
Moreover, for vector-valued function $f(Z) = (f_1(Z), \dots, f_d(Z))$, we let $\hP f^2(Z)  \coloneqq  (\hP f^2_1(Z), \dots, \hP f^2_d(Z))$. 
For any $x \in \mathbb{R}^d$, we denote the open ball centered at $x$ with radius $\delta$ as $\mathcal{B}(x; \delta)$.
For $p>0$ and a probability measure $\mathbb Q$, we denote $\|f\|_{\mathbb Q, p} = \prns{\int |f|^p \diff \mathbb Q}^{1/p}$. 
For a set of functions $\calF$, we define the covering number $N(\epsilon,\calF,\|\cdot\|_{\mathbb Q, 2})$ as the minimal number $N$ of functions $f_1,\dots,f_N$ such that $\sup_{f\in\calF}\inf_{i=1,\dots,N}\|f-f_i\|_{\mathbb Q, 2}\leq\epsilon$.
For positive deterministic sequence $a_n$ and random variable sequence $X_n$, $X_n = o_{\hP}(a_n)$ means $\hP(|X_n|/a_n > \epsilon) \to 0\,\forall\epsilon > 0$ and $X_n = O_{\hP}(a_n)$ means
for any $\epsilon > 0$, there exists $M > 0$ such that $\limsup_{n\to\infty}\hP(|X_n|/a_n \ge M) \le \epsilon$.

\section{Method}\label{sec: method}
We next present our methodology, first motivating the localization technique, and then explicitly stating our meta-algorithm.
\subsection{Motivation}
Ideally, if the nuisances $\tnua$ and $\tnub$ were both known, then \cref{eq: est-equation} suggests that $\tth$ could be estimated by solving the following estimating equation:
\begin{align}\label{eq: full-equation}
    \hP_{N} \left[\psi(Z; \theta, \tnua(Z; \theta_1), \tnub(Z))\right] = 0.
\end{align}
Under standard regularity conditions for $Z$-estimation \citep{vaart_1998}, the resulting oracle estimator $\tilde{\theta}$ that solves \cref{eq: full-equation} is asymptotically linear (and hence $\sqrt{N}$-consistent and asymptotically normal):
\begin{align}\label{eq: infeasible-est}
     &\sqrt{N}(\tilde{\theta} - \tth) = \frac{1}{\sqrt{N}}\sum_{i = 1}^N {J^*}^{-1}\psi(Z_i; \tth, \tnua(Z_i; \theta_1^*), \tnub(Z_i)) + o_{\hP}(1),
    \\&\notag\text{where}\quad J^* = \partial_{\theta^\top}\left\{\hP\left[\psi(Z; {\theta}, \tnua(Z; \theta_1), \tnub(Z))\right]\right\}\vert_{\theta = \tth}.
\end{align}
Furthermore, if ${J^*}^{-1}\psi(Z; \theta^*, \tnua(Z; \theta_1^*), \tnub(Z))$ is the semiparametrically efficient influence function for $\theta^*$, then $\tilde{\theta}$ also achieves the efficiency lower bound, that is, has minimal asymptotic variance among all regular estimators \citep{vaart_1998}.

Since $\tnua$ and $\tnub$ are actually unknown, the oracle estimator $\tilde\theta$ is of course infeasible. Instead, we must estimate the nuisance functions. A direct application of DML would require us to learn the whole functions $\tnua$ and $\tnub$. That is, in order to solve \cref{eq: full-equation} we would need to estimate infinitely many nuisance functions, $H_1=\{\tnua(\cdot,\theta_1):\theta_1\in\Theta_1\}$.

To avoid the daunting task of estimating infinitely many nuisances, we will instead attempt to target the following alternative oracle estimating equation
\begin{align}\label{eq: part-equation}
\hP_{N} \left[\psi(Z; \theta, \tnua(Z; \theta_1^*), \tnub(Z))\right] = 0.
\end{align}
Although \cref{eq: part-equation} appears very similar to \cref{eq: full-equation}, it only involves $\tnua(Z; \theta_1)$ at the \emph{single} value $\theta_1=\tth_1$, as opposed to the infinitely many possible values for $\theta_1$. In other words, among the whole family of nuisances $H_1$, \textit{only} $\tnua(Z; \tth_1)\in H_1$ is relevant for \cref{eq: part-equation}. This formulation considerably reduces the need of nuisance estimation: now we only need to estimate $\tnua(Z; \tth_1)$ and $\tnub(Z)$, both functions \emph{only} of $Z$.

The (infeasible) estimators that solve each of \cref{eq: full-equation,eq: part-equation} have the same leading asymptotic behavior as long as the respective associated Jacobian matrices coincide, as posited by the following assumption.
\begin{assumption}[Invariant Jacobian]\label{assump: jacob}
        $\partial_{\theta^\top}\{\hP\left[\psi(Z; {\theta}, \tnua(Z; \theta_1^*), \tnub(Z))\right]\}\vert_{\theta = \tth} = J^*$.
\end{assumption}
In \cref{sec: jacobian}, we provide a general sufficient condition for \cref{assump: jacob} in terms of a Fr\'echet-differentiation variant of the Neyman orthogonality condition (\cref{assump: identification} condition \ref{assump: identification: orthogoanlity}). 
But it is easy to directly show that 
this invariant Jacobian assumption holds for estimating equations with incomplete data presented in \cref{sec: estimating equation example}. 
In particular, the estimating equation $\psi$ in \cref{eq: est-eq-causal} satisfies that 
\begin{align}\label{eq: invariant-jacobian-example}
\hP \left[\psi(Z;\theta,\eta_1^*(Z;\theta_1),\eta^*_2(Z))\right]    &= \hP\bracks{\frac{\indic{T=1}}{\eta_2^*(Z)}U(Y; \theta_1) - \frac{\indic{T=1} - \eta_2^*(Z)}{\eta_2^*(Z)}\tnua(Z; \theta_1)} +  V(\theta_2) \nonumber \\
    &= \hP\bracks{U\prns{Y(1); \theta_1}} + V(\theta_2),
\end{align}
which does not depend on $\tnua(Z; \theta_1)$ at all. Thus whether fixing $\tnua(Z; \theta_1)$ at $\theta_1 = \theta_1^*$ or not, the Jacobian matrix of the estimating equation $\psi$ remains the same.
This means that solving \cref{eq: full-equation} or \cref{eq: part-equation} will have the same asymptotic behavior. Both, however, are infeasible since they involve unknown nuisances. Nonetheless, \cref{eq: part-equation} motivates a new algorithm that eschews estimating $H_1=\{\tnua(\cdot\ ;\theta_1):\theta_1\in\Theta_1\}$ in full.

\subsection{The LDML Meta-Algorithm}\label{sec: LDML-algorithm}
\begin{figure}[t!]\centering\large%
\includegraphics[width=\textwidth]{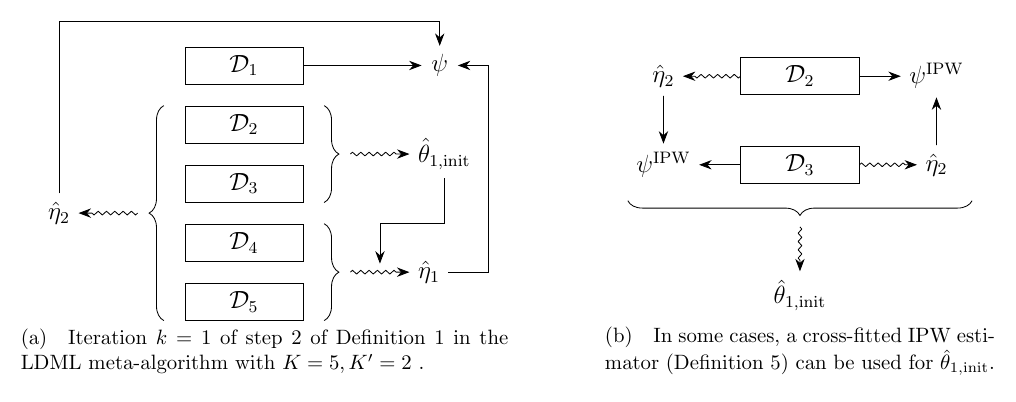}
\caption{Sketch of the LDML estimation procedure and a possible initial guess estimator. Squiggly arrows ``$\rightsquigarrow$'' denote estimation. Plain arrows ``$\rightarrow$'' denote plugging in.}\label{fig: schema}%
\bigskip\end{figure}

Motivated by the new (infeasible) estimating equation in \cref{eq: part-equation}, we propose to estimate $\tth$ by the following (feasible) three-way sample splitting method, which we term localized debiased machine learning (LDML).
The algorithm has two parts: three-way-cross-fold nuisance estimation and solving the estimating equation.

We start by discussing how we estimate the nuisances that we will then plug into \cref{eq: part-equation}.
\begin{definition}[3-way-cross-fold nuisance estimation]
\label{def: splitting}
Fix integers $K\geq3,\,K'\in[1, K-2]$.
\begin{enumerate}
\item Randomly permute the data indices and let $\Dcal_k=\{\lceil(k-1)N/K\rceil+1,\dots,\lceil kN/K\rceil\},\,k=1,\dots,K$ be a random even $K$-fold split of the data.
\item \label{def: splitting per fold step} For $k = 1, \dots, K$:
\begin{enumerate}
    \item Set $\mathcal H_{k, 1}=\{1,\dots,K'+\indic{k\leq K'}\}\setminus \{k\},\,\mathcal H_{k, 2}=\{K'+\indic{k\leq K'}+1,\dots, K\}\setminus \{k\}$.
    \item Use only $\Dcal_k^{C,1}=\left\{Z_i:i\in\bigcup_{k'\in\mathcal H_{k,1}}\Dcal_{k'}\right\}$ to construct an initial estimator $\hthinit\pk$ of $\theta_1^*$. 
    \label{step: nuisance est} 
    Use only $\Dcal_k^{C,2}=\left\{Z_i:i\in\bigcup_{k'\in\mathcal H_{k,2}}\Dcal_{k'}\right\}$ to construct estimator $\hnua\pk(\cdot\ ; \hthinit\pk)$ of $\tnua(\cdot\ ; \hthinit\pk)$. Use only $\Dcal_{k}^{C, 1} \cup \Dcal_{k}^{C, 2}$ to construct estimator  $\hnub\pk$ of $\tnub$.
\end{enumerate}
\end{enumerate}
\end{definition}

For illustration the first iteration of step \ref{def: splitting per fold step} above is sketched in \cref{fig: schema}(a) along with the plugging of estimated nuisances into the estimating equation (see \cref{def: LDML1,def: LDML2} below).

Notice that since $\Dcal_k^{C,1}$ and $\Dcal_k^{C,2}$ are disjoint, 
$\tnua(\cdot\ ; \hthinit\pk)$ is a fixed, nonrandom function with respect to the data $\Dcal_k^{C,2}$. That is, the $\tnua$ nuisance estimation task in step \ref{step: nuisance est} appears as estimating a \emph{single} $\tnua(\cdot\ ;\theta_1')\in H_1$ for $\theta_1' = \hthinit\pk$
rather than the estimation of \emph{all} of $H_1$. 

A natural question is, what might be a reasonable initial estimator.
In the examples given in \cref{sec: ex qte,sec: estimating equation example}, we can use an IPW estimate for $\hthinit\pk$ (see \cref{fig: schema}(b) and \cref{def: IPW_init}).

Given these nuisance estimates, we can obtain the LDML estimator for $\theta^*$ by approximately 
solving the average of the estimate of \cref{eq: part-equation} in each fold.
\begin{definition}[LDML]
\label{def: LDML2}
We let the estimator $\hat\theta$ be given by (approximately) solving
\begin{align}
    \overline\Psi(\theta)=\frac{1}{N}\sum_{k = 1}^K\sum_{i \in \Dcal_{k}}\psi(Z_i; \theta, \hnua\pk(Z_i; \hthinit\pk), \hnub\pk(Z_i)) = 0.
\end{align}
In fact, we allow for an approximate least-squares solution, which is useful if the empirical estimating equation has no exact solution. Namely, we
let $\hat\theta$ be any satisfying 
\begin{align}\label{eq: approx-LDML2}
    &\textstyle\|
    \overline\Psi(\hth)
    \| 
    \le \inf_{\theta \in \Theta}\|
    \overline\Psi(\theta)
    \| + \varepsilon_N. 
\end{align}
\end{definition}
In \cref{sec: alternative-est-eq} \cref{def: LDML1}, we give an alternative LDML estimator obtained by 
 averaging solutions to \cref{eq: part-equation} estimated in each fold separately.
 These two LDML estimators are asymptotically equivalent and all results in this paper apply to both, thus we focus on \cref{def: LDML2} in the main text. 
 Moreover, 
both estimators depend on the random splitting in \cref{def: splitting}.
To reduce the variance from this, we may aggregate estimates from multiple different sample splitting realizations. See \cref{sec: practical} for a detailed discussion.

\section{Theoretical Analysis}\label{sec: general}
In this section, we provide the sufficient conditions that guarantee the proposed estimator $\hth$ in \cref{def: LDML1,def: LDML2} to be consistent and asymptotically normal. In particular, although the proposed estimator relies on plug-in nuisance estimators, it is asymptotically equivalent to the \textit{infeasible} estimator based on \cref{eq: full-equation} with \emph{true} nuisances, that is, it satisfies \cref{eq: infeasible-est}. While some of our conditions are analogous to those in \cite{ChernozhukovVictor2018Dmlf}, some are not and our proof takes a different approach that enables weaker conditions for convergence rates of the nuisance estimators. 

Our asymptotic normality results may be stated uniformly over a sequence of models $\mathcal P_N$ for any data generating distribution $\hP\in\Pcal_N$.
Our first set of assumptions ensure that $\tth$ is reasonably identified by the given estimating equation for all $\hP\in\Pcal_N$.
We also assume that our estimating equation satisfies the Neyman orthogonality condition with respect to a nuisance realization set $\mathcal{T}_N\subset[\mathcal Z\to\Rl]^2$ that contains the nuisance estimates $\hnua(\cdot\ ; \hthinit)$ and $\hnub(\cdot)$ with high probability. Note the set $\mathcal{T}_N$ consists of pairs of functions of the data $Z$ alone and \emph{not} $\theta_1$. Therefore, we denote members of the set as $(\eta_1(\cdot\ ; \theta_1'),\eta_2(\cdot))\in\mathcal T_N$, where $\eta_1(\cdot\ ; \theta_1')$ is simply understood as a symbol representing of some fixed function of $Z$ alone.
\begin{assumption}[Regularity of Estimating Equations]\label{assump: identification}
Assume there exist positive constants $c_1$ to $c_7$ such that the following conditions hold for all $\hP \in \mathcal P_N$:
\begin{enumerate}[label=\roman*.]
    \item\label{assump: identification: interior} 
    $\Theta$ is a compact set and it contains a ball of radius $c_1 N^{-1/2}\log N$ centered at 
    $\tth$.
    \item\label{assump: identification: diffable} The map $(\theta, a, b) \mapsto \hP\left[\psi(Z; {\theta}, a,b)\right]$ is twice continuously G\^ateaux-differentiable.  
    \item\label{assump: identification: identification} 
    For any $\theta\in\Theta$, $2\|\hP\left[\psi(Z; {\theta}, \tnua(Z; \theta_1^*), \tnub(Z))\right]\| \ge \|J^*(\theta - \tth)\| \wedge c_2$. 
     \item\label{assump: identification: conditioning} $J^*$ is non-singular with singular values bounded between positive constants $c_3$ and $c_4$.
    \item \label{assump: identification: moment} Singular values of the covariance matrix $\Sigma$ are bounded betwen constants $c_5$ and $c_6$: 
        \begin{equation}\label{eq:covariancematrix}
    \Sigma \coloneqq \expect\left[{J^*}^{-1}\psi(Z; \tth, \tnua(Z; \theta_1^*), \tnub(Z))\psi(Z; \tth, \tnua(Z; \theta_1^*), \tnub(Z))^\top{J^*}^{-\top}\right].\end{equation}
 \item\label{assump: identification: metric entropy} 
The nuisance realization set $\mathcal T_N$ contains the true nuisance parameters $\left(\tnua(\cdot\ ; \tth_1), \tnub(\cdot)\right)$. Moreover, the parameter space $\Theta$ is bounded and for each $(\eta_1(\cdot\ ; \theta_1'),\eta_2(\cdot))\in\mathcal T_N$, the function class $\calF_{\eta, \theta_1'} = \{Z\mapsto\psi_j(Z; \theta, \nua(Z; \theta_1'), \nub(Z)): j = 1, \dots, d, \theta \in \Theta\}$ is suitably measurable and its uniform covering entropy satisfies the following condition: for positive constants $a$, $v$, and $q > 2$,
    $
        \sup_{\mathbb Q}\log N(\epsilon\|F_{\eta, \theta_1'}\|_{\mathbb Q, 2}, \calF_{\eta, \theta_1'}, \|\cdot\|_{\mathbb Q, 2}) \le v\log(a\epsilon)$ $\forall \epsilon\in(0,1]
   $,
    where $F_{\eta, \theta_1'}$ is a measurable envelope for $\calF_{\eta, \theta_1'}$ that satisfies $\|F_{\eta, \theta_1'}\|_{\hP, q} \le c_7$.
    \item\label{assump: identification: orthogoanlity} 
   $\partial_{r}\braces{\hP \psi(Z;\theta^*,\eta_1^*(Z;\theta_1^*)+r (\eta_1(Z;\theta'_1)-\eta_1^*(Z;\theta^*_1)),\eta_2^*(Z)+r(\eta_2(Z)-\eta_2^*(Z))}\big|_{r=0}=0$ for all $(\eta_1(\cdot\ ; \theta_1'),\eta_2(\cdot))\in\mathcal T_N$.
\end{enumerate}
\end{assumption}
\Cref{assump: identification} conditions \ref{assump: identification: interior}--\ref{assump: identification: moment} constitute standard identification and regularity conditions for $Z$-estimation (with uniform guarantees; see also \cref{remark: uniform} below). \Cref{assump: identification} condition \ref{assump: identification: metric entropy} requires that $\psi$ is a well-estimable function of $\theta$ for any \emph{fixed} set of nuisances. Importantly, while it imposes a metric entropy condition on $\psi$, this condition does \emph{not} impose metric entropy conditions on our nuisance estimators, so flexible machine learning nuisance estimators are allowed. This assumption is very mild as $\Theta$ is finite-dimensional, so it can be ensured by some continuity and compactness condition.
Finally, \cref{assump: identification} condition \ref{assump: identification: orthogoanlity} is the  Neyman orthogonality condition \citep{ChernozhukovVictor2018Dmlf}. We will show how these conditions are ensured in the incomplete data setting in \cref{sec: estimating equation example}.

Our second set of assumptions involve conditions on our nuisance estimators.
\begin{assumption}[Nuisance Estimation Conditions]\label{assump: error}
For any $\hP \in \mathcal P_N$: 
\begin{enumerate}[label=\roman*.]
    \item \label{assump: error: nuisance-set}
    For some sequence of constants $\Delta_N\to0$,
    the nuisance estimates $(\hnua\pk(\cdot\ ; \hthinit\pk), \hnub\pk(\cdot))$ belong to the realization set $\calT_N$ for all $k=1,\dots,K$ with probability at least $1 - \Delta_N$.
    \item \label{assump: error: rate-condition} For some sequence of constants $\delta_N, \tau_N \to 0$, the statistical rates $r_N$, $r'_N$, $\lambda'_N\prns{\theta}$ satisfy:
    \begin{align*}
        &\textstyle\hspace{-2.5em}r_N 
            \coloneqq \sup_{(\nua(\cdot; \theta_1'), \nub) \in \mathcal{T}_N, \theta \in \Theta} \left\|\hP\left[\psi(Z; \theta, \nua(Z; \theta_1'), \nub(Z))\right] - \hP\left[\psi(Z; \theta, \tnua(Z; \tth_1), \tnub(Z))\right]\right\| \le \delta_N\tau_N, \\
        &\textstyle\hspace{-2.5em}r_N' 
            \coloneqq \sup_{\begin{subarray}{l}\theta \in \mathcal{B}(\tth; \tau_N) , \\(\nua(\cdot; \theta_1'), \nub) \in \mathcal{T}_N\end{subarray}}\left\|(\hP \left[\psi(Z; \theta, \nua(Z; \theta_1'), \nub(Z)) - \psi(Z; \theta, \tnua(Z; \tth_1), \tnub(Z)) \right]^2)^{1/2}\right\| \le \frac{\delta_N}{\log N},\\
        &\textstyle\hspace{-2.5em}\lambda_N'(\theta)
            \coloneqq \sup_{\begin{subarray}{l}r \in (0, 1), \\(\nua(\cdot; \theta_1'), \nub) \in \mathcal{T}_N\end{subarray}} \|\partial_r^2 f(r; \theta, \nua(\cdot; \theta_1'), \nub)\| \le \left(\|\theta - \theta^*\| + N^{-1/2}\right)\delta_N, ~~ \forall\theta \in \mathcal{B}(\tth; \tau_N),\\
        &\textstyle\hspace{-2.5em} \text{where } f(r; \theta, \nua(\cdot\ ;\theta_1'), \nub) \coloneqq \hP[\psi(Z;  \tth + r(\theta - \tth), \eta_{1}\prns{Z; \theta_1', r},  \eta_{2}\prns{Z; r})], \\
        &\textstyle\hspace{-2.5em} \phantom{\text{where }} \eta_{1}\prns{Z; \theta_1', r} \coloneqq \tnua(Z; \theta_1^*) + r(\nua(Z; \theta_1') - \tnua(Z; \theta_1^*)), ~ \eta_{2}\prns{Z; r} \coloneqq \tnub(Z) + r(\nub(Z) - \tnub(Z)). 
    \end{align*}
    \item \label{assump: error: approximation} The 
    solution approximation 
    error in \eqref{eq: approx-LDML2} satisfies $\varepsilon_N \le \delta_{N}N^{-1/2}$.
\end{enumerate}
\end{assumption}
Here our condition on $\lambda_N'\prns{\theta}$ differs from the counterpart condition in \cite{ChernozhukovVictor2018Dmlf}, which also leads to a different proof strategy. Our condition and proof generally requires weaker conditions for convergence rates of nuisance estimators. See the discussions in \cref{sec: compare-discussion} for more details.
Moreover, the constants $\Delta_N, \delta_N, \tau_N$ are all prespecified and do not depend on any particular instance $\hP$.

Our key result in this paper is the
following theorem, which shows that the asymptotic distribution of our estimator is identical to the (infeasible) oracle estimator solving the estimating equation in \cref{eq: full-equation} with known nuisances. 
\begin{theorem}[Asymptotic Behavior of LDML]\label{thm: general-split}
Assume \cref{assump: jacob,assump: identification,assump: error} hold with 
\begin{align}\label{eq: asymp-mild}
\begin{array}{l}\max\{\log^2 N(1 + N^{-1/2 + 1/q}),\, \delta_N \log N\}/\sqrt{N}\le \tau_N \le \delta_N,\\
\max\{r'_N {\log^{1/2}(1/r'_N)},\, N^{-1/2 + 1/q}\log(1/r’_N)\} \le \delta_N.
\end{array}
\end{align}
Then the estimator $\hat \theta$  given in \cref{def: LDML2} is asymptotically linear and converges to a Gaussian distribution uniformly over $\hP \in \mathcal P_N$:  
\begin{align}
&\sqrt{N}\Sigma^{-1/2}(\hth - \tth)
= 
\frac{1}{\sqrt{N}}\sum_{i = 1}^N \Sigma^{-1/2}{J^*}^{-1}\psi(Z_i; \tth, \tnua(Z_i; \theta_1^*), \tnub(Z_i)) + O_{\hP}(\rho_N) \leadsto \mathcal N(0, I_d),\notag
\end{align}
where $\Sigma$ is given in \cref{eq:covariancematrix},
the remainder term satisfies $\rho_N = (N^{-1/2 + 1/q} + r'_N){\log N} +r'_N {\log^{1/2}(1/r'_N)} + N^{-1/2 + 1/q}\log(1/r’_N) + \delta_N  \lesssim \delta_N$,
and the $O_\hP$ term depends only on constants pre-specified in \cref{assump: jacob,assump: identification,assump: error}
 and no instance-specific constants. 
\end{theorem}
The conditions in \cref{eq: asymp-mild} and $\rho_N \lesssim \delta_N$ are fairly mild because \cref{assump: identification} condition \ref{assump: identification: metric entropy} requires $q > 2$ (so $N^{-1/2 + 1/q} \to 0$) and \cref{assump: error} condition \ref{assump: error: rate-condition} requires $r_N' \le \frac{\delta_N}{\log N}$.
\begin{remark}[Uniform vs non-uniform convergence]\label{remark: uniform}
To obtain a non-uniform convergence result, we need only need set $\Pcal_N=\{\hP\}$ as a constant singleton in \cref{thm: general-split}. In this case, much of \cref{assump: identification} simplifies: the existence of the constants $c_4,c_6$ is trivial, the non-singularity of $J^*$ is enough for $c_3$ to exist, and $\theta^*$ being in the interior of $\Theta$ is enough for $c_1$ to exist.
Further, we can relax condition \ref{assump: identification: conditioning} by allowing $c_5$ to be zero (in which case we rephrase the asymptotic normality in \cref{thm: general-split} by putting $\Sigma$ on the right-hand side of the limit rather than inverting it). 
Uniformity, however, is important in practice.
Without uniformity, 
for any given sample size $N$ there may always exist some bad instance such that the normal approximation suggested by the convergence is inaccurate \citep{kasy2019uniformity}. 
\end{remark}

\section{Variance Estimation and Inference}\label{sec: inference}

In the previous section we established the asymptotic normality of the LDML estimator under lax conditions. This suggests that if we can estimate its asymptotic variance, then 
we can easily construct confidence intervals on $\theta$. In this section we provide a variance estimator and prove its consistency, resulting in asymptotically calibrated confidence intervals. For DML, \citet{ChernozhukovVictor2018Dmlf} provides variance estimates only for estimating functions $\psi$ that are linear in $\theta$, which already excludes estimand-dependent nuisances. Our results are therefore notable both for handling nonlinear and non-differentiable estimating equations and for handling estimand-dependent nuisances.
\begin{definition}[LDML variance estimator]\label{def: var est}
Given $\hat\theta$ from  \cref{def: LDML2} and $\hat J$, set
$$
\widehat\Sigma=
\frac{1}{N}\sum_{k = 1}^K\sum_{i \in \Dcal_{k}}
\hat J^{-1}
\psi(Z_i; \hat\theta, \hnua\pk(Z_i; \hthinit\pk), \hnub\pk(Z_i))\psi(Z_i; \hat\theta, \hnua\pk(Z_i; \hthinit\pk), \hnub\pk(Z_i))^\top
\hat J^{-\top}.
$$
\end{definition}
We next establish the consistency of  $\widehat\Sigma$, which relies on the following assumption.
\begin{assumption}\label{assump:variance}
Assume that $\|\hat J-J^{*}\|=\rho_{J,N}\lesssim \delta_N$ and that for some $C,\beta>0$,
\begin{align}
    &m_N\textstyle:=\sup_{(\nua(\cdot; \theta_1'), \nub) \in \mathcal{T}_N}\hP[\|\psi(Z;\theta^{*},\eta_1(Z; \theta_1'),\eta_2(Z))\|^{4}]^{1/4}\leq C\quad\forall \theta \in \mathcal{B}(\theta^{*};\tau_N),\notag\\
\label{eq:diff_theta}
&\hP[\|\psi(Z; \theta, \nua^*(Z_i; \theta^*_1), \nub^*(Z_i))-\psi(Z; \theta^*, \nua^*(Z_i; \theta^*_1), \nub^*(Z_i)) \|^2 ] \leq C \|\theta-\theta^{*}\|^{\beta}. 
\end{align}
\end{assumption}
Here, \cref{eq:diff_theta} implies continuity $\theta \mapsto \psi(Z;\theta,\nua^*(Z; \theta^*_1), \nub^*(Z))$ in terms of $L_2$ norm in the 
range space. Note that this condition can be satisfied even if $\theta \mapsto \psi(Z;\theta,\nua^*(Z; \theta^*_1), \nub^*(Z))$ is non-differentiable. For example, in the estimation of QTEs, the efficient estimating equation in \cref{eq: quantile-only} involves the indicator function $\indic{Y \le \theta_1}$, so the map $\theta \mapsto \psi(Z;\theta,\nua^*(Z; \theta^*_1), \nub^*(Z))$ is obviously not differentiable. However, the condition in \cref{eq:diff_theta} amounts to
\begin{align*}
    \hP[\prns{\hP\prns{T=1\mid X}}^{-1}\prns{\mathbb{I}[Y\leq \theta_1]-\mathbb{I}[Y\leq \theta^{*}_1]}^2 ] \leq C|\theta_1-\theta^*_1|^{\beta}.
\end{align*} 
In \cref{assump: unconf}, we will assume that $\hP\prns{T=1\mid X} \ge \epsilon_{\pi}$ for a positive constant $\epsilon_{\pi}$. Then the condition above follows if the cumulative distribution function of $Y\prns{1}$ is smooth enough, so that  
$\abs{\overline\hP(Y(1) \le \theta_1) - \overline\hP(Y(1) \le \theta_1^*)} \le C\epsilon_{\pi}\abs{\theta_1-\theta^*_1}^{\beta}$ for any $\theta_1 \in \mathcal{B}(\theta^{*}_1;\tau_N)$.

Under \cref{assump:variance}, we now show that the variance estimator in \cref{def: var est} is consistent and it leads to asymptotically valid confidence intervals.
\begin{theorem}\label{thm:cf}
Assume the assumptions in \cref{thm: general-split} and \cref{assump:variance}. Then, 
\begin{align*}
      &\hat \Sigma=\Sigma+O_{\hP}(\rho''_N) \to \Sigma, ~~ \text{uniformly over } \hP\in \mathcal{P}_N, \\
      &\text{where } \rho''_N= N^{-1/2+1/q}(\log N)^{1/2}  + N^{-1/4}(\log N)^{1/2}+r'_N+\rho_{J,N}+N^{-\beta/4} \lesssim \delta_N. 
\end{align*}
Given some $\zeta\in\R d$, the confidence interval $\mathrm{CI} \coloneqq \fbracks{\zeta^{\top}\hat \theta  \pm \Phi^{-1}(1-\alpha/2)\sqrt{\zeta^{\top}\hat \Sigma^2\zeta/N}}$ obeys
\begin{align*}
 \textstyle\sup_{\hP\in \mathcal{P}_N}\abs{{\hP}(\zeta^{\top}\theta^{*}\in \mathrm{CI})-(1-\alpha)} \to 0, ~ \text{as } N \to \infty.
\end{align*}
\end{theorem}
In \cref{def: var est}, we assumed that we have a consistent estimator $\hat J$ for $J^*$. How to construct such an estimator depends on the problem. When $\theta \mapsto \psi(Z;\theta,\nua^*(Z; \theta^*_1), \nub^*(Z))$ is differentiable, an estimator may easily be constructed as follows:
$$\hat J=\frac1N\sum_{k=1}^K\sum_{i\in\Dcal_k}\partial_{\theta^{\top}}\psi(Z;\theta,\hnua\pk(Z; \hthinit), \hnub\pk(Z))|_{\theta=\hat \theta}.$$ 
However, the estimating equation for QTE is  not differentiable. Thus we rely on deriving the form of $J^{*}$ and estimate it directly, which we discuss in detail in \cref{rem:variancee_quantile}. 

With finite sample, the variance of the LDML estimator also depends on the uncertain sample splitting in \cref{def: splitting}. This uncertainty can be additionally accounted for when multiple sample splitting realizations are used, which we discuss in \cref{sec: practical}.

\begin{remark}[Estimating and Conducting Inference on Treatment Effects]\label{remark: effect variance}
Suppose we have two sets of parameters, $\theta^{*(0)},\,\theta^{*(1)}$, each identified by its own estimating equation, $\psi^{(0)},\,\psi^{(1)}$, and we are interested in estimating the difference, $\tau^*=\theta^{*(1)}-\theta^{*(0)}$. For example, $\theta^{*(0)},\,\theta^{*(1)}$ can be the quantile and/or CVaR of $Y(0),\,Y(1)$, respectively, and we are interested in the QTE and/or CVaR treatment effect. To do this, we can concatenate the two estimating equations and augment them with the additional equation $\theta^{*(1)}-\theta^{*(0)}-\tau^*=0$. Estimating this set of estimating equations with LDML is equivalent to applying LDML to each of $\psi^{(0)},\,\psi^{(1)}$ and letting $\hat\tau$ be the difference of the estimates $\hat\theta^{(0)},\hat\theta^{(1)}$, where we may use the same data and the same folds for the two LDML procedures. For QTE and for other estimating equations with incomplete data, we can even share the nuisance estimates of the propensity score (\ie, $\hnub^{(1),(k)} = \hnub^{(0),(k)}$ in the below equation). The variance estimate one would derive for $\hat\tau$ from the augmented estimating equations is equivalent to
\begin{align*}
\widehat\Sigma_\tau=
\frac{1}{N}\sum_{k = 1}^K\sum_{i \in \Dcal_{k}}
\omega_{i,k}\omega_{i,k}^\top,~~\text{where}~~
\omega_{i,k}&=
(\hat J^{(1)})^{-1}
\psi^{(1)}(Z_i; \hat\theta^{(1)}, \hnua^{(1),(k)}(Z_i; \hthinit^{(1),(k)}), \hnub^{(1),(k)}(Z_i))
\\&\phantom{=}-
(\hat J^{(0)})^{-1}
\psi^{(0)}(Z_i; \hat\theta^{(0)}, \hnua^{(0),(k)}(Z_i; \hthinit^{(0),(k)}), \hnub^{(0),(k)}(Z_i)).
\end{align*}
\end{remark}
\section{Estimating Equations with Incomplete Data}\label{sec: bandit}
In this section, we apply our method and theory to general estimating equations with incomplete data presented in \cref{eq: est-eq-causal-complete-case}, which subsumes the estimation of QTEs, quantile of potential outcomes, CVaR treatment effect, CVaR of potential outcomes, expectile treatment effect, and expectile of potential outcomes.
We will proceed to further specialize these results to quantile and CVaR estimation, deferring the case of expectiles to the appendix (\cref{sec: expectiles}). 
We also defer the case of using IVs to estimate the solution to \emph{local} estimating equations, such as those that describe the LQTE, to appendix (\cref{sec: IV}).

As motivated in \cref{sec: ex qte}, under unconfoundedness, there is a very natural initial estimator: the IPW estimator. As we will show, the LDML estimate for this problem using the IPW initial estimator can be computed using just blackbox algorithms for (possibly binary) regression, which is the standard supervised learning task in machine learning. And, under lax conditions, the estimate is efficient, asymptotically normal, and amenable to inference. 

Recall that $\theta$ is defined by the complete-data estimating equations in \cref{eq: est-eq-causal-complete-case}, namely, $\overline\hP[U(Y(1);\theta_1)+V(\theta_2)]=0$. Assuming ignorability and overlap, $\theta$ is identified from the incomplete-data observations $Z=(X,T,Y)$ where $Y=Y(T)$. In particular, \cref{eq: est-eq-causal} provides a Neyman-orthogonal estimating equation identifying $\theta$. 
For better interpretability, we give our nuisances names:
we denote $\tpib(t \mid x) = \hP(T = t \mid X = x)$, $\tq_j(x, t; \theta_1) = \Eb{U_j(Y; \theta_1) \mid X = x, T = t}$, and $\tq(x, t; \theta_1) = [\tq_1(x, t; \theta_1), \dots, \tq_d(x, t; \theta_1)]^\top$. For estimating parameters corresponding to $Y(1)$, our estimand-independent nuisance is the propensity score $\eta^*_2(Z)=\pi^*(1\mid X)$, and our estimand-dependent nuisance is $\eta^*_1(Z;\theta_1)=\tq(X,1;\theta_1)$. The case for $Y(0)$ is symmetric; and it also need the symmetric ignorability and overlap assumptions for identifiability: $Y(0)\independent T\mid X$ and $\hP(T=1\mid X)<1$. Treatment effects (\eg, QTEs) can be estimated by differences of estimates, where we can use the same data, the same fold splits, and the same estimates of $\tpib$ for both treatments (see \cref{remark: effect variance}).

This problem also admits a simpler but unstable (\ie, non-orthogonal) estimating equation using IPW, which suggests a possible initial estimator, using $K'\geq2$ in \cref{def: splitting}:
\begin{definition}[IPW Initial Estimator]\label{def: IPW_init}
For each $k = 1, \dots, K$ and $l \in  \mathcal{H}_{k, 1}$ as in \cref{def: splitting}, use only the data in $\Dcal_k^{C,1,l}=\left\{Z_i:i\in\bigcup_{k'\in\mathcal H_{k,1}\setminus\{l\}}\Dcal_{k'}\right\}$ to construct a propensity score estimator $\hat\pi^{(k, l)}(1\mid \cdot)$ for $\pi^*(1\mid \cdot)$. Then let $\hthinit\pk$ be given by solving the following estimating equation (or, its least squares solution up to approximation error of $\epsilon_N$): 
\[
\frac{1}{|{\Dcal_k^{C,1}}|}\sum_{l \in \mathcal{H}_{k, 1}}\sum_{i \in \Dcal_l}\psi^{\text{IPW}}(Z_i; \theta, \hat\pi^{(k, l)}) = 0,
~~\text{where}~~\psi^{\text{IPW}}(Z; \theta, \pi) = \frac{\ind(T = 1)}{\pi(1\mid X)}U(Y;\theta_1) + V(\theta_2).\] 
\end{definition}
This procedure is illustrated in \cref{fig: schema}(b).
Note that, given a fixed $\theta_1'$, both $\tpib(1\mid \cdot)$ and $\tq(\cdot,1;\theta_1')$ are conditional expectations of observable variables given $X$.
Thus, in this setting, the whole LDML estimate using the IPW initial estimate can be computed given just blackbox algorithms for (possibly binary) regression.

\subsection{Theoretical Analysis}\label{sec:Theoretical}

We first study the LDML estimate for estimating equations with incomplete data by leveraging our general theory in \cref{thm: general-split}. To this end, we assume a strong form of the overlap condition and specify the convergence rates of the initial estimator and nuisance estimators used. We consider a generic treatment level $t \in \{0, 1\}$ in these two assumptions.
\begin{assumption}[Strong Overlap]\label{assump: unconf}
Assume that there exists a positive constant $\varepsilon_{\pi}>0$ such that for any $\hP \in \mathcal P_N$,
$\pi(t \mid X) \ge \varepsilon_{\pi}$ almost surely.
\end{assumption}
\begin{assumption}[Nuisance Estimation Rates]\label{assump: nuisance-rate}
Assume that 
for any $\hP \in \mathcal P_N$:
condition \ref{assump: error: nuisance-set} of \cref{assump: error} holds for a sequence of constants $\Delta_N\to0$; with probability at least $1 - \Delta_N$, $\hpib\pk(t \mid X) \ge \varepsilon_{\pi}$ for almost all realizations of $X$, and 
\begin{align*}
    &\|\fprns{\hP \fprns{\hq\pk(X, t; \hthinit\pk ) - \tq(X, t; \hthinit\pk )  }^2}^{1/2}\|   \le \rho_{\mu, N},  
    \\
    &
    \fprns{\hP \fprns{\hpib\pk(t \mid X) - \tpib(t \mid X)}^2}^{1/2} \le \rho_{\pi, N}, ~~ \|\hthinit\pk - \theta^*\| \le \rho_{\theta, N}.
 \end{align*}
\end{assumption}
The following theorem establishes that the asymptotic distribution of our proposed estimator is similar to the (infeasible) one that solves the semiparametric efficient estimating equation  in \cref{eq: est-eq-causal} with known nuisances. This theorem is proved by verifying conditions in \cref{thm: general-split}, namely \cref{assump: jacob,assump: identification,assump: error}. 
\begin{theorem}[LDML for Estimating Equations with Incomplete Data]\label{thm: causal}
Fix $t = 1$ and let the estimator $\hat \theta$ be given by 
applying \cref{def: LDML2} to the estimating equation in \cref{eq: est-eq-causal}.
Suppose \cref{assump: unconf,assump: nuisance-rate} hold and that there exist positive constants $c'$, $C$, and $c_1$ to $c_7$ such that for any $\hP\in\mathcal P_N$ the following conditions hold: 
\begin{enumerate}[label=\roman*.]
\item \label{assump: causal: regularity} Conditions \ref{assump: identification: interior} (with $c_1$), \ref{assump: identification: diffable}, \ref{assump: identification: moment} (with $c_5,c_6$), and \ref{assump: identification: metric entropy} (with $c_7$)
 of \cref{assump: identification} and 
  condition \ref{assump: error: approximation} of \cref{assump: error} for the estimating equation in \cref{eq: est-eq-causal}. 
\item \label{assump: causal: Jacobian} For $j = 1, \dots, d$, $\theta \mapsto \hP\left[U_j(Y(t); \theta_1) + V(\theta_2)\right]$ is differentiable at any $\theta$ in a compact set $\Theta$, and each component of its gradient is $c'$-Lipschitz continuous at $\tth$. Moreover, for any $\theta \in \Theta$ with $\|\theta - \theta^*\| \ge \frac{c_3}{2\sqrt{d}c'}$, we have $2\|\hP\left[U(Y(t); \theta_1) + V(\theta_2)\right]\| \ge c_2$.
\item \label{assump: causal: identification} The singular values of $\partial_{\theta^\top}\hP\left[U(Y(t); \theta_1) + V(\theta_2)\right]\vert_{\theta = \tth}$ are bounded between $c_3$ and $c_4$. 
\item \label{assump: causal: exchange-int-diff}    For any $\theta \in \mathcal{B}(\tth; \frac{4C\sqrt{d}\rho_{\pi, N}}{\delta_N\varepsilon_\pi}) \cap \Theta$,  $r \in (0, 1)$, and  $j = 1, \dots, d$, there exist $h_1(x, t; \theta_1),h_2(x, t; \theta_1)$ such that $\hP\left[h_1(X, t; \theta_1)\right] < \infty,\,\hP\left[h_2(X, t; \theta_1)\right] < \infty$ and almost surely 
\[\left|\partial_{r}\tq_j(X, t; \tth_1 + r(\theta_1 - \tth_1))\right| \le h_1(X, t; \theta_1),~~\left|\partial^2_{r}\tq_j(X, t; \tth_1 + r(\theta_1 - \tth_1))\right| \le h_2(X, t; \theta_1).\]
\item \label{assump: causal: boundedness} For $j = 1, \dots, d$ and any $\theta \in \Theta$, we have $\fprns{\hP \fprns{\tq_j(X,t; \theta_1)}^2}^{1/2} \le C$.
\item \label{assump: causal: boundedness2} 
For $j = 1, \dots, d$ and any $\theta \in \mathcal{B}(\tth; \max\{\frac{4C\sqrt{d}\rho_{\pi, N}}{\delta_N\varepsilon_\pi}, \rho_{\theta, N}\})\cap \Theta$.
\begin{align*}
&\left\{\hP \left[\tq_j(X,t; \theta_1) - \tq_j(X,t; \tth_1)\right]^2\right\}^{1/2} \le C\|\theta_1 - \tth_1 \|, ~~ \left\|\left\{\hP\left[\partial_{\theta_{1}}\tq_j(X, t; \theta_1)\right]^2\right\}^{1/2}\right\| \le C,  \\
&\qquad\qquad\qquad \sigma_{\max}\left(\hP \bracks{\partial_{\theta_1}\partial_{\theta^\top_1}\tq_j(X, t; \theta_1)}\right) \le C, ~ \sigma_{\max}\left(\partial_{\theta_2}\partial_{\theta_2^\top}V_j(\theta_2)\right) \le C. 
\end{align*}
\item \label{assump: causal: rate} $\rho_{\pi, N}(\rho_{\mu, N} + C\rho_{\theta, N}) \le \frac{\varepsilon_\pi^3}{3}\delta_N N^{-1/2}$, $\rho_{\pi, N} \le \frac{\delta_N^3}{\log N}$, $\rho_{\mu, N} + C\rho_{\theta, N} \le \frac{\delta^2_N}{\log N}$, $\delta_N \le \frac{4C^2\sqrt{d} + 2\varepsilon_\pi}{\varepsilon_\pi^2}$, and $\delta_N \le \min\{\frac{\varepsilon_\pi^2}{8C^2 d}\log N, \sqrt{\frac{\varepsilon_\pi^3}{2C\sqrt{d}}}\log^{1/2} N\}$. 
\end{enumerate}
Then $\hat\theta$ satisfies the 
conclusion of
\cref{thm: general-split} 
for $\psi(Z; \tth, \tnua(Z; \tth_1), \tnub(Z))$ given in \cref{eq: est-eq-causal}.
\end{theorem}

An analogous result for the estimating equations involving $Y(0)$ holds when 
we change $t=1$ to $t=0$ everywhere in \cref{thm: causal}. See \cref{remark: effect variance} regarding estimation of the difference of the parameters (\ie, the treatment effects) and inference thereon.

In \cref{thm: causal}, conditions \ref{assump: causal: Jacobian} and \ref{assump: causal: identification} guarantee the identification conditions \ref{assump: identification: identification} and \ref{assump: identification: conditioning} of \cref{assump: identification}.  Condition \ref{assump: causal: exchange-int-diff} enables exchange of integration, which together with conditions \ref{assump: causal: boundedness}, \ref{assump: causal: boundedness2}, and \ref{assump: causal: rate} imply the rate condition \ref{assump: error: rate-condition} of \cref{assump: error}. 
Note condition \ref{assump: causal: rate} permits nonparametric rates for nuisance estimators. Focusing on the order in the sample size and up to polylog factors, the condition allows for  $\rho_{\pi,N}\rho_{\mu,N}=o(N^{-1/2}),\,\rho_{\pi,N}\rho_{\theta,N}=o(N^{-1/2}),\,\rho_{\pi,N}=o(1),\,\rho_{\mu,N}=o(1),\,\rho_{\theta,N}=o(1).$
Note the first two restrictions are on \emph{products}, permitting a trade-off between rates for different nuisances (see also \cref{sec: compare-discussion}).
 \begin{remark}[Rate Conditions with IPW Initial Estimator]\label{remark: IPW-init}
In \cref{sec: ipw initial}, we prove that if the propensity nuisance estimators used to construct  the IPW initial estimators (\cref{def: IPW_init}) also have convergence rate $\rho_{\pi, N}$, then the initial estimators' convergence rates satisfy that $\rho_{\theta, N} = O\prns{\rho_{\pi, N}}$. 
In this case, we are essentially imposing $\rho_{\pi, N} = o(N^{-1/4})$: condition \ref{assump: causal: rate} of \cref{thm: causal}  requires $\rho_{\pi, N}\rho_{\theta, N} = o(N^{-1/2})$, so  unless $\rho_{\theta, N}$ is somehow even faster than $\rho_{\pi, N}$, we must need both $\rho_{\theta, N}$ and $\rho_{\pi, N}$ to be $o(N^{-1/4})$. 
 \end{remark}

\subsection{Quantile and CVaR} 

Now we consider estimating quantile and (possibly) CVaR based on the semiparametrically efficient estimating equation in \cref{eq: quantile-only}.
Instantiating \cref{eq: est-eq-causal} for the simultaneous estimation of quantile and CVaR and rearranging, we obtain the following estimating equation:
\begin{align}
&\psi(Z;\theta,\eta^*_1(Z;\theta_1),\eta^*_2(Z))=
    \frac{\indic{T=1}}{\eta_2^*(Z)}
        \begin{bmatrix} 
            \ind[Y \leq \theta_1 ] - \eta^*_{1,1}(Z;\theta_1)\\
            \frac{1}{1 - \gamma}\big(\max(Y - \theta_1,0) - \eta^*_{1,2}(Z;\theta_1)\big)
        \end{bmatrix}
         + \begin{bmatrix}
            \eta_{1,1}^*(Z;\theta_1) - \gamma \\
            \theta_1 + \frac{1}{1 - \gamma}\eta^*_{1,2}(Z;\theta_1) -\theta_2
        \end{bmatrix},\nonumber\\
&\text{where}\quad\eta^*_1(Z;\theta_1)
=\begin{bmatrix} 
\Prb{Y\leq \theta_1\mid X,T=1}\\
\Eb{\max(Y - \theta_1,0) \mid X, T=1}
\end{bmatrix}, ~~ \eta_2^*(Z)=\Prb{T=1\mid X}.\label{eq: quantile-cvar}
\end{align}

We use $F_{t}(\cdot \mid x)$ and $F_t(\cdot)$ to denote the conditional and unconditional cumulative distribution function of $Y(t)$, respectively: 
for any $y$, $F_{t}(y \mid x) = \hP(Y(t) \le y \mid X = x)$ and $F_{t}(y) = \hP(Y(t) \le y )$.
The following proposition gives the asymptotic behavior of our proposed estimators for the quantile and CVaR of  $Y(1)$. This conclusion is proved by verifying all conditions in \cref{thm: causal}.
Analogous conclusions also hold for $Y(0)$ when all assumptions hold for $t = 0$ instead of $t=1$.
\begin{proposition}[LDML for Quantile and CVaR]\label{thm: quantile only}
Fix $t = 1$ and Let the estimator $\hat \theta$ be given by applying \cref{def: LDML2} to the estimating function in \cref{eq: quantile-cvar}.
Suppose \cref{assump: unconf,assump: nuisance-rate} hold and there exist positive constants $c_1' \sim c_5'$ and $C \ge 1$, such that for any $\hP \in \mathcal P_N$, the following conditions hold:
\begin{enumerate}[label=\roman*.]
\item \label{assump: quantile only: regularity} Conditions \ref{assump: identification: interior} (with $c_1$), \ref{assump: identification: diffable}, \ref{assump: identification: moment} (with $c_5,c_6$)
 of \cref{assump: identification},
  condition \ref{assump: error: approximation} of \cref{assump: error}, and condition \ref{assump: causal: rate} of \cref{thm: causal} for the estimating function in \cref{eq: quantile-cvar}  and the corresponding nuisance estimators.
\item\label{thm: quantile only: diffable marg} $F_t(\theta_1)$ is twice differentiable with derivatives $f_t(\theta_1), \dot{f}_t(\theta_1)$ satisfying $0 < c_1' \le f_t(\theta_1^*)$, $f_t(\theta_1) \le c_2'$, $|\dot{f}_t(\theta_1)| \le c_3'$ $\forall\theta_1 \in \Theta_1$. Moreover, $|F_t(\theta^*_1) - F_t(\theta_1)| \ge c_4'$ for $|\theta_1 - \tth_1| \ge 
\frac{c_1'}{2c_3'}$.
\item\label{thm: quantile only: diffable cond} At any $\theta \in \mathcal{B}(\tth; \max\{\frac{4C\sqrt{d}\rho_{\pi, N}}{\delta_N\varepsilon_\pi}, \rho_{\theta, N}\})\cap \Theta$, $F_{t}(\theta_1 \mid X)$ is twice differentiable almost surely with first two order derivatives $f_{t}(\theta_1 \mid X)$ and $\dot{f}_{t}(\theta_1 \mid X)$ that satisfy $f_t(\theta_1 \mid X)\le C$ and $|\dot{f}_{t}(\theta_1 \mid X)| \le C$ almost surely.
\item\label{thm: quantile only: cvar cond} $2\|{\overline\hP[U(Y(t); \theta_1) + V(\theta_2)]}\| \ge c_5'$ for $\|\theta - \tth\| \ge \frac{\min\{\gamma, \prns{1 - \gamma}c_1', \gamma c_1'\}}{4\sqrt{2}\gamma\max\{c_2', c_3'\}}$ and $U(Y(t); \theta_1) + V(\theta_2)$ as given in \cref{eq: est-eq-cvar-original}. 
\item\label{thm: quantile only: second-norm} $\prns{\hP{\prns{\expect[\max(Y - \theta_1, 0) \mid X,T=t]^2}}}^{1/2}\le C$ for any $\theta \in \Theta$. 
\end{enumerate}
Then $\hat\theta$ satisfies the 
conclusion of
\cref{thm: general-split} 
for
$\psi(Z; \tth, \tnua(Z; \tth_1), \tnub(Z))$ given in \cref{eq: quantile-cvar} and for $J^*=\op{diag}\prns{f_{t}(\tth_1),\,-1}$.
Moreover, under all conditions above except conditions \ref{thm: quantile only: cvar cond} and \ref{thm: quantile only: second-norm},  the  quantile estimator $\hat\theta_1$ alone still satisfies the analogous asymptotic linear expansion for $\psi(Z; \tth, \tnua(Z; \tth_1), \tnub(Z))$ given in \cref{eq: quantile-only} and for $J^* = f_{t}(\tth_1)$.
\end{proposition}
\begin{remark}[Estimating $f_{t}(\tth_1)$ for Variance Estimation]\label{rem:variancee_quantile}
If we want to conduct inference on the quantile or QTE using our method from \cref{sec: inference}, we need to estimate $f_{t}(\tth_1)$. We only need to do this consistently, regardless of rate, in order to get the correct asymptotic coverage. One simple approach is to use cross-fitted IPW kernel density estimation at $\hat\theta_1$:
$$\hat J=
\frac1{Nh}{\sum_{k=1}^K\sum_{i\in\Dcal_k}\frac{\indic{T_i=1}}{\hpib\pk(1\mid X_i)}\kappa((Y_i-\hat\theta_1)/h)},
$$
where $\kappa(u)$ is a kernel function such as $\kappa(u)=(2\pi)^{-1/2}\exp(-u^2/2)$ and $h\to0$ is a bandwidth.
Under \cref{assump: unconf}, $h\asymp N^{-1/5}$ would be the optimal bandwidth.
While this together with any consistent estimate $\hpib\pk$ suffices for asymptotic coverage, the estimate may be unstable. It is therefore recommended to use self-normalization by dividing the above by $\frac1{n}{\sum_{k=1}^K\sum_{i\in\Dcal_k}\frac{\indic{T_i=1}}{\hpib\pk(1\mid X_i)}}$ and to potentially clip propensities. 
\end{remark}

\section{Empirical Results}

We first study the behavior of LDML in a simulation study. We then demonstrate its use in estimating the QTE of 401(k) eligibility on net financial assets. In \cref{sec: IV}, we additionally consider estimating the \emph{L}QTE of 401(k) participation using eligibility as IV.

\subsection{Simulation Study}

First, we consider a simulation study to compare the performance of LDML estimates to benchmarks. We consider estimating $\theta^*_1$ as the second tertile of $Y(1)$ from incomplete data. The distribution $\hP$ is as follows. First, we draw $20$-dimensional covariates $X$ from the uniform distribution on $[0,1]^{20}$. Then, we draw $T$ from $\operatorname{Bernoulli}(\Phi(3(1-X_1-X_3)))$, where $\Phi$ is the standard normal cumulative distribution function, and we draw $Y(1)$ from $\mathcal N(\mathbb I[X_1+X_2\leq 1],\,2X_3)$. We only observe $Y(1)$ when $T=1$.

We consider estimating $\theta^*_1$ using four different methods. First, we consider LDML applied to the efficient estimating equation (\cref{eq: quantile-only}) with $K=5,\,K'=2$, $\hthinit\pk$ estimated using 2-fold cross-fitted IPW with random-forest-estimated propensities, and $\hpib\pk(1\mid X),\,\hq\pk(X,1;\hthinit\pk)$ similarly estimated by random forests. Second, we consider $K=5$-fold cross-fitted IPW with random-forest-estimated propensities. Third, we consider DML with $K=5$ and the estimand-dependent nuisance estimated using a discretization approach similar to the suggestion of \citet{belloni2018high}: for $j=1,\dots,99$,  fix $\theta_{1,j}$ to be the $j/100$ marginal quantile of $Y$ and fit $\hq\pk(X,1;\theta_{1,j})$ using random forests; then apply DML with the restricted discretized estimand range $\{\theta_{1,j}:j=1,\dots,99\}$. We refer to this method as DML-D for \emph{discretized}. Fourth, we consider DML with $K=5$ and where the estimand-dependent nuisance is estimated using an approach similar to \citet{meinshausen2006quantile,bertsimas2014predictive}: namely, fit a random forest regression to the out-of-fold data $\{(X_i,Y_i):i\notin\Dcal_k,T_i=1\}$ to obtain $B$ regression trees $\tau_j:\operatorname{support}(X)\to\{1,\dots,\ell_j\}$, then set $\hq\pk(X,1;\theta_1)=\sum_{i\notin\Dcal_k:T_i=1}\frac{\mathbb I[Y_i\leq \theta_1]}B\sum_{j=1}^B\frac{\mathbb I[\tau_j(X_i)=\tau_j(X)]}{\sum_{i'\notin\Dcal_k:T_{i'}=1}\mathbb I[\tau_j(X_i)=\tau_j(X_{i'})]}$ for all $\theta_1$. We refer to this method as DML-F for \emph{forest}. For each method, we run it three times with new random fold splits (with the same data) and take the median of the three results to be the estimate.

\begin{figure}\centering%
\includegraphics[width=\textwidth]{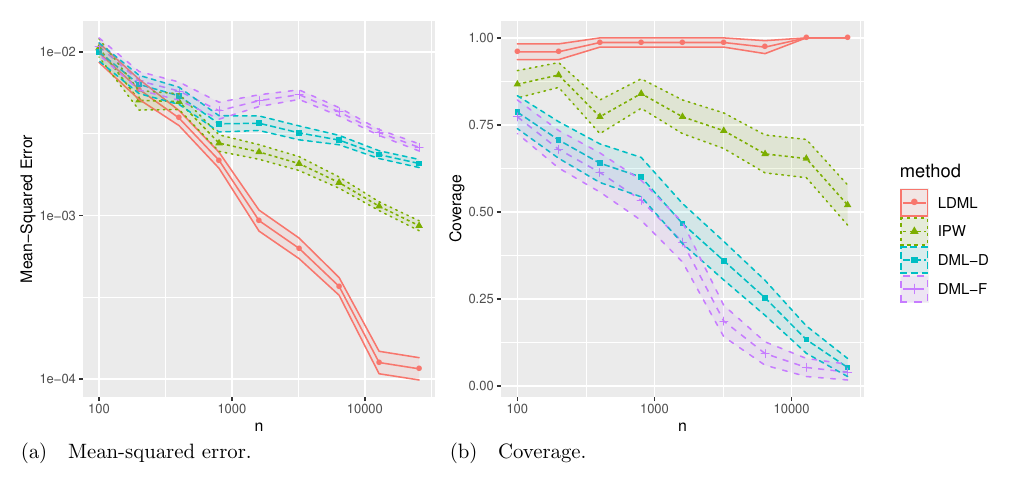}
\caption{Results for the simulation study. Shaded regions denote plus/minus one standard error for estimated mean-squared error or mean coverage computed over 75 replications.}
\label{fig: sim MSE}
\end{figure}

For each of $n=100,200,\dots,25600$, we consider 75 replications of drawing a dataset of size $n$ and constructing each of the above four estimates. We plot the mean-squared error of each method and $n$ over the 75 replications in \cref{fig: sim MSE}(a). The shaded regions show plus/minus one standard error of this as the sample mean of 75 squared errors. We clearly see that LDML offers significant improvements over the other methods when we use flexible machine learning methods to tackle estimand-dependent nuisances. 

In \cref{fig: sim MSE}(b), we additionally report the coverage of the true parameter using the standard error estimate proposed in \cref{rem:variancee_quantile} and \cref{sec: practical}.
 Namely, for each of the three random runs of each method, we take the sample standard deviation of the estimated influence function evaluated at the final estimand and with the cross-fitted nuisances and divide it by $\sqrt{n}$ times an estimate of $f_t(\theta_1^*)$ given by cross-fitted IPW kernel density estimation at the estimand.
(We do the same for the IPW estimate for the sake of comparison but note IPW's asymptotic variance may also depend on the propensity-estimation variance, unlike LDML and DML.) 
We take the median of these standard errors over the three runs and add to it the standard deviation of estimands over the three runs divided by $\sqrt{3}$. Then we consider the 95\% confidence interval given by the estimand plus/minus 1.96 of this estimated standard error. \Cref{fig: sim MSE}(b) shows the sample mean coverage of $\theta_1^*$ over the 75 replications, and the shaded region shows plus/minus one standard error of this sample mean. 
{ LDML offers good, calibrated coverage (the 100\% coverage for some $n$ can be attributed to only observing 75 replications with 95\% success probability each). The other methods have poor coverage, which may be attributed to significant bias so that confidence intervals based only on standard errors of the sample average would undercover and even get worse as samples grow and standard errors shrink relative to bias. In particular, the IPW estimate's convergence directly depends on that of the random-forest-estimated propensities, so convergence may be slower than $\sqrt{n}$ and/or the true standard error may be far larger than that of the cross-fitted sample average. Similarly, using DML to estimate the control-variate term using a discretization or a forest need not converge, so ultimately their convergence and standard errors may be similar to IPW's, again leading to underconvering.}

\begin{table}[t!]
\caption{The QTE of 401(k) eligibility in thousand dollars (and standard error) estimated by LDML using different regression methods, and raw unadjusted differences of quantiles.}
\centering{\footnotesize\spacingset{1}\begin{tabular}{llllllllll}
\toprule
$\gamma$ & $K$ & LASSO & Neural Net & Boosting & Forest & Raw\\ 
  \midrule
 \multirow{3}{*}{25\%} & 5 & 0.95 (0.24) & 1.05 (0.19) & 1.00 (0.20) & 0.93 (0.29) & \multirow{3}{*}{1.50 (0.25)} \\ 
    & 15 & 0.95 (0.24) & 1.06 (0.20) & 1.00 (0.20) & 0.93 (0.28) \\ 
    & 25 & 0.95 (0.24) & 1.03 (0.20) & 1.00 (0.20) & 0.93 (0.29) \\ \midrule
   \multirow{3}{*}{50\%} & 5 & 4.74 (0.68) & 5.56 (0.69) & 4.47 (0.85) & 3.64 (1.87) & \multirow{3}{*}{8.98 (0.41)} \\ 
    & 15 & 4.68 (0.68) & 5.59 (0.68) & 4.47 (0.85) & 3.46 (1.85) \\ 
    & 25 & 4.68 (0.68) & 5.55 (0.67) & 4.47 (0.85) & 3.45 (1.85) \\ \midrule
   \multirow{3}{*}{75\%} & 5 & 14.00 (4.14) & 17.12 (4.10) & 13.28 (5.11) & 13.88 (11.32) & \multirow{3}{*}{29.67 (1.35)} \\ 
    & 15 & 13.94 (4.12) & 16.86 (4.01) & 13.29 (5.20) & 14.30 (12.11) \\ 
    & 25 & 13.93 (4.13) & 16.87 (4.00) & 13.29 (5.16) & 14.29 (12.23) \\ 
 \bottomrule
\end{tabular}}\label{table: 401k qte}
\end{table}

\subsection{Effect of 401(k) Eligibility on Net Financial Assets}\label{sec: 401k eligibility}
Next we consider an empirical case study to demonstrate the estimation of QTE using LDML in practice and with a variety of machine learning nuisance estimators.
We use data from \citet{chernozhukov2004effects} to estimate the QTEs of 401(k) retirement plan eligibility on net financial assets. Eligibility for 401(k) (here considered the treatment, $T$) is not randomly assigned, but is argued in \citet{chernozhukov2004effects} to be ignorable conditioned on certain covariates: age, income, family size, years of education, marital status, two-earner household status, availability of defined benefit pension plan to household, IRA participation, and home ownership status. Net financial assets (the outcome, $Y$) are defined as the sum of IRA and 401(k) balances, bank accounts, and other interest-earning accounts and assets minus non-mortgage debt. While \citet{chernozhukov2004effects} considered controlling for these in a low-dimensional linear specification, it is not clear whether such is sufficient to account for all confounding. Consequently, \citet{belloni2017program} considered including higher-order terms and interactions, but needed to theoretically construct a continuum of LASSO estimates and may not be able to use generic black-box regression methods. Finally, \citet{ChernozhukovVictor2018Dmlf} considered using generic machine learning methods, but only tackled ATE estimation.

In contrast, we will use LDML to estimate and conduct inference on the QTEs of 401(k) eligibility on net assets using a variety of flexible black-box regression methods. First, to understand the effect of different choices in the application of LDML to the problem, we consider estimating the 25\%, 50\%, and 75\% QTE while varying $K$ in $\{5,15,25\}$ and varying the nuisance estimators. We consider estimating both propensity score $\eta_2^*$ and conditional cumulative distribution $\eta_1^*$ with each of: boosting (using \emph{R} package gbm), LASSO (using \emph{R} package hdm), and a one-hidden-layer neural network (using \emph{R} package nnet).
For LASSO, we use a 275-dimensional expansion of the covariates by considering higher-order terms and interactions.
In each instantiation of LDML, 
we construct folds so to ensure a balanced distribution of treated and untreated units,
we let $K'=(K-1)/2$, we use the IPW initial estimator for $\hthinit$,
we normalize propensity weights to have mean 1 within each treatment group,
we use estimates given by solving the grand-average estimating equation as in \cref{def: LDML2}, and for variance estimation we estimate $J^*$ using IPW kernel density estimation as in \cref{rem:variancee_quantile}. The solution to the LDML-estimated empirical estimating equation must occur at an observed outcome $Y_i$ and that we can find the solution using binary search after sorting the data along outcomes.
We re-randomize the fold construction and repeat each instantiation 100 times. We then remove the outlying 2.5\% from each end and report $\hat\theta^\text{mean},\,\widehat\Sigma^\text{mean}$ as in \cref{sec: practical}.
The resulting estimates and standard errors are shown in \cref{table: 401k qte}.
The estimates appear overall roughly stable across methods and $K$. 

\begin{figure}[t!]\centering%
\includegraphics[width=0.9\textwidth]{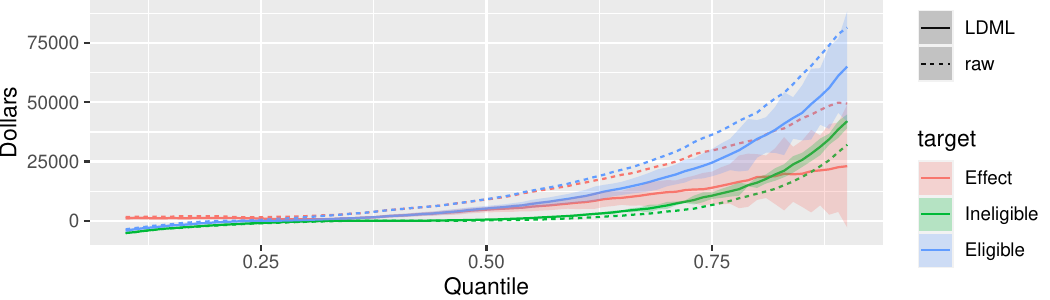}
\caption{LDML estimates of a range of quantiles and QTEs with confidence 90\% intervals and comparison to raw unadjusted marginal quantiles by treatment group.%
}
\label{fig: 401k qte}
\bigskip\end{figure}

Next, we consider estimating a range of QTEs. We focus on nuisance estimation using LASSO and fix $K=15$. We then estimate the $10\%,11\%,\,\dots,\,89\%,\,$ and $90\%$ quantiles and QTEs. We plot the resulting LDML estimates with $90\%$ confidence intervals in \cref{fig: 401k qte} and compare these to the raw unadjusted marginal quantiles within each treatment group.

\section{Related Literature}  \label{sec:related}

\paragraph*{Semiparametric Estimation, Neyman orthogonality, and Debiased Machine Learning.}
Our work is closely related to the classical semiparametric estimation literature on constructing $\sqrt{N}$-consistent and asymptotically normal estimators for low dimensional target parameters in the presence of infinitely dimensional nuisances, typically estimated by conventional nonparametric estimators such as kernel or series estimators \citep[e.g.,][]{newey1990semiparametric,newey1994the,newey1998undersmoothing,ibragimov1981statistical,levit1976on,bickel98,bickel1982on,robinson1988root,vaart1991on,andrews1994asymptotics,robins1995semiparametric,linton1996edgeworth,chen2003estimation,ai2003efficient,laan2011targeted,AiChunrong2009Sebf}.
Our work builds on the Neyman orthogonality condition introduced by \cite{Neyman1959}). 
This condition plays a critical role in many works that go beyond such nonparametric estimators, such as targeted learning \citep[e.g.,][]{laan2011targeted,van2018targeted},  inference for coefficients in high dimensional linear models \citep[e.g.,][]{belloni2016post,belloni2014pivotal,zhang2014confidence,van2014asymptotically,javanmard2014confidence,chernozhukov2015valid,ning2017general}, and semiparametric estimation with nuisances that involve  high dimensional covariates \citep[e.g., ][]{belloni2017program,smucler2019unifying,chernozhukov2018plug,farrell2015robust, belloni2014high,belloni2014inference,bradic2019sparsity,bravo2020two}. 

\cite{ChernozhukovVictor2018Dmlf}  further advocate
the use of cross-fitting in addition to orthogonal estimating equations, so that the traditional Donsker assumption on nuisance estimators can be relaxed, and a broad array of  black-box machine learning algorithms can be used instead. 
They refer to this generic approach as DML, which provides a principled framework to estimate low-dimensional target parameters with strong asymptotic guarantees when leveraging modern machine learning methods in nuisance estimation.  
Similar forms of sample splitting and cross-fitting have also appeared in
\citet{klaassen1987consistent,zheng2011cross,fan2012variance,bickel1982on,robins2013new,schick1986asymptotically,robins2008higher,robins2017minimax}.
Since the DML framework was introduced, numerous works have applied it in many different problems, such as heterogeneous treatment effect estimation \citep{kennedy2020optimal,nie2017quasi,curth2020semiparametric,semenova2020debiased,oprescu2019orthogonal,fan2020estimation}, causal effects of continuous treatments \citep{colangelo2020double,oprescu2019orthogonal}, instrumental variable estimation \citep{singh2019biased,syrgkanis2019machine}, partial identification \citep{bonvini2019sensitivity,kallus2019assessing,semenova2017machine,yadlowsky2018bounds}, difference-in-difference models \citep{lu2019robust,chang2020double,zimmert2018efficient}, off-policy evaluation \citep{kallus2020double,demirer2019semi,zhou2018offline,athey2017efficient}, generalized method of moments \citep{chernozhukov2016locally,belloni2018high}, improved machine learning nuisance estimation \citep{farrell2018deep,cui2019bias}, statistical learning with nuisances \citep{foster2019orthogonal}, causal inference with surrogate observations \citep{kallus2020role}, linear functional estimation \citep{chernozhukov2018learning,chernozhukov2018double,bradic2019minimax}, etc.
Our work complements this line of research by proposing a simple but effective way to handle estimand-dependent nuisances. This type of nuisances frequently appears in efficient estimation of complex causal effects such as QTEs, and applying DML directly would require estimating a continuum of nuisances, which is challenging in practice.

\paragraph*{Efficient estimation of (L)QTE.}\cite{FirpoSergio2007ESEo} first considered efficient estimation of QTE and proposed an IPW estimator based on propensity scores estimated by a logistic sieve estimator. Under strong smoothness conditions, this IPW estimator is $\sqrt{N}$-consistent and achieves the semiparametric efficiency bound.
\cite{frolich2007unconditional} consider a weighted estimator for LQTE with weights estimated by local linear regressions using high-order kernels and show that their estimator is also semiparametrically efficient. 
Although these purely weighted methods bypass the estimation of nuisances that depend on the estimand, their favorable behavior is restricted to certain nonparametric weight estimators and strong smoothness requirements.
\cite{diaz2017efficient} proposed a Targeted Minimum Loss Estimator (TMLE) estimator for efficient QTE estimation. Built on the efficient influence function with nuisances that depends on the quantile itself, this estimator requires estimating a whole conditional cumulative distribution function, which as discussed may be very challenging in practice using flexible machine learning methods.   
\cite{belloni2017program} similarly consider efficient estimation of LQTE with high-dimensional covariates by using a Neyman-orthogonal estimating equation and discretizing a continuum of LASSO estimators for the estimand-dependent nuisance.
In contrast, our proposed estimator can leverage a wide variety of flexible machine learning methods for the standard regression task to estimate nuisances, since we require estimating conditional cumulative distribution function only at a \emph{single} point, which amounts to a binary regression problem. 

\paragraph*{Estimand-dependent nuisances.} Besides (local) quantiles and CVaR, many efficient estimation problems involve nuisances that depends on the estimand \citep[e.g.,][]{tsi,chen2005measurement}.
Previous approaches estimate the whole continuum of the estimand-dependent nuisances either by positing simple parametric model for conditional distributions \citep[Chap 10]{tsi}, using sieve estimators \citep{chen2005measurement}, or discretizing a hypothetical continuum of regression estimators \citep{belloni2017program}. In contrast, our proposed method obviates the need to estimate infinitely many nuisances by fitting  nuisances only at a preliminary estimate of the parameter of interest.
This idea was briefly mentioned by 
\cite{robins1994estimation}, focusing on  parametric  models for nuisance estimation. Our paper rigorously develops this approach  
 and
 admits flexible machine learning methods for estimating nuisances that depend on the estimand.

\section{Conclusion}

In many causal inference and missing data settings, the efficient influence function involves nuisances that depend on the estimand of interest. A key example provided was that of QTE under ignorable treatment assignment and LQTE estimation using an IV, where in both cases the efficient influence function depends on the conditional cumulative distribution function evaluated at the quantile of interest. This structure, common to many other important problems, makes the application of existing debiased machine learning methods difficult in practice. In quantile estimation, it requires we learn the whole conditional cumulative distribution function. To avoid this difficulty, we proposed the LDML approach, which localized the nuisance estimation step to an initial rough guess of the estimand. 
This was motivated by the fact that in many applications, the oracle estimating equation is asymptotically equivalent to one where the nuisance is evaluated at the true parameter value, which our localization approach targets. 
Assuming only standard identification conditions, Neyman orthogonality, and lax rate conditions on our nuisance estimates, we proved the LDML enjoys the same favorable asymptotics as the oracle estimator that solves the estimating equation with the \emph{true} nuisance functions.
This newly enables the practical efficient estimation of important quantities such as QTEs using machine learning.

{ An interesting future direction is to consider a uniform estimation way over the quantile $\gamma$ though we consider the setting in which $\tau$ is fixed to emphasize our main point. This might be possible by combining recent technique in quantile regression \citep{ota2019quantile,bradic2017uniform}; however, the rigorous result is left as future research. }

\bibliographystyle{plainnat}
\bibliography{arxiv_lit}

\begin{thebibliography}{86}
\providecommand{\natexlab}[1]{#1}
\providecommand{\url}[1]{\texttt{#1}}
\expandafter\ifx\csname urlstyle\endcsname\relax
  \providecommand{\doi}[1]{doi: #1}\else
  \providecommand{\doi}{doi: \begingroup \urlstyle{rm}\Url}\fi

\bibitem[Ai and Chen(2003)]{ai2003efficient}
Chunrong Ai and Xiaohong Chen.
\newblock Efficient estimation of models with conditional moment restrictions
  containing unknown functions.
\newblock \emph{Econometrica}, 71\penalty0 (6):\penalty0 1795--1843, 2003.

\bibitem[Ai and Chen(2012)]{AiChunrong2009Sebf}
Chunrong Ai and Xiaohong Chen.
\newblock Semiparametric efficiency bound for models of sequential moment
  restrictions containing unknown functions.
\newblock \emph{Journal of Econometrics}, 170:\penalty0 442--457, 2012.

\bibitem[{Andrews}(1994)]{andrews1994asymptotics}
Donald W.~K. {Andrews}.
\newblock Asymptotics for semiparametric econometric models via stochastic
  equicontinuity.
\newblock \emph{Econometrica}, 62\penalty0 (1):\penalty0 43--72, 1994.

\bibitem[Athey and Wager(2017)]{athey2017efficient}
Susan Athey and Stefan Wager.
\newblock Efficient policy learning.
\newblock \emph{arXiv preprint arXiv:1702.02896}, 2017.

\bibitem[Belloni et~al.(2014{\natexlab{a}})Belloni, Chernozhukov, and
  Hansen]{belloni2014high}
Alexandre Belloni, Victor Chernozhukov, and Christian Hansen.
\newblock High-dimensional methods and inference on structural and treatment
  effects.
\newblock \emph{Journal of Economic Perspectives}, 28\penalty0 (2):\penalty0
  29--50, 2014{\natexlab{a}}.

\bibitem[Belloni et~al.(2014{\natexlab{b}})Belloni, Chernozhukov, and
  Hansen]{belloni2014inference}
Alexandre Belloni, Victor Chernozhukov, and Christian Hansen.
\newblock Inference on treatment effects after selection among high-dimensional
  controls.
\newblock \emph{The Review of Economic Studies}, 81\penalty0 (2):\penalty0
  608--650, 2014{\natexlab{b}}.

\bibitem[Belloni et~al.(2014{\natexlab{c}})Belloni, Chernozhukov, and
  Wang]{belloni2014pivotal}
Alexandre Belloni, Victor Chernozhukov, and Lie Wang.
\newblock Pivotal estimation via square-root lasso in nonparametric regression.
\newblock \emph{The Annals of Statistics}, 42\penalty0 (2):\penalty0 757--788,
  2014{\natexlab{c}}.

\bibitem[Belloni et~al.(2016)Belloni, Chernozhukov, and Wei]{belloni2016post}
Alexandre Belloni, Victor Chernozhukov, and Ying Wei.
\newblock Post-selection inference for generalized linear models with many
  controls.
\newblock \emph{Journal of Business \& Economic Statistics}, 34\penalty0
  (4):\penalty0 606--619, 2016.

\bibitem[Belloni et~al.(2017)Belloni, Chernozhukov, Fern{\'a}ndez-Val, and
  Hansen]{belloni2017program}
Alexandre Belloni, Victor Chernozhukov, Ivan Fern{\'a}ndez-Val, and Christian
  Hansen.
\newblock Program evaluation and causal inference with high-dimensional data.
\newblock \emph{Econometrica}, 85\penalty0 (1):\penalty0 233--298, 2017.

\bibitem[{Belloni} et~al.(2018){Belloni}, {Chernozhukov}, {Chetverikov},
  {Hansen}, and {Kato}]{belloni2018high}
Alexandre {Belloni}, Victor {Chernozhukov}, Denis {Chetverikov}, Christian
  {Hansen}, and Kengo {Kato}.
\newblock High-dimensional econometrics and regularized gmm.
\newblock \emph{arXiv preprint arXiv: 1806.01888}, 2018.

\bibitem[Bertsimas and Kallus(2014)]{bertsimas2014predictive}
Dimitris Bertsimas and Nathan Kallus.
\newblock From predictive to prescriptive analytics.
\newblock \emph{arXiv preprint arXiv:1402.5481}, 2014.

\bibitem[{Bickel}(1982)]{bickel1982on}
P.~J. {Bickel}.
\newblock On adaptive estimation.
\newblock \emph{Annals of Statistics}, 10\penalty0 (3):\penalty0 647--671,
  1982.

\bibitem[Bickel et~al.(1998)Bickel, Klaassen, Ritov, and Wellner]{bickel98}
P.~J. Bickel, C.~A.~J. Klaassen, Y.~Ritov, and J.~A. Wellner.
\newblock \emph{Efficient and Adaptive Estimation for Semiparametric Models}.
\newblock Springer, 1998.

\bibitem[Bonvini and Kennedy(2019)]{bonvini2019sensitivity}
Matteo Bonvini and Edward~H Kennedy.
\newblock Sensitivity analysis via the proportion of unmeasured confounding.
\newblock \emph{arXiv preprint arXiv:1912.02793}, 2019.

\bibitem[Bradic and Kolar(2017)]{bradic2017uniform}
Jelena Bradic and Mladen Kolar.
\newblock Uniform inference for high-dimensional quantile regression: linear
  functionals and regression rank scores.
\newblock \emph{arXiv preprint arXiv:1702.06209}, 2017.

\bibitem[{Bradic} et~al.(2019){Bradic}, {Chernozhukov}, {Newey}, and
  {Zhu}]{bradic2019minimax}
Jelena {Bradic}, Victor {Chernozhukov}, Whitney~K. {Newey}, and Yinchu {Zhu}.
\newblock Minimax semiparametric learning with approximate sparsity.
\newblock \emph{arXiv preprint arXiv:1912.12213}, 2019.

\bibitem[Bradic et~al.(2019)Bradic, Wager, and Zhu]{bradic2019sparsity}
Jelena Bradic, Stefan Wager, and Yinchu Zhu.
\newblock Sparsity double robust inference of average treatment effects.
\newblock \emph{arXiv preprint arXiv:1905.00744}, 2019.

\bibitem[Bravo et~al.(2020)Bravo, Escanciano, and Van~Keilegom]{bravo2020two}
Francesco Bravo, Juan~Carlos Escanciano, and Ingrid Van~Keilegom.
\newblock Two-step semiparametric empirical likelihood inference.
\newblock \emph{The Annals of Statistics}, 48\penalty0 (1):\penalty0 1--26,
  2020.

\bibitem[Chang(2020)]{chang2020double}
Neng-Chieh Chang.
\newblock Double/debiased machine learning for difference-in-differences
  models.
\newblock \emph{The Econometrics Journal}, 23\penalty0 (2):\penalty0 177--191,
  2020.

\bibitem[{Chen} et~al.(2003){Chen}, {Linton}, and
  {Keilegom}]{chen2003estimation}
Xiaohong {Chen}, Oliver {Linton}, and Ingrid~Van {Keilegom}.
\newblock Estimation of semiparametric models when the criterion function is
  not smooth.
\newblock \emph{LSE Research Online Documents on Economics}, 2003.

\bibitem[Chen et~al.(2005)Chen, Hong, and Tamer]{chen2005measurement}
Xiaohong Chen, Han Hong, and Elie Tamer.
\newblock Measurement error models with auxiliary data.
\newblock \emph{The Review of Economic Studies}, 72\penalty0 (2):\penalty0
  343--366, 2005.

\bibitem[Chernozhukov and Hansen(2004)]{chernozhukov2004effects}
Victor Chernozhukov and Christian Hansen.
\newblock The effects of 401(k) participation on the wealth distribution: an
  instrumental quantile regression analysis.
\newblock \emph{Review of Economics and statistics}, 86\penalty0 (3):\penalty0
  735--751, 2004.

\bibitem[Chernozhukov et~al.(2015)Chernozhukov, Hansen, and
  Spindler]{chernozhukov2015valid}
Victor Chernozhukov, Christian Hansen, and Martin Spindler.
\newblock Valid post-selection and post-regularization inference: An
  elementary, general approach.
\newblock 2015.

\bibitem[Chernozhukov et~al.(2016)Chernozhukov, Escanciano, Ichimura, Newey,
  and Robins]{chernozhukov2016locally}
Victor Chernozhukov, Juan~Carlos Escanciano, Hidehiko Ichimura, Whitney~K
  Newey, and James~M Robins.
\newblock Locally robust semiparametric estimation.
\newblock \emph{arXiv preprint arXiv:1608.00033}, 2016.

\bibitem[Chernozhukov et~al.(2018{\natexlab{a}})Chernozhukov, Chetverikov,
  Demirer, Duflo, Hansen, Newey, and Robins]{ChernozhukovVictor2018Dmlf}
Victor Chernozhukov, Denis Chetverikov, Mert Demirer, Esther Duflo, Christian
  Hansen, Whitney Newey, and James Robins.
\newblock Double/debiased machine learning for treatment and structural
  parameters.
\newblock \emph{Econometrics Journal}, 21:\penalty0 C1--C68,
  2018{\natexlab{a}}.

\bibitem[Chernozhukov et~al.(2018{\natexlab{b}})Chernozhukov, Nekipelov,
  Semenova, and Syrgkanis]{chernozhukov2018plug}
Victor Chernozhukov, Denis Nekipelov, Vira Semenova, and Vasilis Syrgkanis.
\newblock Plug-in regularized estimation of high-dimensional parameters in
  nonlinear semiparametric models.
\newblock \emph{arXiv preprint arXiv:1806.04823}, 2018{\natexlab{b}}.

\bibitem[Chernozhukov et~al.(2018{\natexlab{c}})Chernozhukov, Newey, Robins,
  and Singh]{chernozhukov2018double}
Victor Chernozhukov, Whitney Newey, James Robins, and Rahul Singh.
\newblock Double/de-biased machine learning of global and local parameters
  using regularized riesz representers.
\newblock \emph{arXiv preprint arXiv:1802.08667}, 2018{\natexlab{c}}.

\bibitem[Chernozhukov et~al.(2018{\natexlab{d}})Chernozhukov, Newey, and
  Singh]{chernozhukov2018learning}
Victor Chernozhukov, Whitney~K Newey, and Rahul Singh.
\newblock Learning l2 continuous regression functionals via regularized riesz
  representers.
\newblock \emph{arXiv preprint arXiv:1809.05224}, 8, 2018{\natexlab{d}}.

\bibitem[Colangelo and Lee(2020)]{colangelo2020double}
Kyle Colangelo and Ying-Ying Lee.
\newblock Double debiased machine learning nonparametric inference with
  continuous treatments.
\newblock \emph{arXiv preprint arXiv:2004.03036}, 2020.

\bibitem[Cui and Tchetgen(2019)]{cui2019bias}
Yifan Cui and Eric~Tchetgen Tchetgen.
\newblock Bias-aware model selection for machine learning of doubly robust
  functionals.
\newblock \emph{arXiv preprint arXiv:1911.02029}, 2019.

\bibitem[Curth et~al.(2020)Curth, Alaa, and van~der
  Schaar]{curth2020semiparametric}
Alicia Curth, Ahmed~M Alaa, and Mihaela van~der Schaar.
\newblock Semiparametric estimation and inference on structural target
  functions using machine learning and influence functions.
\newblock \emph{arXiv preprint arXiv:2008.06461}, 2020.

\bibitem[Demirer et~al.(2019)Demirer, Syrgkanis, Lewis, and
  Chernozhukov]{demirer2019semi}
Mert Demirer, Vasilis Syrgkanis, Greg Lewis, and Victor Chernozhukov.
\newblock Semi-parametric efficient policy learning with continuous actions.
\newblock \emph{arXiv preprint arXiv:1905.10116}, 2019.

\bibitem[D{\'\i}az(2017)]{diaz2017efficient}
Iv{\'a}n D{\'\i}az.
\newblock Efficient estimation of quantiles in missing data models.
\newblock \emph{Journal of Statistical Planning and Inference}, 190:\penalty0
  39--51, 2017.

\bibitem[Fan et~al.(2012)Fan, Guo, and Hao]{fan2012variance}
Jianqing Fan, Shaojun Guo, and Ning Hao.
\newblock Variance estimation using refitted cross-validation in ultrahigh
  dimensional regression.
\newblock \emph{Journal of the Royal Statistical Society: Series B (Statistical
  Methodology)}, 74\penalty0 (1):\penalty0 37--65, 2012.

\bibitem[Fan et~al.(2020)Fan, Hsu, Lieli, and Zhang]{fan2020estimation}
Qingliang Fan, Yu-Chin Hsu, Robert~P Lieli, and Yichong Zhang.
\newblock Estimation of conditional average treatment effects with
  high-dimensional data.
\newblock \emph{Journal of Business \& Economic Statistics}, \penalty0
  (just-accepted):\penalty0 1--39, 2020.

\bibitem[Farrell(2015)]{farrell2015robust}
Max~H Farrell.
\newblock Robust inference on average treatment effects with possibly more
  covariates than observations.
\newblock \emph{Journal of Econometrics}, 189\penalty0 (1):\penalty0 1--23,
  2015.

\bibitem[Farrell et~al.(2018)Farrell, Liang, and Misra]{farrell2018deep}
Max~H Farrell, Tengyuan Liang, and Sanjog Misra.
\newblock Deep neural networks for estimation and inference.
\newblock \emph{arXiv preprint arXiv:1809.09953}, 2018.

\bibitem[Firpo(2007)]{FirpoSergio2007ESEo}
Sergio Firpo.
\newblock Efficient semiparametric estimation of quantile treatment effects.
\newblock \emph{Econometrica}, 75:\penalty0 259--276, 2007.

\bibitem[Foster and Syrgkanis(2019)]{foster2019orthogonal}
Dylan~J Foster and Vasilis Syrgkanis.
\newblock Orthogonal statistical learning.
\newblock \emph{arXiv preprint arXiv:1901.09036}, 2019.

\bibitem[{Frölich} and {Melly}(2013)]{frolich2007unconditional}
Markus {Frölich} and Blaise~Stéphane {Melly}.
\newblock Unconditional quantile treatment effects under endogeneity.
\newblock \emph{Journal of Business \& Economic Statistics}, 31\penalty0
  (3):\penalty0 346--357, 2013.

\bibitem[{Ibragimov} and {Hasminskii}(1981)]{ibragimov1981statistical}
I.~A. {Ibragimov} and R.~Z. {Hasminskii}.
\newblock Statistical estimation : asymptotic theory.
\newblock 1981.

\bibitem[Imbens and Angrist(1994)]{imbens1994identification}
Guido~W Imbens and Joshua~D Angrist.
\newblock Identification and estimation of local average treatment effects.
\newblock \emph{Econometrica}, pages 467--475, 1994.

\bibitem[Javanmard and Montanari(2014)]{javanmard2014confidence}
Adel Javanmard and Andrea Montanari.
\newblock Confidence intervals and hypothesis testing for high-dimensional
  regression.
\newblock \emph{The Journal of Machine Learning Research}, 15\penalty0
  (1):\penalty0 2869--2909, 2014.

\bibitem[Kallus and Mao(2020)]{kallus2020role}
Nathan Kallus and Xiaojie Mao.
\newblock On the role of surrogates in the efficient estimation of treatment
  effects with limited outcome data.
\newblock \emph{arXiv preprint arXiv:2003.12408}, 2020.

\bibitem[Kallus and Uehara(2020)]{kallus2020double}
Nathan Kallus and Masatoshi Uehara.
\newblock Double reinforcement learning for efficient off-policy evaluation in
  markov decision processes.
\newblock \emph{Journal of Machine Learning Research}, 21\penalty0
  (167):\penalty0 1--63, 2020.

\bibitem[Kallus et~al.(2019)Kallus, Mao, and Zhou]{kallus2019assessing}
Nathan Kallus, Xiaojie Mao, and Angela Zhou.
\newblock Assessing algorithmic fairness with unobserved protected class using
  data combination.
\newblock \emph{arXiv preprint arXiv:1906.00285}, 2019.

\bibitem[{Kasy}(2019)]{kasy2019uniformity}
Maximilian {Kasy}.
\newblock Uniformity and the delta method.
\newblock \emph{Journal of Econometric Methods}, 8\penalty0 (1):\penalty0
  1--19, 2019.

\bibitem[Kennedy(2020)]{kennedy2020optimal}
Edward~H Kennedy.
\newblock Optimal doubly robust estimation of heterogeneous causal effects.
\newblock \emph{arXiv preprint arXiv:2004.14497}, 2020.

\bibitem[Klaassen(1987)]{klaassen1987consistent}
Chris~AJ Klaassen.
\newblock Consistent estimation of the influence function of locally
  asymptotically linear estimators.
\newblock \emph{The Annals of Statistics}, pages 1548--1562, 1987.

\bibitem[{Levit}(1976)]{levit1976on}
B.~Ya. {Levit}.
\newblock On the efficiency of a class of non-parametric estimates.
\newblock \emph{Theory of Probability and Its Applications}, 20\penalty0
  (4):\penalty0 723--740, 1976.

\bibitem[{Linton}(1996)]{linton1996edgeworth}
Oliver~B. {Linton}.
\newblock Edgeworth approximation for minpin estimators in semiparametric
  regression models.
\newblock \emph{Econometric Theory}, 12\penalty0 (1):\penalty0 30--60, 1996.

\bibitem[Lu et~al.(2019)Lu, Nie, and Wager]{lu2019robust}
Chen Lu, Xinkun Nie, and Stefan Wager.
\newblock Robust nonparametric difference-in-differences estimation.
\newblock \emph{arXiv preprint arXiv:1905.11622}, 2019.

\bibitem[Meinshausen(2006)]{meinshausen2006quantile}
Nicolai Meinshausen.
\newblock Quantile regression forests.
\newblock \emph{Journal of Machine Learning Research}, 7\penalty0
  (Jun):\penalty0 983--999, 2006.

\bibitem[{Newey}(1990)]{newey1990semiparametric}
Whitney~K. {Newey}.
\newblock Semiparametric efficiency bounds.
\newblock \emph{Journal of Applied Econometrics}, 5\penalty0 (2):\penalty0
  99--135, 1990.

\bibitem[{Newey}(1994)]{newey1994the}
Whitney~K. {Newey}.
\newblock The asymptotic variance of semiparametric estimators.
\newblock \emph{Econometrica}, 62\penalty0 (6):\penalty0 1349--1382, 1994.

\bibitem[Newey and Powell(1987)]{newey1987asymmetric}
Whitney~K Newey and James~L Powell.
\newblock Asymmetric least squares estimation and testing.
\newblock \emph{Econometrica: Journal of the Econometric Society}, pages
  819--847, 1987.

\bibitem[{Newey} et~al.(1998){Newey}, {Hsieh}, and
  {Robins}]{newey1998undersmoothing}
Whitney~K. {Newey}, Fushing {Hsieh}, and James {Robins}.
\newblock Undersmoothing and bias corrected functional estimation.
\newblock 1998.

\bibitem[Neyman(1959)]{Neyman1959}
Jerzy Neyman.
\newblock Optimal asymptotic tests of composite statistical hypotheses.
\newblock \emph{Probability and Statistics}, pages 416--44, 1959.

\bibitem[Nie and Wager(2017)]{nie2017quasi}
Xinkun Nie and Stefan Wager.
\newblock Quasi-oracle estimation of heterogeneous treatment effects.
\newblock \emph{arXiv preprint arXiv:1712.04912}, 2017.

\bibitem[Ning et~al.(2017)Ning, Liu, et~al.]{ning2017general}
Yang Ning, Han Liu, et~al.
\newblock A general theory of hypothesis tests and confidence regions for
  sparse high dimensional models.
\newblock \emph{The Annals of Statistics}, 45\penalty0 (1):\penalty0 158--195,
  2017.

\bibitem[Oprescu et~al.(2019)Oprescu, Syrgkanis, and Wu]{oprescu2019orthogonal}
Miruna Oprescu, Vasilis Syrgkanis, and Zhiwei~Steven Wu.
\newblock Orthogonal random forest for causal inference.
\newblock In \emph{International Conference on Machine Learning}, pages
  4932--4941. PMLR, 2019.

\bibitem[Ota et~al.(2019)Ota, Kato, and Hara]{ota2019quantile}
Hirofumi Ota, Kengo Kato, and Satoshi Hara.
\newblock Quantile regression approach to conditional mode estimation.
\newblock \emph{Electronic Journal of Statistics}, 13\penalty0 (2):\penalty0
  3120--3160, 2019.

\bibitem[Robins et~al.(2008)Robins, Li, Tchetgen, van~der Vaart,
  et~al.]{robins2008higher}
James Robins, Lingling Li, Eric Tchetgen, Aad van~der Vaart, et~al.
\newblock Higher order influence functions and minimax estimation of nonlinear
  functionals.
\newblock In \emph{Probability and statistics: essays in honor of David A.
  Freedman}, pages 335--421. Institute of Mathematical Statistics, 2008.

\bibitem[{Robins} and {Rotnitzky}(1995)]{robins1995semiparametric}
James~M. {Robins} and Andrea {Rotnitzky}.
\newblock Semiparametric efficiency in multivariate regression models with
  missing data.
\newblock \emph{Journal of the American Statistical Association}, 90\penalty0
  (429):\penalty0 122--129, 1995.

\bibitem[Robins et~al.(1994)Robins, Rotnitzky, and Zhao]{robins1994estimation}
James~M Robins, Andrea Rotnitzky, and Lue~Ping Zhao.
\newblock Estimation of regression coefficients when some regressors are not
  always observed.
\newblock \emph{Journal of the American statistical Association}, 89\penalty0
  (427):\penalty0 846--866, 1994.

\bibitem[Robins et~al.(2013)Robins, Zhang, Ayyagari, Logan, Tchetgen, Li,
  Lumley, van~der Vaart, Committee, et~al.]{robins2013new}
James~M Robins, Peng Zhang, Rajeev Ayyagari, Roger Logan, Eric~Tchetgen
  Tchetgen, Lingling Li, Thomas Lumley, Aad van~der Vaart, HEI Health~Review
  Committee, et~al.
\newblock New statistical approaches to semiparametric regression with
  application to air pollution research.
\newblock \emph{Research report (Health Effects Institute)}, \penalty0
  (175):\penalty0 3, 2013.

\bibitem[Robins et~al.(2017)Robins, Li, Mukherjee, Tchetgen, van~der Vaart,
  et~al.]{robins2017minimax}
James~M Robins, Lingling Li, Rajarshi Mukherjee, Eric~Tchetgen Tchetgen, Aad
  van~der Vaart, et~al.
\newblock Minimax estimation of a functional on a structured high-dimensional
  model.
\newblock \emph{The Annals of Statistics}, 45\penalty0 (5):\penalty0
  1951--1987, 2017.

\bibitem[{Robinson}(1988)]{robinson1988root}
Peter~M {Robinson}.
\newblock Root-n-consistent semiparametric regression.
\newblock \emph{Econometrica}, 56\penalty0 (4):\penalty0 931--954, 1988.

\bibitem[Rockafellar and Uryasev(2002)]{rockafellar2002conditional}
R~Tyrrell Rockafellar and Stanislav Uryasev.
\newblock Conditional value-at-risk for general loss distributions.
\newblock \emph{Journal of banking \& finance}, 26\penalty0 (7):\penalty0
  1443--1471, 2002.

\bibitem[Schick(1986)]{schick1986asymptotically}
Anton Schick.
\newblock On asymptotically efficient estimation in semiparametric models.
\newblock \emph{The Annals of Statistics}, pages 1139--1151, 1986.

\bibitem[Semenova(2017)]{semenova2017machine}
Vira Semenova.
\newblock Machine learning for set-identified linear models.
\newblock \emph{arXiv preprint arXiv:1712.10024}, 2017.

\bibitem[Semenova and Chernozhukov(2020)]{semenova2020debiased}
Vira Semenova and Victor Chernozhukov.
\newblock Debiased machine learning of conditional average treatment effects
  and other causal functions.
\newblock \emph{The Econometrics Journal}, 2020.

\bibitem[Singh and Sun(2019)]{singh2019biased}
Rahul Singh and Liyang Sun.
\newblock De-biased machine learning for compliers.
\newblock \emph{arXiv preprint arXiv:1909.05244}, 2019.

\bibitem[Smucler et~al.(2019)Smucler, Rotnitzky, and
  Robins]{smucler2019unifying}
Ezequiel Smucler, Andrea Rotnitzky, and James~M Robins.
\newblock A unifying approach for doubly-robust $\ell_1$ regularized estimation
  of causal contrasts.
\newblock \emph{arXiv preprint arXiv:1904.03737}, 2019.

\bibitem[Syrgkanis et~al.(2019)Syrgkanis, Lei, Oprescu, Hei, Battocchi, and
  Lewis]{syrgkanis2019machine}
Vasilis Syrgkanis, Victor Lei, Miruna Oprescu, Maggie Hei, Keith Battocchi, and
  Greg Lewis.
\newblock Machine learning estimation of heterogeneous treatment effects with
  instruments.
\newblock In \emph{Advances in Neural Information Processing Systems}, pages
  15193--15202, 2019.

\bibitem[Tsiatis(2006)]{tsi}
Anastasios. Tsiatis.
\newblock \emph{Semiparametric Theory and Missing Data}.
\newblock Springer, New York, 2006.

\bibitem[Van~de Geer et~al.(2014)Van~de Geer, B{\"u}hlmann, Ritov, Dezeure,
  et~al.]{van2014asymptotically}
Sara Van~de Geer, Peter B{\"u}hlmann, Ya’acov Ritov, Ruben Dezeure, et~al.
\newblock On asymptotically optimal confidence regions and tests for
  high-dimensional models.
\newblock \emph{The Annals of Statistics}, 42\penalty0 (3):\penalty0
  1166--1202, 2014.

\bibitem[van~der {Laan} and {Rose}(2011)]{laan2011targeted}
Mark~J. van~der {Laan} and Sherri {Rose}.
\newblock \emph{Targeted Learning: Causal Inference for Observational and
  Experimental Data}.
\newblock 2011.

\bibitem[Van~der Laan and Rose(2018)]{van2018targeted}
Mark~J Van~der Laan and Sherri Rose.
\newblock \emph{Targeted learning in data science}.
\newblock Springer, 2018.

\bibitem[van~der Vaart(1998)]{vaart_1998}
A.~W. van~der Vaart.
\newblock \emph{Asymptotic Statistics}.
\newblock Cambridge Series in Statistical and Probabilistic Mathematics.
  Cambridge University Press, 1998.
\newblock \doi{10.1017/CBO9780511802256}.

\bibitem[{var der Vaart}(1991)]{vaart1991on}
Aad {var der Vaart}.
\newblock On differentiable functionals.
\newblock \emph{Annals of Statistics}, 19\penalty0 (1):\penalty0 178--204,
  1991.

\bibitem[Yadlowsky et~al.(2018)Yadlowsky, Namkoong, Basu, Duchi, and
  Tian]{yadlowsky2018bounds}
Steve Yadlowsky, Hongseok Namkoong, Sanjay Basu, John Duchi, and Lu~Tian.
\newblock Bounds on the conditional and average treatment effect with
  unobserved confounding factors.
\newblock \emph{arXiv preprint arXiv:1808.09521}, 2018.

\bibitem[Zhang and Zhang(2014)]{zhang2014confidence}
Cun-Hui Zhang and Stephanie~S Zhang.
\newblock Confidence intervals for low dimensional parameters in high
  dimensional linear models.
\newblock \emph{Journal of the Royal Statistical Society: Series B: Statistical
  Methodology}, pages 217--242, 2014.

\bibitem[Zheng and van~der Laan(2011)]{zheng2011cross}
Wenjing Zheng and Mark~J van~der Laan.
\newblock Cross-validated targeted minimum-loss-based estimation.
\newblock In \emph{Targeted Learning}, pages 459--474. Springer, 2011.

\bibitem[Zhou et~al.(2018)Zhou, Athey, and Wager]{zhou2018offline}
Zhengyuan Zhou, Susan Athey, and Stefan Wager.
\newblock Offline multi-action policy learning: Generalization and
  optimization.
\newblock \emph{arXiv preprint arXiv:1810.04778}, 2018.

\bibitem[Zimmert(2018)]{zimmert2018efficient}
Michael Zimmert.
\newblock Efficient difference-in-differences estimation with high-dimensional
  common trend confounding.
\newblock \emph{arXiv preprint arXiv:1809.01643}, 2018.

\end{thebibliography}

\newpage
\appendix

\section{LDML Estimates for Local Estimating Equations using Instrumental Variable}\label{sec: IV}
In \cref{sec: estimating equation example}, we mention that without the ignorability assumption, we can rely on an instrumental variable to identify \emph{local} parameters, namely, solutions $\tth = \prns{\tth_1, \tth_2}$ to the following \emph{local} estimating equation: 
\begin{equation}\label{eq: local-est}
\overline\hP[U(Y(1);\theta_1)+V(\theta_2)\mid T(1)> T(0)]=0.
\end{equation}
We assume standard instrumental variable identification conditions: for potential treatments $T(w)$ and potential outcomes $Y(t,w)$, we have exclusion $Y(t):=Y(t,w)=Y(t,1-w)$, exogeneity $(Y(t), T(w)) \perp W \mid X$, overlap $\hP(W=1\mid X)\in(0,1)$, relevance $\overline\hP(T(1)=1)>\overline\hP(T(0)=1)$, and monotonicity $T(1)\geq T(0)$. 
Following \citet{belloni2017program},
a Neyman orthogonal estimating equation for $\tth$ is given by
\begin{align}\label{eq: IV-equation}
\psi(Z;\theta, \theta^{\op{aux}}_2,\eta_1(Z;\theta_1),\eta_2(Z))
 = 
 \begin{bmatrix}
 \psi_1(Z;\theta,\eta_1(Z;\theta_1),\eta_2(Z)) \\
  \psi_2(Z; \theta^{\op{aux}}_2,\eta_2(Z))
 \end{bmatrix},
\end{align}
where 
\begin{align*}
&\psi_1(Z;\theta,\eta_1(Z;\theta_1),\eta_2(Z))=
\biggl(
\eta_{1,1}(Z;\theta_1)-\eta_{1,2}(Z;\theta_1)
+
\frac{W}{\eta_{2, 1}(Z)}\prns{TU(Y;\theta_1)-\eta_{1,1}(Z;\theta_1)}\nonumber
\\
&\phantom{\psi(Z;\theta_1,\eta_1(Z;\theta_1),\eta_2(Z))=\biggl(}-
\frac{1-W}{1-\eta_{2, 1}(Z)}\prns{TU(Y;\theta_1)-\eta_{1,2}(Z;\theta_1)}
\biggr)\times\frac{1}{\theta^{\op{aux}}_2}
+ V\prns{\theta_2},\\\nonumber
&\psi_2(Z;\theta^{\op{aux}}_2,\eta_2(Z))
  = \eta_{2, 2}\prns{Z} - \eta_{2, 3}\prns{Z} + \frac{W}{\eta_{2, 1}\prns{Z}}\prns{T - \eta_{2, 2}\prns{Z}} - \frac{1-W}{1-\eta_{2,1}\prns{Z}}\prns{T - \eta_{2, 3}\prns{Z}} - \theta^{\op{aux}}_2.
\end{align*}
with nuisance functions 
\begin{align}\label{eq: nuisance2}
&\eta^*_1(Z;\theta_1)
=\begin{bmatrix} 
\Eb{TU(Y; \theta_1)\mid X, W=1}\\
\Eb{TU(Y; \theta_1) \mid X, W=0}
\end{bmatrix}, ~~~ \eta_2^*(Z)=
\begin{bmatrix}
\Prb{W=1\mid X} \\
\hP\prns{T=1\mid X,W=1} \\
\hP\prns{T=1\mid X,W=0}
\end{bmatrix}.
\end{align}
Here the second estimating equation $\Eb{\psi_2(Z;\theta^{\op{aux}}_2,\eta_2(Z))} = 0$ identifies the  compliance probability, denoted by the following auxiliary parameter $\theta^{\op{aux}*}_2$: 
\begin{align*}
\theta^{\op{aux}*}_2 = \Eb{\Prb{T = 1 \mid X, W = 1} - \Prb{T = 1 \mid X, W = 0}} = \Prb{T\prns{1} > T\prns{0}}.
\end{align*}
By redefining $\tilde\theta_1 = \theta_1$, $\tilde\theta_2 = \prns{\theta_2, \theta^{\op{aux}}_2}$, and $\tilde \theta = \prns{\tilde\theta_1, \tilde\theta_2}$, the estimating equation becomes 
\begin{align}\label{eq: overall-eq}
\hP\bracks{\psi(Z;\tilde\theta,\eta_1(Z;\tilde\theta_1),\eta_2(Z))} = \mb{0},
\end{align}
which apparently fits into our general framework in 
\cref{eq: est-equation}. Therefore, we can directly apply our LDML algorithm in \cref{sec: LDML-algorithm} to estimate the local parameters $\tth = \prns{\tth_1, \tth_2}$. We can also use the theory in \cref{sec: general,sec: inference} to analyze the asymptotic distribution of the resulting estimators and estimate their asymptotic variances.

\subsection{Estimating Local Quantiles}
In particular, we take the local quantile estimation as an example, namely, the solution $\tth_1$ to the local estimating equation in \cref{eq: local-est} with 
\begin{align}\label{eq: local-quantile-equation}
U\prns{Y; \theta_1} = \indic{Y \le \theta_1}, ~~ V\prns{\theta_2} = - \gamma.
\end{align}
Its orthogonal estimating equation involves the following nuisance functions:
\begin{align}\label{eq: nuisance1}
\eta^*_1(Z;\theta_1)
=\begin{bmatrix} 
\Prb{T = 1, Y \le \theta_1\mid X, W=1}\\
\Prb{T = 1, Y \le \theta_1 \mid X, W=0}
\end{bmatrix}.
\end{align}
For better readability, we denote the event of being a complier, \ie, $T(1) > T(0)$, as $\mathcal C$, the nuisance functions as $\tilde\pi^*(X) = \hP(W = 1 \mid X)$, $\nu_w^*\prns{X} = \hP\prns{T=1\mid X,W=w}$, and $\tilde\mu^*_w(X; \theta_1) = \hP\prns{T = 1, Y \le \theta_1 \mid X, W = w}$ for $w \in \braces{0, 1}$.
 We fit estimators for the  nuisance functions based on the sample-splitting scheme given in \cref{def: splitting}, which we denote as $\hat{\tilde \pi}^{(k)}(X)$,  $\hat{\nu}^{(k)}_w(X)$ and $\hat{\tilde \mu}^{(k)}(X; \hthinit) = (\hat{\tilde \mu}_{1}^{(k)}(X; \hthinit), \hat{\tilde \mu}_{0}^{(k)}(X; \hthinit))$ respectively for $k = 1, \dots, K$. 
 Finally, we obtain the estimator $\hat\theta = \prns{\hat\theta_1, \hat\theta_2^{\text{aux}}}$ by searching approximate solutions over $\Theta = \Theta_1 \times \Theta_2 \subseteq \mathbb{R} \times \mathbb{R}$ 
 to the empirical estimating equations  in \cref{def: LDML2} or \cref{def: LDML1}, specialized to \cref{eq: IV-equation,eq: local-quantile-equation}.

We next assume a strong form of the overlap and relevance assumptions and specify the convergence rates of the initial estimator and nuisance estimators. We again consider a generic treatment level $t \in \{0, 1\}$ in these two assumptions.
\begin{assumption}[Strong Overlap and Relevance Assumptions]\label{assump: IV}
Assume that there exists a positive constant $\epsilon >0$ such that for any $\hP \in \mathcal P_N$, $\epsilon \le \tilde \pi^*(X)  \le 1-\epsilon$ holds almost surely, and $\theta^{\op{aux}*}_2 \ge \epsilon$. 
\end{assumption}
\begin{assumption}[Nuisance Estimation Rates]\label{assump: nuisance-rate-iv}
Assume that 
for any $\hP \in \mathcal P_N$: with probability at least $1 - \Delta_N$, for $w = 0, 1$, 
\begin{align*}
    &\left\|\bigg\{\hP \left[\hat{\tilde \mu}_w\pk\left(X; \hthinit\pk)\right) - \tilde{\mu}^*_w\left(X; \hthinit\pk)\right)  \right]^2\bigg\}^{1/2}\right\|   \le \tilde \rho_{\mu, N}, ~ \bigg\{\hP \left[\hat{\nu}^{(k)}_w(X) - {\nu}^*_w(X)\right]^2\bigg\}^{1/2} \le \tilde\rho_{\nu, N}, \\
  &\qquad\qquad\qquad\qquad \bigg\{\hP \left[\hat{\tilde \pi}^{(k)}(X) - \tilde{\pi}^*(X)\right]^2\bigg\}^{1/2} \le \tilde\rho_{{\pi}, N}, ~~ |\hthinit\pk - \theta^*_1| \le \tilde \rho_{\theta, N},
\end{align*}
and $\epsilon \le \hat{\tilde\pi}\pk( X) \le 1- \epsilon,  0 \le \hat{\tilde \mu}_w\pk\left(X; \hthinit\pk)\right) \le 1$, $0 \le \hat{\nu}^{(k)}_w(X) \le 1$ almost surely.
\end{assumption}

In the following theorem, we derive the asymptotic distribution of the local quantile estimator, which is proved by verifying all assumptions in \cref{thm: general-split}.
\begin{proposition}[LDML for Local Quantile]\label{thm: iv}
Fix $t = 1$ and let $\Theta = \prns{\Theta_1, \Theta_2} \subseteq \R{2}$ be a compact set where $\theta_2^{\op{aux}} \ge \epsilon$ for any $\theta_2^{\op{aux}} \in \Theta_2$ and $\epsilon$ given in \cref{assump: IV}. Let $(\hat\theta_1, \hat\theta_2^{\op{aux}})$ be the LDML estimator given in either \cref{def: LDML2} or \cref{def: LDML1}, specialized to \cref{eq: IV-equation,eq: local-quantile-equation}. Suppose that there exist constants $c', C$ such that the following conditions hold for any instance $\hP \in \mathcal P_N$:
\begin{enumerate}[label=\roman*.]
\item \label{thm: iv: 0}  Conditions \ref{assump: identification: interior} (with $c_1$), \ref{assump: identification: diffable}, \ref{assump: identification: moment} (with $c_5,c_6$)  and 
  condition \ref{assump: error: approximation} of \cref{assump: error} 
 for the estimating equation in \cref{eq: IV-equation,eq: local-quantile-equation}.
\item \label{thm: iv: 1}For any $\theta_1 \in \Theta_1    $, the  distribution function of $Y(t)$ for compliers, denoted as $F_t(\theta_1 \mid \mathcal C)$, is twice continuously differentible. Its first two order derivatives $f_t(\theta_1 \mid \mathcal C)$ and $\dot{f}_t(\theta_1 \mid \mathcal C)$ satisfy that $f_t(\theta_1 \mid \mathcal C) \le c_1'$, $\abs{\dot{f}_t(\theta_1 \mid \mathcal C)} \le c'_2$ for any $\theta_1 \in \Theta_1$, and $f_t(\theta_1^* \mid \mathcal C) \ge c_3' > 0$. 
\item $2\|\hP\left[\psi(Z;\theta, \theta^{\op{aux}}_2,\eta^*_1(Z;\theta^*_1),\eta^*_2(Z))\right]\| \ge c_2$ for all $\theta = \prns{\theta_1, \theta_2^{\op{aux}}} \in\Theta$ such that $\|\theta - \theta^*\| \ge \frac{c_3}{2\sqrt{d}c_{\op{Lip}}}$ where $c_{\op{Lip}} \coloneqq \max\braces{\sqrt{\prns{\frac{c_1'}{\epsilon^2}}^2 + \prns{\frac{c'_2}{\epsilon}}^2}, \sqrt{\prns{\frac{2}{\epsilon^3}}^2 + \prns{\frac{c_1'}{\epsilon^2}}^2}}$.
\item \label{thm: iv: 2} For any $\theta_1 \in \mathcal{B}(\tth_1;\max\{\frac{4\tilde\rho_{\pi, N}}{\epsilon^2\prns{1-\epsilon}\delta_N}, \rho_{\theta, N}\}) \cap \Theta$ and $w \in \braces{0, 1}$, the conditional distribution of $Y\prns{t}$ given $X, T\prns{w} = 1$, denoted as $F_{t, w}(\theta_1 \mid X)$, is twice differentiable almost surely with first two order derivatives $f_{t, w}(\theta_1 \mid X)$ and $\dot{f}_{t, w}(\theta_1 \mid X)$ that satisfy $f_{t, w}(\theta_1 \mid X)\le C$ and $\left|\dot{f}_{t, w}(\theta_1 \mid X)\right| \le C$ almost surely.
\item\label{thm: iv: rate} The nuisance estimator convergence rates satisfy that $\tilde\rho_{\pi, N} \le \frac{\delta_N^3}{\log N}$, $\tilde\rho_{\mu, N} + C\tilde\rho_{\theta, N} \le \frac{\delta^2_N}{\log N}$, 
$\tilde\rho_{\pi, N}\prns{\tilde\rho_{\mu, N} + C\tilde\rho_{\theta, N}} \le \frac{\epsilon^4\prns{1-\epsilon}^3}{4\prns{\epsilon^3 + \prns{1-\epsilon}^3}}\delta_N N^{-1/2}$, 
$\tilde\rho_{\pi, N}{\tilde\rho_{\nu, N}} \le  \frac{\epsilon^3\prns{1-\epsilon}^3}{8\prns{\epsilon^3 + \prns{1-\epsilon}^3}}\delta_N N^{-1/2}$
 with $\delta_N$ satisfying that $\delta_N \le \frac{\epsilon^3\prns{1-\epsilon}^2}{4C + 3\epsilon^2\prns{1-\epsilon}}$,
$\frac{\delta_N}{\log N} \le \frac{1}{C_{\epsilon}}$ for a positive constant $C_{\epsilon}$ given in \cref{eq: Cepsilon}
. 
\end{enumerate}
Then $(\hat\theta_1, \hat\theta_2^{\op{aux}})$ satisfies the 
conclusion of
\cref{thm: general-split} 
for $\psi(Z;\theta^*,\eta^*_1(Z;\theta_1^*),\eta^*_2(Z))$ given in \cref{eq: IV-equation} and 
\begin{align*}
J^{*-1} = \begin{bmatrix}
\frac{1}{f_1\prns{\theta_1^* \mid \mathcal{C}}} & -\frac{\gamma}{\theta^{\op{aux}*}_2f_1\prns{\theta_1^* \mid \mathcal{C}}} \\
0 & - 1
\end{bmatrix}.
\end{align*}
In particular, the local quantile estimator $\hat\theta_1$ is asymptotically linear with the following influence function:
\begin{align*}
\frac{1}{f_1\prns{\theta_1^* \mid \mathcal{C}}}\psi_1(Z_i;\theta^*,\eta^*_1(Z_i;\theta^*_1),\eta^*_2(Z_i))-\frac{\gamma}{\theta^{\op{aux}*}_2f_1\prns{\theta_1^* \mid \mathcal{C}}}  \psi_2(Z_i; \theta^{\op{aux}*}_2,\eta^*_2(Z_i)),
\end{align*}
where $\psi_1(Z_i;\theta^*,\eta^*_1(Z_i;\theta^*_1),\eta^*_2(Z_i))$ and $\psi_2(Z_i; \theta^{\op{aux}*}_2,\eta^*_2(Z_i))$ are given in \cref{eq: IV-equation}.
Analogous conclusion for local quantiles of $Y(0)$  holds when all assumptions above hold for $t = 0$.
\end{proposition}

\subsection{Effect of 401(k) Participation on Net Financial Assets}
\begin{table}[b!]
\caption{The LQTE of 401(k) participation in thousand dollars (and standard error) estimated by LDML using different regression methods, and raw unadjusted differences of marginal quantiles by eligibility.}\label{table: 401k lqte}
\centering
{\footnotesize\spacingset{1}\begin{tabular}{llllllllll}
  \toprule
$\gamma$ & $K$ & LASSO & Neural Net & Boosting & Forest & Raw \\ 
  \midrule
 \multirow{3}{*}{25\%} & 5 & 1.75 (0.23) & 2.06 (0.25) & 1.57 (0.26) & 1.91 (0.44) & \multirow{3}{*}{4.18 (0.37)} \\ 
    & 15 & 1.74 (0.23) & 2.04 (0.25) & 1.57 (0.26) & 1.88 (0.44) \\ 
    & 25 & 1.75 (0.23) & 2.07 (0.25) & 1.58 (0.26) & 1.87 (0.44) \\ 
   \multirow{3}{*}{50\%} & 5 & 8.64 (0.60) & 10.38 (0.66) & 7.54 (0.60) & 6.32 (1.12) & \multirow{3}{*}{15.05 (0.67)} \\ 
    & 15 & 8.55 (0.59) & 10.64 (0.68) & 7.53 (0.60) & 6.12 (1.11) \\ 
    & 25 & 8.52 (0.59) & 10.45 (0.67) & 7.51 (0.60) & 6.08 (1.11) \\ 
   \multirow{3}{*}{75\%} & 5 & 22.02 (1.87) & 31.86 (1.77) & 20.54 (2.05) & 19.28 (4.81) & \multirow{3}{*}{38.59 (1.71)} \\ 
    & 15 & 21.78 (1.86) & 32.73 (1.73) & 20.48 (2.05) & 19.91 (5.07) \\ 
    & 25 & 21.72 (1.89) & 33.01 (1.76) & 20.45 (2.04) & 19.96 (5.24) \\ 
   \bottomrule
\end{tabular}}
\end{table}
\begin{figure}[t!]\centering%
\includegraphics[width=0.9\textwidth]{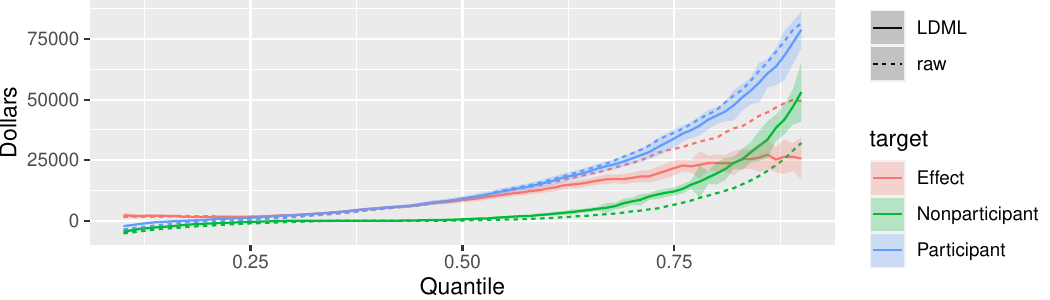}
\caption{LDML estimates of a range of local quantiles and LQTEs with confidence 90\% intervals and comparison to raw unadjusted marginal quantiles by treatment group.%
}
\label{fig: 401k lqte}
\bigskip\end{figure}

Next, we estimate the effect of 401(k) participation on net assets.
Participation in a 401(k) plan (here considered the treatment, $T$) is not randomly assigned: individuals with a preference for saving may save more in non-retirement accounts than others whether they were to participate in retirement savings or not. There may be many other confounding factors, such as the possibility of higher financial acumen of savers leading to higher net worth otherwise.
It is unlikely that we can control for all these factors using observable covariates. Instead, we rely on instrumenting on eligibility since, as argued in \cref{sec: 401k eligibility}, eligibility is ignorable given covariates. Additionally, one cannot participate if one is ineligible, ensuring monotonicity, and some eligible individuals do participate, ensuring relevance. Assuming that eligibility cannot affect net assets except through its effect on participation, we have that eligibility for a 401(k) (here considered as $W$) is valid IV. We can therefore use it to estimate local quantiles by and LQTEs of 401(k) Participation on the population of individuals that would participate if eligible.

We use LDML applied to the Neyman orthogonal estimating equation \cref{eq: IV-equation,eq: local-quantile-equation}. Again, we consider the impact of different choices in the application of LDML. We repeat the same specification as above, using each possible nuisance estimator to fit the conditional probabilities \cref{eq: nuisance1,eq: nuisance2}. 
We display the results for the 25\%, 50\%, and 75\% quantiles while varying $K$ and the nuisance estimators in \cref{table: 401k lqte}. The qualitative results regarding the stability of LDML across methods and $K$ remain the same. Then, focusing as before on nuisance estimation using LASSO and on $K=15$, we also estimate a range of local quantiles and QTEs, which we plot along with 90\% confidence intervals in \cref{fig: 401k qte}. Again, we compare to the raw unadjusted marginal quantiles within each treatment group.

\section{LDML Estimates for Expectiles}\label{sec: expectiles}
We can also apply our method and analysis to estimating the $\gamma$-expectile $\theta_1$ of $Y(1)$, as defined in \cref{eq: expectile-complete}. Instantiating \cref{eq: est-eq-causal} for expectiles and rearranging, we get the following efficient estimating function from incomplete data:
\begin{align}
&\psi(Z; \theta_1, \eta_1^*(Z; \theta_1), \eta_{2}^*(Z)) \nonumber\\     
  =& \frac{\ind(T = 1)}{\eta^*_{2, 2}(Z)}\left[(1 - \gamma)\left(Y - \eta_{2, 1}^*(Z)\right) - (1 - 2\gamma)\left(\max\left(Y - \theta_1, 0\right) - \eta_{1}^*(Z; \theta_1)\right)\right] \nonumber \\
  +&\left[(1 - \gamma)\eta_{2, 1}^*(Z) - (1 - 2\gamma)\eta_{1}^*(Z; \theta_1)\right], \label{eq: est-eq-expectile}\\
  \text{where}&\qquad \eta_1^*(Z; \theta_1)=\Eb{\max(Y - \theta_1,0) \mid X, T=1},  \nonumber\\
\phantom{\text{where}}&\qquad \eta^*_2(Z;\theta_1)
=\begin{bmatrix} 
\expect\left[Y\mid X,T=1\right]\\
\Prb{T=1\mid X}
\end{bmatrix}. \nonumber
\end{align} 
The next result gives the asymptotic behavior of LDML applied to these equations.

\begin{proposition}\label{thm: expectile}
Fix $t = 1$ and let the estimator $\hat \theta_1$ be given by applying either \cref{def: LDML2} or \cref{def: LDML1} to the estimating function in \cref{eq: est-eq-expectile}.
Suppose \cref{assump: unconf,assump: nuisance-rate} hold and there exist positive constants $C$, $c_1', c_2'$, such that for any $\hP\in \mathcal P_N$, the following conditions hold: 
\begin{enumerate}[label=\roman*.]
\item Conditions \ref{assump: identification: interior} (with $c_1$), \ref{assump: identification: diffable}, \ref{assump: identification: moment} (with $c_5,c_6$)
 of \cref{assump: identification},
  condition \ref{assump: error: approximation} of \cref{assump: error}, and condition \ref{assump: causal: rate} of \cref{thm: causal} for the estimating function in \cref{eq: est-eq-expectile}  and the corresponding nuisance estimators.
\item $F_t(\theta_1)$ is continuous at $\theta_1^*$, and $|-(1 - 2\gamma)F_t(\theta_1^*) - \gamma| \ge c_1' > 0$. Moreover, for any $\theta \in \Theta$ such that $\|\theta - \theta^*\| \ge \frac{c_1'}{2}$,  $2\abs{\hP\left[U(Y(t); \theta_1)\right]} \ge c_2'$ for $U(Y(t); \theta_1)$ given in \cref{eq: expectile-complete}.
\item At any $\theta_1 \in \mathcal{B}(\tth;\max\{\frac{4C\sqrt{d}\rho_{\pi, N}}{\delta_N\varepsilon_\pi}, \rho_{\theta, N}\}) \cap \Theta_1$ , $F_t(\theta_1 \mid X)$ is almost surely differentiable with first-order derivative $f_t(\theta_1 \mid X)$, and second-order derivative $\dot{f}_{t}(\theta_1 \mid X)$ that satisfies  ${{f}_{t}(\theta_1 \mid X)} \le C$ and $\abs{\dot{f}_{t}(\theta_1 \mid X)} \le C$ almost surely;
\item For any $\theta_1 \in \Theta_1$,
\begin{align*}
\left\{\hP\left[\expect[\max\{Y(t) - \theta_1, 0\}\mid X]\right]^2\right\}^{1/2} \le C, ~~~ \left\{\hP\left[\expect\left[Y(t) \mid X\right]\right]^2\right\}^{1/2} \le C.
\end{align*} 
\end{enumerate}
Then $\hat\theta_1$ satisfies the 
conclusion of
\cref{thm: general-split} 
for $\psi(Z; \tth_1, \tnua(Z; \tth_1), \tnub(Z))$ given in \cref{eq: est-eq-expectile} and $J^* = -\gamma - (1 - 2\gamma)F_t(\theta^*_1)$. Analogous conclusion for expectile of $Y(0)$  holds when all assumptions above hold for $t = 0$.
\end{proposition}
When constructing confidence intervals, we only need to estimate $F_t(\theta_1^*)$ to estimate  $J^*$. This can be easily estimated by the inverse propensity reweighted estimator $$\frac{1}{N}\sum_{k}\sum_{i \in \Dcal_k}\frac{\indic{T_i = t}}{\hat\pi\pk(t \mid X_i)}\indic{Y \le \hat\theta_1}.$$
Alternatively, it can be estimated by an imputation estimator based on $\hat\mu\pk$ or a LDML estimator that uses both $\hat\pi\pk$ and $\hat\mu\pk$ (see \cref{rem:variancee_quantile}). 

\section{Theoretical Analysis of IPW Initial Estimator}\label{sec: ipw initial}
In this part, we show that the IPW initial estimator given in \cref{def: IPW_init} can satisfy the conditions on $\hthinit$ in \cref{assump: nuisance-rate}.
\begin{proposition}[IPW Initial Estimator Rate]\label{thm: IPW}
Fix $t= 1$ and 
let the initial estimator $\hthinit\pk$ 
be constructed according to \cref{def: IPW_init} for $k = 1, \dots, K$. Assume the following (for $t=1$):
\begin{enumerate}[label=\roman*.]
\item\label{thm: IPW: rate} For each $k \in \{1, \dots, K\}$ and $l \in \mathcal H_{k, 1}$, $\hat\pi^{(k, l)}$ satisfies the same conditions as for $\hat\pi\pk$ in \cref{assump: nuisance-rate}.
\item Conditions \ref{assump: causal: Jacobian}, \ref{assump: causal: identification}, and \ref{assump: causal: boundedness} in \cref{thm: causal} (with constants $c_2$ to $c_4$ and $C$) hold.
\item There exists a nuisance realization set $\Pi_N$ that contains the true propensity score $\pi^*$ and also the propensity score estimators $\hat\pi^{(k, l)}$ for $k = 1, \dots, K$ and $l \in \mathcal{H}_{k, 1}$ with at least probability $1 - \Delta_N$. Moreover, any $\pi \in \Pi_N$ satisfies that $\pi(t \mid X)\ge \epsilon_\pi$.
\item For each $\pi \in \Pi_N$, the function class $\mathcal G_\pi = \{(X,T,Y)\mapsto\frac{\indic{T = t}}{\pi(t \mid X)}U_j(Y; \theta_1) + V_j(\theta_2): j = 1, \dots, d, \theta \in \Theta\}$ is suitably measurable and its uniform covering entropy satisfies the following condition: for positive constants $a', v'$ and $q' > 2$, 
$
        \sup_{\mathbb Q}\log N(\epsilon\|G_{\pi}\|_{\mathbb Q, 2}, \mathcal G_{\pi}, \|\cdot\|_{\mathbb Q, 2}) \le v'\log(a'\epsilon)$ $\forall \epsilon \in(0, 1]
 $,
 where $G_{\pi}$ is a measurable envelope for $\mathcal G_\pi$. 
There exists a positive constant $c_8$ such that for any $\hP \in \mathcal P_N$, $\|G_{\pi}\|_{\hP, q'} \le c_8$. 
\item $\prns{\frac{K'}{N}}^{1/2}\log\prns{\frac{K'}{N}} +  \prns{\frac{K'}{N}}^{1 - \frac{1}{q'}}\log\prns{\frac{K'}{N}} \le \delta_N\rho_{\pi, N}$;
\end{enumerate}
Then there exists a constant $c$ that only depends on pre-specified constants in the conditions above such that with probability $1 - c\prns{\log N}^{-1}$,
$
    \rho_{\theta, N} \le 2c^{-1}_3\prns{C\sqrt{d}\epsilon_\pi^{-1} + 1 }\rho_{\pi, N}.
$
\end{proposition}
In \cref{remark: IPW-init}, we discuss the corresponding rate conditions on other nuisance estimators when using the IPW initial estimator, based on the conclusion in \cref{thm: IPW}.

\section{An Alternative LDML Estimator}\label{sec: alternative-est-eq}
In \cref{def: LDML2}, we construct an LDML estimator by first averaging estimates of the  equation in \cref{eq: part-equation} over all folds and then solving the grand-average equation approximately. 
Below we provide an alternative LDML estimator that first solves the estimate of \cref{eq: part-equation} from each fold separately and then averages these  solutions. 
\begin{definition}[LDML2]
\label{def: LDML1}
For $k = 1, \dots, K$, 
construct $\hth\pk$ by (approximately) solving
\begin{equation}
    \overline\Psi^{(k)}(\theta)=\frac{1}{\abs{\Dcal_{k}}} \sum_{i \in \Dcal_{k}}\psi(Z_i; \theta, \hnua\pk(Z_i; \hthinit\pk), \hnub\pk(Z_i)) = 0.
\end{equation}
In fact, we allow for an approximate least-squares solution, 
 which is useful if the empirical estimating equation has no exact solution. Namely, we let $\hth^{k}$ be any satisfying
\begin{align}\label{eq: approx-LDML1}
    &\textstyle\|
    \overline\Psi^{(k)}(\hat\theta\pk)
    \| 
    \le \inf_{\theta \in \Theta}\|
    \overline\Psi^{(k)}(\theta)
    \| + \varepsilon_N. 
\end{align}
Then, we let the final estimator be 
\begin{align}
    \hth = \frac{1}{K}\sum_{k = 1}^K\hth\pk.
\end{align}
\end{definition}
We can easily follow previous proofs for the LDML estimator in \cref{def: LDML2} to show that \cref{thm: general-split,thm:cf,thm: causal,thm: quantile only} also apply to $\hat\theta$ in \cref{def: LDML1}, provided that $\epsilon_N$ in \cref{eq: approx-LDML1} is $o\prns{N^{-1/2}}$ (\ie, condition \ref{assump: error: approximation} in \cref{assump: error}). For example, we demonstrate this at the end of the proof for \cref{thm: general-split}.
Thus the two LDML estimators in \cref{def: LDML2} and \cref{def: LDML1} are asymptotically equivalent.

\section{Practical Considerations}\label{sec: practical}
The proposed LDML estimator $\hat\theta$ in \cref{def: LDML2} or \cref{def: LDML1} relies on nuisance estimates based on random sample splitting (\cref{def: splitting}). 
Although the uncertainty due to sample splitting does not affect the asymptotic theory, it may influence the finite-sample performance of the LDML estimator.

To make the results more robust to sample splitting, we may consider aggregating the estimates over different random splitting realizations. 
In particular, it is possible to use many other different ways of splitting data. For example, in both \cref{def: LDML1,def: LDML2} we may average more than just $K$ solutions or equations. For each $k$, we can permute over all $\binom{K-1}{K'}$ splits of $\{1,\dots,K\}\setminus\{k\}$ into $K'$ and $K-1-K'$ folds used for fitting $\hthinit\pk$ and $\hnua\pk(\cdot; \hthinit\pk),\hnub\pk$. Or, we could even permute over all $\sum_{K'=1}^{K-2}\binom{K-1}{K'}$ ways to split $\{1,\dots,K\}\setminus\{k\}$ into two. Or, we can even repeat the initial random splitting into $K$ folds many times over and average the resulting estimates from either \cref{def: LDML1} or \ref{def: LDML2}, or take their median to avoid outliers, or solve the grand-mean of estimating equations. All of these procedures can provide improved finite-sample performance in practice as they can only reduce variance without affecting bias, and we do recommend these, but they have no effect on the leading asymptotic behavior, which remains the same whether you use one or more splits of the data into folds and/or one or more splits of $\{1,\dots,K\}\setminus\{k\}$ into two.

With estimates from multiple random splitting realizations, we may also improve variance estimation and to account for the variance due to random splitting. In particular, letting $\hat\theta_s,\,\widehat\Sigma_s$ be the parameter and variance estimates for each run of LDML for $s=1,\dots,S$, we can let $\hat\theta^\text{mean}=\frac1S\sum_{s=1}^S\hat\theta_s$ and $\widehat\Sigma^\text{mean}=\frac1S\sum_{s=1}^S(\widehat\Sigma_s+\frac1S(\hat\theta_s-\hat\theta^\text{mean})(\hat\theta_s-\hat\theta^\text{mean})^\top)$  be the final parameter and variance estimates.
Like $\hat\theta^\text{mean}$, the first term in $\widehat\Sigma^\text{mean}$ reduces the variance in the estimate $\widehat\Sigma_s$ itself. The second term in $\widehat\Sigma^\text{mean}$ accounts for the variance of $\hat\theta^\text{mean}$ due to random splitting.
Notice that the second term vanishes as $S\to\infty$; indeed then $\hat\theta^\text{mean}$ has no variance due to random splitting as it is fully averaged over. Because $\hat\theta_s$ are each consistent, the second term also vanishes as $N\to\infty$. Removing the $\frac1S$ factor in the second term we can instead get an estimate of the variance of each single $\hat\theta_s$, rather than of $\hat\theta^\text{mean}$, accounting for random splitting.
This procedure extends a similar proposal by \cite{ChernozhukovVictor2018Dmlf} for inference in linear estimating equations.

\section{Invariant Jacobian Matrix}\label{sec: jacobian}
The key condition that motivates our LDML approach is the invariant Jacobian condition in \cref{assump: jacob}. Below \cref{assump: jacob}, we directly show that this condition is satisfied for efficient estimating equations in incomlete data settings. 
In the following proposation, we provide a more general sufficient condition for the invariant Jacobian condition.

\begin{proposition}[Sufficient Conditions for Invariant Jacobian]\label{prop: frechet-diff}
Assume that the map $(\theta,  \nua(\cdot; \theta_1')) \mapsto \hP \left[\psi(Z; \theta, \nua(Z; \theta_1'), \tnub(Z))\right]$ is Fr\'echet differentiable at $(\tth,  \tnua(\cdot, \tth_1))$. Namely, assume that there exists a bounded linear operator
$\mathcal{D}_{\tnua}$, such that for any $(\theta,  \nua'(\cdot, \theta_1'))$ within a small open neighborhood $\mathcal{N}$ around $(\tth,  \tnua(\cdot, \tth_1))$,
\begin{align*}
   &\|\hP \left[\psi(Z; \theta, \nua'(Z, \theta_1'), \tnub(Z))\right] - \hP \left[\psi(Z; \tth, \tnua(Z; \tth_1), \tnub(Z))\right] \\
    &\qquad\qquad\qquad - \partial_{\theta^\top}\{\hP\left[\psi(Z; {\theta}, \tnua(Z; \theta_1^*), \tnub(Z))\right]\}\vert_{\theta = \tth}(\theta - \tth) -  \mathcal{D}_{\tnua}[\nua'(\cdot,\theta_1') - \tnua(\cdot, \tth_1)]\| \\
   &= o(\|\theta - \tth\|) + o(\{\hP\left[\nua'(Z,\theta_1') - \tnua(Z; \tth_1)\right]^2\}^{1/2}).
\end{align*}
Assume further that there exists $C > 0$ such that for any  $(\theta,  \nua'(\cdot, \theta_1')) \in \mathcal{N}$
\begin{align}
    \mathcal{D}_{\tnua}[\nua'(\cdot,\theta_1') - \tnua(\cdot, \tth_1)] &= 0, \label{eq: frechet-orth}\\
    {\hP\left[\magd{\nua^*(Z,\theta_1') - \tnua(Z; \tth_1)}^2\right]^{1/2}}& \le C{\|\theta_1' - \tth_1\|}. 
    \nonumber
\end{align}
Then 
\cref{assump: jacob} is satisfied.
\end{proposition}
Here the condition in \cref{eq: frechet-orth} is an orthogonality condition using the Fr\'echet derivative, which is stronger than the G\^ateaux differentiability required in Neyman orthogonality (see \cref{assump: identification} condition \ref{assump: identification: orthogoanlity}). 
In \cref{eq: invariant-jacobian-example}, we already show that $\hP \left[\psi(Z;\theta,\eta_1(Z;\theta_1'),\eta^*_2(Z))\right]$ for the efficient estimating function in the incomplete data setting does not depend on $\eta_1$ at all. Thus, its Fr\'echet derivative with respect to $\eta_1$ trivially exists and is always $0$, and therefore our \cref{assump: jacob} will be satisfied per \cref{prop: frechet-diff}.

\section{Literature Review on Semiparametric Inference and Debiased Machine Learning}\label{sec: app-literature}
Our work is closely related to the classical semiparametric estimation literature on constructing $\sqrt{N}$-consistent and asymptotically normal estimators for low dimensional target parameters in the presence of infinitely dimensional nuisances, typically estimated by conventional nonparametric estimators such as kernel and series estimators \citep[e.g.,][]{newey1990semiparametric,newey1994the,newey1998undersmoothing,ibragimov1981statistical,levit1976on,bickel98,bickel1982on,robinson1988root,vaart1991on,andrews1994asymptotics,robins1995semiparametric,linton1996edgeworth,chen2003estimation,laan2011targeted,AiChunrong2009Sebf}.
Our work builds on the Neyman orthogonality condition introduced by \cite{Neyman1959} (see \cref{eq: frechet-orth} in \cref{prop: frechet-diff} below). 
This condition plays a critical role in many works that go beyond such nonparametric estimators, such as targeted learning \citep[e.g.,][]{laan2011targeted,van2018targeted},  inference for coefficients in high dimensional linear models \citep[e.g.,][]{belloni2016post,belloni2014pivotal,zhang2014confidence,van2014asymptotically,javanmard2014confidence,chernozhukov2015valid,ning2017general}, and semiparametric estimation with nuisances that involve  high dimensional covariates \citep[e.g., ][]{belloni2017program,smucler2019unifying,chernozhukov2018plug,farrell2015robust, belloni2014high,belloni2014inference,bradic2019sparsity}. 

\cite{ChernozhukovVictor2018Dmlf}  further advocate
the use of cross-fitting in addition to orthogonal estimating equations, so that the traditional Donsker assumption on nuisance estimators can be relaxed, and a broad array of  black-box machine learning algorithms can be used instead. 
They refer to this generic approach as DML, which provides a principled framework to estimate low-dimensional target parameters with strong asymptotic guarantees when leveraging modern machine learning methods in nuisance estimation.  
Other forms of sample splitting and cross-fitting have also appeared in
\citet{klaassen1987consistent,zheng2011cross,fan2012variance,bickel1982on,robins2013new,schick1986asymptotically,vaart_1998,robins2008higher,robins2017minimax}.
Since the DML framework was introduced, numerous works have applied it in many different problems, such as heterogeneous treatment effect estimation \citep{kennedy2020optimal,nie2017quasi,curth2020semiparametric,semenova2020debiased,oprescu2019orthogonal,fan2020estimation}, causal effects of continuous treatments \citep{colangelo2020double,oprescu2019orthogonal}, instrumental variable estimation \citep{singh2019biased,syrgkanis2019machine}, partial identification \citep{bonvini2019sensitivity,kallus2019assessing,semenova2017machine,yadlowsky2018bounds}, difference-in-difference models \citep{lu2019robust,chang2020double,zimmert2018efficient}, off-policy evaluation \citep{kallus2020double,demirer2019semi,zhou2018offline,athey2017efficient}, generalized method of moments \citep{chernozhukov2016locally,belloni2018high}, improved machine learning nuisance estimation \citep{farrell2018deep,cui2019bias}, statistical learning with nuisances \citep{foster2019orthogonal}, causal inference with surrogate observations \citep{kallus2020role}, linear functional estimation \citep{chernozhukov2018learning,chernozhukov2018double,bradic2019minimax}, etc.
Our work complements this line of research by proposing a simple but effective way to handle estimand-dependent nuisances. This type of nuisances frequently appears in efficient estimation of complex causal effects such as QTEs, and applying DML directly would require estimating a continuum of nuisances, which is challenging in practice. 

\section{Comparison with \texorpdfstring{\cite{ChernozhukovVictor2018Dmlf}}{a}}\label{sec: compare-discussion}
Our proof of \cref{thm: general-split} and the proof of Theorem 3.3 in \cite{ChernozhukovVictor2018Dmlf} are overall similar, but critically differ in Step II. In Step II, both proofs are based on the following decomposition:  
\begin{align}\label{eq: key-decomp}
     \|{J^*}^{-1}\sqrt{N}\hP_N[\psi(Z; \tth, \tnua(Z; \theta_1^*), \tnub(Z))] +\sqrt{N}(\hth - \tth)\| \le \varepsilon_N {N}^{1/2} + 2\calI_4 + 2\calI_5, 
\end{align}
where 
\begin{align*}
    \calI_4 
        &\coloneqq \sqrt{N}\sup_{r \in (0, 1), (\nua(\cdot; \theta_1'), \nub) \in \mathcal{T}_N} \|\partial_r^2 f(r; \hth, \nua(\cdot; \theta_1'), \nub)\|,  \\
    \calI_5 
        &\coloneqq \hG_N\left[\psi(Z; \hth, \hnua(Z, \hthinit), \hnub(Z)) - \psi(Z; \tth, \tnua(Z; \tth_1), \tnub(Z))\right]\|,
\end{align*}
and $\calI_5 = O_{\hP}(\delta_N)$ is proved analogously in both proofs. 

However, our proof and the proof in \cite{ChernozhukovVictor2018Dmlf} assume different rate on $\lambda_N'$ and thus $\calI_4$:
\begin{align}
\text{Our condition} ~~~  &\lambda_N'\prns{\theta} \le \left(\|\hth - \theta^*\| + {N}^{-1/2}\right)\delta_N, \label{eq: our-condition}\\
\text{Condition in \cite{ChernozhukovVictor2018Dmlf}} ~~~ &\lambda_N'\prns{\theta} \le  {N}^{-1/2}\delta_N. \label{eq: chernochukov-condition}
\end{align}

 Under our condition, $\calI_4 \le \left(\sqrt{N}\|\hth - \theta^*\| + 1\right)\delta_N$, then jointly considering the left hand side and right hand side in \cref{eq: key-decomp} gives $\|\hth - \tth\| = O_p(N^{-1/2})$, which in turn implies that $\calI_4 = O(\delta_N)$, and thus the asserted conclusion in \cref{thm: general-split}. In contrast, the counterpart condition in \cite{ChernozhukovVictor2018Dmlf} guarantees that $\calI_4 = O(\delta_N)$ directly without needing to consider both sides of \cref{eq: key-decomp} jointly.

Now we use the example of estimating equation for incomplete data to show that the condition \cref{eq: chernochukov-condition} in \cite{ChernozhukovVictor2018Dmlf} generally requires stronger conditions for the convergence rates of nuisance estimators than our condition \cref{eq: our-condition}.

According to \cref{eq: thm2eq1}, under suitable regularity conditions, 
\begin{align*}
    \|\partial^2_{r}f(r; \hth, \mu(X, T; \theta_1'), \pi)\|  
    &= O(\rho_{\pi, N}\rho_{\mu, N}) + O_p(\rho_{\pi, N}\rho_{\theta, N})+ O(\|\hth - \tth\|^2) + O(\rho_{\pi, N}\|\hth_1 - \tth_1\|) %
\end{align*}
Since Step I in the proof of \cref{thm: general-split} already proves that $\|\hth - \tth\| \le \frac{\rho_{\pi, N}}{\delta_N}$, we need $\rho_{\pi, N}\rho_{\mu, N} \le \delta_N N^{-1/2},  \rho_{\pi, N}\rho_{\theta, N} \le  \delta_N N^{-1/2}$, and $\rho_{\pi, N} \le \delta_N^2$ to guarantee our condtion. Thus our condition in \cref{eq: our-condition} only requires that the product error rates to vanish faster than $O(N^{-1/2})$, which is common in debiased machine learning for linear estimating equation \citep{ChernozhukovVictor2018Dmlf}. 

In contrast, to guarantee the condition in \cite{ChernozhukovVictor2018Dmlf} given in \cref{eq: chernochukov-condition}, we need to assume that $\rho_{\pi, N} \le \delta_N^{3/2}N^{-1/4}$, besides the conditions on product error rates. Therefore, following the proof in \cite{ChernozhukovVictor2018Dmlf} directly will require the propensity score to converge faster than $O(N^{-1/4})$, no matter how fast the initial estimator $\hthinit$ and the regression estimator $\hq(\cdot, \hthinit)$ converge. 

\section{Proofs}
\subsection{Proofs for \texorpdfstring{\cref{sec: method}}{a}}
\begin{proof}[Proof for \cref{prop: frechet-diff}]
For any $\theta = (\theta_1, \theta_2)$ such that $(\theta,  \tnua(\cdot, \theta)) \in \mathcal{N}$, the asserted Fr\'echet differentiability and orthogonality condition imply that
\begin{align*}
   &\|\hP \left[\psi(Z; \theta, \nua^*(Z, \theta_1), \tnub(Z))\right] - \hP \left[\psi(Z; \tth, \tnua(Z; \tth_1), \tnub(Z))\right] \\
   &- \partial_{\theta}\{\hP\left[\psi(Z; {\theta}, \tnua(Z; \theta_1^*), \tnub(Z))\right]\}\vert_{\theta = \tth}(\theta - \tth) \| = o(\|\theta - \tth\|).
\end{align*}
This means that $J^* = \partial_{\theta}\{\hP\left[\psi(Z; {\theta}, \tnua(Z; \theta_1), \tnub(Z))\right]\}\vert_{\theta = \tth}$. 
\end{proof}

\subsection{Proofs for \texorpdfstring{\cref{sec: general}}{a}}

\begin{proof}[Proof for \cref{thm: general-split}]
Fix any sequence $\{P_N\}_{N \ge 1}$ that generates the observed data \\ $(Z_1, \dots, Z_N)$ and satisfies that $P_N \in \mathcal P_N$ for all $N \ge 1$. Because this sequence is chosen arbitrarily, to prove that the asserted conclusion holds uniformly over $P \in \mathcal P_N$, we only need to prove 
\begin{align*}
\sqrt{N}\Sigma^{-1/2}(\hth - \tth) = \frac{1}{\sqrt{N}}\sum_{i = 1}^N \Sigma^{-1/2}\left[{J^*}^{-1}\psi(Z; \tth, \tnua(Z; \theta_1^*), \tnub(Z))\right] + O_{P_N}(\rho_N) \overset{d}{\to} \mathcal N(0, I_d).
\end{align*}
For $k = 1, \dots, K$, we use $\hP_{N, k}$ to represent the empirical average operator based on $\Dcal_k$. For example, $\hP_{N, k}\left[\psi(Z; \tth, \tnua(Z; \theta_1^*), \tnub(Z))\right] = \frac{1}{|\Dcal_{k}|}\sum_{i \in \Dcal_k}\psi(Z_i; \tth, \tnua(Z_i; \theta_1^*), \tnub(Z_i))$. 
Analogously, $\hP_{N}$ is the empirical average operator for the whole dataset, i.e., $\hP_{N}f(Z) = \frac{1}{N}\sum_{i = 1}^N f(Z_i)$. $\mathbb G_{N, k}$ is the empirical process operator $\sqrt{N}\left(\hP_{N, k} - \hP\right)$. 
Moreover, for a given $N$, $\hP_{N, k}$, $\hP_{N}$ and the population average operator $\hP$ are all derived from the underlying true distribution $P_N$, but we supress such dependence for ease of notation. 
Throughout the proof, we condition on the event  $(\hnua(\cdot, \hthinit), \hnub(\cdot)) \in \mathcal{T}_N$, which happens with at least $P_N$-probability $1 - \Delta_N$ according to \cref{assump: error} condition \ref{assump: error: nuisance-set}.
All statements involving $o(\cdot)$, $O_{P_N}(\cdot)$ or $\lesssim$ notations in this proof depend on only constants pre-specified in \cref{assump: identification,assump: error}, and do not depend on constants specific to the instance $P_N$. This should be clear from the proof, and the fact that the maximal inequality in Lemma 6.2 of \cite{ChernozhukovVictor2018Dmlf} only depend on pre-specified parameters. 
Here we prove the asymptotic distribution of $\hth$ given in \cref{def: LDML2} first.

\textbf{Step I: Prove a preliminary convergence rate for $\hth$:} $\|\hth - \tth\| \le \tau_N$ with $P_N$-probability $1-o(1)$. Here we prove this by showing that with $P_N$-probability $1 - o(1)$, 
\begin{align}\label{eq: thm1eq1}
 \left\|\hP\left[\psi(Z; \hth, \tnua(Z; \theta_1^*), \tnub(Z))\right]\right\| = o(\tau_N)
\end{align}
so that \cref{assump: identification} implies that 
\[
\|J^*(\hth - \tth)\| \wedge c_2 = o(\tau_N).
\]
Since the singular values of $J^*$ are lower bounded by $c_3 > 0$, we can conclude that with $P_N$-probability $1 - o(1)$, $\|\hth - \tth\| \le \tau_N$ for $N$ exceeding an instance-independent threshold. 

In order to prove \cref{eq: thm1eq1}, we use the following decomposition:
\begin{align*}
&\hP\left[\psi(Z; \hth, \tnua(Z; \theta_1^*), \tnub(Z))\right] \\
=& \underbrace{\frac{1}{K}\sum_{k = 1}^K\hP\left[\psi(Z; \hth, \tnua(Z; \theta_1^*), \tnub(Z))\right] - \hP\left[\psi(Z; \hth, \hnua\pk(Z, \hthinit\pk), \hnub\pk(Z))\right]}_{(a)} \\
+& \underbrace{\frac{1}{K}\sum_{k = 1}^K\left\{\hP\left[\psi(Z; \hth, \hnua\pk(Z, \hthinit\pk), \hnub\pk(Z))\right] - \hP_{N, k}\left[\psi(Z; \hth, \hnua\pk(Z, \hthinit\pk), \hnub\pk(Z))\right]\right\}}_{(b)} \\
+& \underbrace{\frac{1}{K}\sum_{k = 1}^K\left\{\hP_{N, k}\left[\psi(Z; \hth, \hnua\pk(Z, \hthinit\pk), \hnub\pk(Z))\right] - \hP_{N, k}\left[\psi(Z; \tth, \hnua\pk(Z, \hthinit\pk), \hnub\pk(Z))\right]\right\}}_{(c)} \\
+& \underbrace{\frac{1}{K}\sum_{k = 1}^K\left\{\hP_{N, k}\left[\psi(Z; \tth, \hnua\pk(Z, \hthinit\pk), \hnub\pk(Z))\right] - \hP\left[\psi(Z; \tth, \hnua\pk(Z, \hthinit\pk), \hnub\pk(Z))\right]\right\}}_{(d)}  \\
+& \underbrace{\frac{1}{K}\sum_{k = 1}^K\left\{\hP\left[\psi(Z; \tth, \hnua\pk(Z, \hthinit\pk), \hnub\pk(Z))\right] - \hP\left[\psi(Z; \tth, \tnua(Z; \tth_1), \tnub(Z))\right]\right\}}_{(e)}.
\end{align*}
Denote 
\begin{align*}
    \calI_{1, k} = \sup_{\theta \in {\Theta}}\|\hP\left[\psi(Z; \theta, \tnua(Z; \theta_1^*), \tnub(Z))\right] - \hP\left[\psi(Z; \theta, \hnua\pk(Z, \hthinit\pk), \hnub\pk(Z))\right]\|, \\
    \calI_{2, k} = \sup_{\theta \in {\Theta}}\|\hP_{N, k}\left[\psi(Z; \theta, \hnua\pk(Z, \hthinit\pk), \hnub\pk(Z))\right] - \hP\left[\psi(Z; \theta, \hnua\pk(Z, \hthinit\pk), \hnub\pk(Z))\right]\|.
\end{align*}
Then obviously, 
\begin{align*}
    (a) + (e) \le \frac{2}{K}\sum_{k = 1}^K\calI_{1, k}, (b) + (d) \le \frac{2}{K}\sum_{k = 1}^K\calI_{2,k}. 
\end{align*}
To bound (c), note that \cref{eq: approx-LDML2} implies 
\begin{align*}
&\|\frac{1}{K}\sum_{k = 1}^K \hP_{N, k}\left[\psi(Z; \hth, \hnua\pk(Z, \hthinit\pk), \hnub\pk(Z))\right]\| \\
 \le& \inf_{\theta \in \Theta}\|\frac{1}{K}\sum_{k = 1}^K\hP_{N, k}\left[\psi(Z; \theta, \hnua\pk(Z, \hthinit\pk), \hnub\pk(Z))\right]\| + \varepsilon_N \\
  \le&  \frac{1}{K}\sum_{k = 1}^K\|\hP_{N, k}\left[\psi(Z; \tth, \hnua\pk(Z, \hthinit\pk), \hnub\pk(Z))\right]\| + \varepsilon_N. 
\end{align*}
Thus 
\begin{align*}
    (c)  
&\le \frac{1}{K}\sum_{k = 1}^K\|\hP_{N, k}\left[\psi(Z; \hth, \hnua\pk(Z, \hthinit\pk), \hnub\pk(Z))\right]\|  + \frac{1}{K}\sum_{k = 1}^K\|\hP_{N, k}\left[\psi(Z; \tth, \hnua\pk(Z, \hthinit\pk), \hnub\pk(Z))\right]\|  \\
&\le \frac{2}{K}\sum_{k = 1}^K\|\hP_{N, k}\left[\psi(Z; \tth, \hnua\pk(Z, \hthinit\pk), \hnub\pk(Z))\right]\| + \varepsilon_N \\
&\le \frac{2}{K}\sum_{k = 1}^K\bigg\|\hP_{N, k}\left[\psi(Z; \tth, \hnua\pk(Z, \hthinit\pk), \hnub\pk(Z))\right] - \hP\left[\psi(Z; \tth, \hnua\pk(Z, \hthinit\pk), \hnub\pk(Z))\right]\bigg\| \\
&+ \frac{2}{K}\sum_{k = 1}^K\bigg\|\hP\left[\psi(Z; \tth, \hnua\pk(Z, \hthinit\pk), \hnub\pk(Z))\right] - \hP\left[\psi(Z; \tth, \tnua(Z; \tth_1), \tnub(Z))\right]\bigg\| +  \varepsilon_N \\
&\le  \frac{2}{K}\sum_{k = 1}^K\calI_{1, k} +  \frac{2}{K}\sum_{k = 1}^K\calI_{2, k} + \varepsilon_N.
\end{align*}

Therefore, 
\[
\hP\left[\psi(Z; \hth, \tnua(Z; \theta_1^*), \tnub(Z))\right]  \le  \frac{4}{K}\sum_{k = 1}^K\calI_{1, k} +  \frac{4}{K}\sum_{k = 1}^K\calI_{2, k} + \varepsilon_N.
\]
Note that \cref{assump: error} condition \ref{assump: error: rate-condition}  implies that $\calI_{1, k} \le \delta_N\tau_N$ and the \cref{assump: error} condition \ref{assump: error: approximation}  implies that $\varepsilon_N \le \delta_N N^{-1/2} = o(\tau_N)$. 

To bound $\calI_{2, k}$, note that conditionally on $\hnua\pk(Z, \hthinit\pk), \hnub\pk(Z)$, 
the function class \\ $\calF_{\hat{\eta}\pk, \hthinit\pk} = \{\psi_j(\cdot; \theta, \hnua\pk(\cdot, \hthinit\pk), \hnub\pk(\cdot)): j = 1, \dots, d, \theta \in \Theta\}$ satisfies the asserted entropy condition in \cref{assump: identification}, and has envelope $F_{1, \hat{\eta}\pk, \hthinit\pk}$ that satisfies 
$$
\sup_{\theta \in \Theta}\hP\left[\psi(Z; \theta, \hnua\pk(Z, \hthinit\pk), \hnub\pk(Z)) \right]^2 \le \hP\left[F^2_{1, \hat{\eta}\pk, \hthinit\pk}\right] < C_{q, c_7}
$$
for a positive constant $C_{q, c_7}$ that only depends on $q$ and $c_7$ specified in \cref{assump: identification}.

Then conditionally on $\hthinit, \hnua\pk(Z, \hthinit\pk), \hnub\pk(Z)$,  we can use Lemma 6.2 eq. (A.1) in \cite{ChernozhukovVictor2018Dmlf} to prove that with $P_N$-probability $1 - o(1)$,  
\begin{align}\label{eq: thm1eq2}
    \sup_{\theta \in \Theta}\hG_{N, k}\left[\psi(Z; \theta, \hnua\pk(Z, \hthinit\pk), \hnub\pk(Z))\right] \lesssim \log N(1 + N^{-1/2 + 1/q}),
\end{align}
which also holds unconditionally according to Lemma 6.1 of in \cite{ChernozhukovVictor2018Dmlf}. 
This further implies that $\calI_{2, k} \lesssim N^{-1/2}\log N(1 + N^{-1/2 + 1/q}) = o(N^{-1/2}\log^2 N(1 + N^{-1/2 + 1/q})) = o(\tau_N)$. Thus
\begin{align*}
\hP\left[\psi(Z; \hth, \tnua(Z; \theta_1^*), \tnub(Z))\right]  
  &\le  4\delta_N\tau_N + 4N^{-1/2}\log N(1 + N^{-1/2 + 1/q}) + \delta_N N^{-1/2}  = o(\tau_N).
\end{align*}

\textbf{Step II: Linearization and $\sqrt{N}-$Consistency.} In Step I, we proved that $\|\hth - \tth\| \le \tau_N$ with  $P_N$-probability $1 - o(1)$. Conditioned on this event,  we will show that 
\begin{align}\label{eq: thm1eq3}
    &\|\sqrt{N}\hP_N[\psi(Z; \tth, \tnua(Z; \theta_1^*), \tnub(Z))] + \sqrt{N}J^*(\hth - \tth)\| \nonumber \\
    \le& \varepsilon_N {N}^{1/2} + \calI_{3} + \calI_4 + \frac{1}{K}\sum_{k = 1}^K\calI_{5, k},
\end{align}
where 
\begin{align*}
    \calI_{3} 
        &\coloneqq \inf_{\theta \in \Theta} \sqrt{N}\left\|\frac{1}{K}\sum_{k = 1}^K\hP_N[\psi(Z; \theta, \hnua\pk(Z, \hthinit\pk), \hnub\pk(Z))]\right\|, \\
    \calI_4 
        &\coloneqq \sqrt{N}\sup_{r \in (0, 1), (\nua(\cdot; \theta_1'), \nub) \in \mathcal{T}_N} \|\partial_r^2 f(r; \hth, \nua(\cdot; \theta_1'), \nub)\|,  \\
    \calI_{5, k} 
        &\coloneqq \sup_{\|\theta - \theta^*\| \le \tau_N}\|\hG_{N, k}\left[\psi(Z; \theta, \hnua\pk(Z, \hthinit\pk), \hnub\pk(Z)) - \psi(Z; \tth, \tnua(Z; \tth_1), \tnub(Z))\right]\|.
\end{align*}

Here \cref{assump: error} condition \ref{assump: error: rate-condition} guarantees that $\calI_4 \le \delta_N \left(1 + \sqrt{N}\|\hth - \tth\|\right)$ and the assumption that  $\varepsilon_N = \delta_N N^{-1/2}$ guarantees that $\varepsilon_N {N}^{1/2} \le \delta_N$. In step III and IV, we will further bound $\calI_{5, k} \lesssim \rho_N' \coloneqq (N^{-1/2 + 1/q} + r'_N){\log N} + r'_N {\log^{1/2}(1/r'_N)} + N^{-1/2 + 1/q}\log(1/r’_N) \lesssim \delta_N$ and $\calI_3 \le \calI_4 + \frac{1}{K}\sum_{k = 1}^K \calI_{5, k}$ respectively.

Consequently, with $P_N$-probability $1 - o(1)$, 
\begin{align}\label{eq: thm1eq4c}
    \|\sqrt{N}\hP_N[\psi(Z; \tth, \tnua(Z; \theta_1^*), \tnub(Z))] +\sqrt{N}{J^*}(\hth - \tth)\| \lesssim \left(\delta_N \left(1 + \sqrt{N}\|\hth - \tth\|\right)\right) + \rho_N' + \delta_N.
\end{align}
This implies that 
\begin{align*}
& \sqrt{N}\|\hth - \tth\| -  \|\sqrt{N}J^{* - 1}\hP_N[\psi(Z; \tth, \tnua(Z; \theta_1^*), \tnub(Z))]\| \\
\le& \|\sqrt{N}(\hth - \tth) + \sqrt{N}J^{* - 1}\hP_N[\psi(Z; \tth, \tnua(Z; \theta_1^*), \tnub(Z))]\| \\
\lesssim& \|J^{*-1}\| \left[\left(\delta_N \left(1 + \sqrt{N}\|\hth - \tth\|\right)\right) + \rho_N' + \delta_N \right]
\end{align*}
and  
\begin{align*}
\sqrt{N}\|\hth - \tth\| \lesssim \frac{1}{c_3}\left[\left(\delta_N \left(2 + \sqrt{N}\|\hth - \tth\|\right)\right) + \rho_N \right] +  \|\sqrt{N}J^{* - 1}\hP_N[\psi(Z; \tth, \tnua(Z; \theta_1^*), \tnub(Z))]\|.
\end{align*}
By \cref{assump: identification} condition \ref{assump: identification: moment} and Markov inequality, $\|\sqrt{N}J^{* - 1}\hP_N[\psi(Z; \tth, \tnua(Z; \theta_1^*), \tnub(Z))]\| = O_{P_N}(\sqrt{c_6})$. Thus, with $P_N$-probability $1 - o(1)$, 
\begin{align*}
\sqrt{N}\|\hth - \tth\| \lesssim \delta_N + \rho_N'. 
\end{align*}
Plugging this back into \cref{eq: thm1eq4c} gives 
\begin{align*}
\|\sqrt{N}\hP_N[\psi(Z; \tth, \tnua(Z; \theta_1^*), \tnub(Z))] +\sqrt{N}{J^*}(\hth - \tth)\| = O_{P_N}(\delta_N + \rho'_N).
\end{align*}
Thus, 
\begin{align*}
  &\|\Sigma^{-1/2}{J^*}^{-1}\sqrt{N}\hP_N[\psi(Z; \tth, \tnua(Z; \theta_1^*), \tnub(Z))] +\Sigma^{-1/2}\sqrt{N}(\hth - \tth)\|  \\
\le&~ \|\Sigma^{-1/2}\|\|{J^*}^{-1}\|\|\sqrt{N}\hP_N[\psi(Z; \tth, \tnua(Z; \theta_1^*), \tnub(Z))] +\sqrt{N}{J^*}(\hth - \tth)\| \\
\lesssim& ~ \delta_N + \rho'_N = \rho_N,
\end{align*}
because $\|{J^*}^{-1}\| \le 1/c_3$ and $\|\Sigma^{-1/2}\| \le 1/\sqrt{c_5}$. 

Now we prove the decomposition \cref{eq: thm1eq3}. Note that for any $\theta \in \Theta$ and $(\nua(\cdot, \theta_1), \nub) \in \mathcal{T}_N$
\begin{align}\label{eq: thm1eq4}
    &\sqrt{N}\left\{\frac{1}{K}\sum_{k = 1}^K\hP_{N, k}\left[\psi(Z; \theta, \nua(Z, \theta_1), \nub(Z))\right]\right\} \nonumber \\
    &= \sqrt{N}\bigg\{\frac{1}{K}\sum_{k = 1}^K\hP_{N, k}\left[\psi(Z; \theta, \nua(Z, \theta_1), \nub(Z))\right] - \hP\left[\psi(Z; \theta, \nua(Z, \theta_1), \nub(Z))\right] \nonumber \\
    &+  \frac{1}{K}\sum_{k = 1}^K\hP\left[\psi(Z; \theta, \nua(Z, \theta_1), \nub(Z))\right] - \hP\left[\psi(Z; \tth, \tnua(Z; \tth_1), \tnub(Z))\right] \nonumber \\
    &+ \frac{1}{K}\sum_{k = 1}^K \hP\left[\psi(Z; \tth, \tnua(Z; \tth_1), \tnub(Z))\right] - \hP_{N, k}\left[\psi(Z; \tth, \tnua(Z; \tth_1), \tnub(Z))\right]\nonumber  \\
    &+ \hP_N\left[\psi(Z; \tth, \tnua(Z; \tth_1), \tnub(Z))\right]\bigg\} \nonumber \\
    &= \frac{1}{K}\sum_{k = 1}^K\hG_{N, k}\left[\psi(Z; \theta, \nua(Z, \theta_1), \nub(Z))) - \psi(Z; \tth, \tnua(Z; \tth_1), \tnub(Z))\right] \nonumber \\
    &+ \frac{1}{K}\sum_{k = 1}^K\sqrt{N}\bigg\{\hP\left[\psi(Z; \theta, \nua(Z, \theta_1), \nub(Z))\right] - \hP\left[\psi(Z; \tth, \tnua(Z; \tth_1), \tnub(Z))\right]  \bigg\}\nonumber \\
    &+ \hP_N\left[\psi(Z; \tth, \tnua(Z; \tth_1), \tnub(Z))\right].
\end{align}

If we apply \cref{eq: thm1eq4} with $\theta = \hth$ and $(\nua(\cdot, \theta_1), \nub)$ equal $(\hnua\pk(\cdot, \hthinit\pk), \hnub\pk)$ for the $k$th fold, and apply \cref{eq: approx-LDML2}, then 
\begin{align}\label{eq: thm1eq5}
    &\bigg\|\frac{1}{K}\sum_{k = 1}^K\hG_{N, k}\left[\psi(Z; \hth, \hnua\pk(Z, \hthinit\pk), \hnub\pk(Z)) - \psi(Z; \tth, \tnua(Z; \tth_1), \tnub(Z))\right] \nonumber \\
    +& \sqrt{N}\bigg\{\frac{1}{K}\sum_{k = 1}^K\hP\left[\psi(Z; \hth, \hnua\pk(Z, \hthinit\pk), \hnub\pk(Z))\right] \nonumber- \hP\left[\psi(Z; \tth, \tnua(Z; \tth_1), \tnub(Z))\right]  \bigg\} \\
    +& \sqrt{N}\hP_{N}\left[\psi(Z; \tth, \tnua(Z; \tth_1), \tnub(Z))\right] \bigg \|\nonumber \\
    =& \sqrt{N}\left\|\frac{1}{K}\sum_{k = 1}^K\hP_{N, k}\left[\psi(Z; \hat\theta, \hnua\pk(Z, \hthinit\pk), \hnub\pk(Z))\right]\right\| \nonumber \\
    \le& \sqrt{N}\inf_{\theta \in \Theta}\left\|\frac{1}{K}\sum_{k = 1}^K\hP_{N, k}\left[\psi(Z; \theta, \hnua\pk(Z, \hthinit\pk), \hnub\pk(Z))\right]\right\| + \varepsilon_N\sqrt{N}.
\end{align}
Here 
\begin{align}\label{eq: thm1Gk}
    \|\hG_{N, k}\left[\psi(Z; \hth, \hnua\pk(Z, \hthinit\pk), \hnub\pk(Z)) - \psi(Z; \tth, \tnua(Z; \tth_1), \tnub(Z))\right]\| \le \calI_{5, k}
\end{align}
and the second order tayler expansion at $r = 0$ gives that for some data-dependent $\tilde{r} \in (0, 1)$, 
\begin{align}\label{eq: thm1eq6}
    &\sqrt{N}\bigg\{\hP\left[\psi(Z; \hth, \hnua\pk(Z, \hthinit\pk), \hnub\pk(Z))\right]
    - \hP\left[\psi(Z; \tth, \tnua(Z; \tth_1), \tnub(Z))\right]\bigg\} \nonumber \\
    =& \sqrt{N}\left[f(1; \hth, \hnua\pk(\cdot, \hthinit\pk), \hnub\pk) - f(0; \hth, \hnua\pk(\cdot, \hthinit\pk), \hnub\pk)\right] \nonumber\\
    =& \sqrt{N}\bigg\{J^*(\hat{\theta} - \theta^*) + \partial_{r}\big\{\hP\left[\psi(Z; \tth, \tnua(Z; \tth_1) + r(\hnua\pk(Z, \hthinit\pk) - \tnua(Z; \tth_1)), \tnub)\right]\big\}\vert_{r = 0} \nonumber\\
    +& \partial_{r}\big\{\hP\left[\psi(Z; \tth, \tnua(Z; \tth_1) , \tnub + r(\hnub\pk(Z) - \tnub(Z)))\right]\big\}\vert_{r = 0}  + \partial_r^2 f(r; \hth, \hnua\pk(\cdot, \hthinit\pk), \hnub\pk)\vert_{r = \tilde{r}}\bigg\} \nonumber \\
    =& \sqrt{N}\bigg\{J^*(\hat{\theta} - \theta^*)+\partial_r^2 f(r; \hth, \hnua\pk(\cdot, \hthinit\pk), \hnub\pk)\vert_{r = \tilde{r}}\bigg\}
\end{align}
where the third equality uses the Neyman orthogonality in \cref{assump: identification} condition \ref{assump: identification: orthogoanlity}. 

Combining \cref{eq: thm1eq5}, \cref{eq: thm1Gk} and \cref{eq: thm1eq6} gives decomposition \cref{eq: thm1eq3}.

\textbf{Step III: bounding $\calI_{5, k}$.} To bound $\calI_{5, k}$, we still condition on  $\hnua\pk(\cdot, \hthinit\pk), \hnub\pk$, and then apply Lemma 6.2 in \cite{ChernozhukovVictor2018Dmlf} with function class 
\[
    \calF'_{\hat{\eta}\pk, \hthinit\pk} = \{\psi_j(\cdot; \theta, \hnua\pk(\cdot, \hthinit\pk), \hnub\pk) - \psi_j(\cdot; \tth, \tnua(\cdot, \tth), \tnub): j = 1, \dots, d, \theta \in \Theta, \|\theta - \tth\| \le \tau_N\}.
\]
We can verify that $\calF'_{\hat{\eta}\pk, \hthinit\pk}$ satisfies similar entropy condition with envelope $F_{1, \hat{\eta}\pk, \hthinit\pk} + F_{1, \eta^*, \theta^*_1}$. Moreover, \cref{assump: error} implies that 
\[
\sup_{\|\theta - \tth\| \le \tau_N}\|\psi(Z; \theta, \hnua\pk(Z, \hthinit\pk), \hnub\pk(Z)) - \psi(Z; \tth, \tnua(Z, \tth), \tnub(Z))\|_{\hP, 2} \le r_N'.
\]
Thus conditionally on $\hthinit, \hnua\pk(Z, \hthinit\pk), \hnub\pk(Z)$, we can use Lemma 6.2 eq. (A.1) in \cite{ChernozhukovVictor2018Dmlf} to show that with $P_N$-probability $1- o(1)$,
\[
    \calI_{5, k} \lesssim (N^{-1/2 + 1/q} + r'_N){\log N} + r'_N {\log^{1/2}(1/r'_N)} + N^{-1/2 + 1/q}\log(1/r’_N),
\]
which also holds unconditionally according to Lemma 6.1 in \cite{ChernozhukovVictor2018Dmlf} . 

\textbf{Step IV: bounding $\calI_3$.} Let $\overline{\theta} = \tth - {J^*}^{-1}\hP_N\left[\psi(Z; \tth, \tnua(Z, \tth), \tnub(Z))\right]$.\\ Since $\hP\left[\psi(Z; \tth, \tnua(Z, \tth), \tnub(Z))\right] = 0$, ${J^*}$ is nonsingular with singular values bounded away from 0 by $c_3$, and $\|\hP_N\left[\psi(Z; \tth, \tnua(Z, \tth), \tnub(Z))\right]\| = O_{P_N}(N^{-1/2})$, $\|\overline{\theta} - \tth\| = O_{P_N}(N^{-1/2}) = o_{P_N}(\tau_N)$. According to \cref{assump: identification} condition \ref{assump: identification: interior}, $\overline{\theta} \in \Theta$ with $P_N$ probability $1 - o(1)$. 
Therefore, 
\[
    \calI_3 \le  \sqrt{N}\left\|\frac{1}{K}\sum_{k = 1}^K\hP_N[\psi(Z; \overline{\theta}, \hnua\pk(Z, \hthinit\pk), \hnub\pk(Z))]\right\|
\]
Then apply the linearization \cref{eq: thm1eq4} and  taylor expansion similar to \cref{eq: thm1eq6} with $\theta = \overline \theta$ and $(\nua(\cdot, \theta_1), \nub)$ equal $(\hnua\pk(\cdot, \hthinit\pk), \hnub\pk)$ for the $k$th fold, we can get that 
\begin{align*}
    &\sqrt{N}\left\|\frac{1}{K}\sum_{k = 1}^K\hP_{N, k}[\psi(Z; \overline{\theta}, \hnua\pk(Z, \hthinit\pk), \hnub\pk(Z))]\right\| \\
    \le& \sqrt{N}\|\hP_N[\psi(Z; \overline{\theta}, \tnua(Z; \theta_1^*), \tnub(Z))] +{J^*}(\overline{\theta} - \tth)\| + \calI_4 + \frac{1}{K}\sum_{k = 1}^K\calI_{5, k} \\
    =& \calI_4 + \frac{1}{K}\sum_{k = 1}^K\calI_{5, k}. 
\end{align*}
where the last equality here holds because $\hP_N[\psi(Z; \overline{\theta}, \tnua(Z; \theta_1^*), \tnub(Z))] +{J^*}(\overline{\theta} - \tth) = 0$ as a consequence of the special construction of $\overline{\theta}$.

\textbf{Extension: $\hat{\theta}$ defined in \cref{def: LDML1}.} By applying step I to IV to sample estimating equation \cref{eq: approx-LDML1}, we can get that for $k = 1, \dots, K$, 
\begin{align*}
\sqrt{N/K}\Sigma^{-1/2}(\hth\pk - \tth) = \frac{1}{\sqrt{N/K}}\sum_{i \in \Dcal_k} \Sigma^{-1/2}{J^*}^{-1}\psi(Z_i; \tth, \tnua(Z_i; \theta_1^*), \tnub(Z_i)) + O_P(\rho_{N/K}).
\end{align*}
Since $K$ is a fixed integer that does not grow with $N$, the equation above implies that the asserted conclusion in \cref{thm: general-split} also holds for $\hat{\theta} = \frac{1}{K}\sum_{k = 1}^K \hat\theta\pk$. 
\end{proof}

\subsection{Proofs for \texorpdfstring{\cref{sec: inference}}{a}}

\begin{proof}[Proof of \cref{thm:cf}]
We still consider data generating processes $\braces{P_N}_{N \ge 1}$ defined in the proof for \cref{thm: general-split}, and define $\otimes a=aa^{\top}$. Now we prove that 
\begin{align}
    &\| \hP_{N,k}[\otimes \psi(Z; \hat \theta, \hnua^{(k)}(Z, \hthall^{(k)}), \hnub^{(k)}(Z))  ]-\hP[ \otimes \psi(Z; \theta^*, \nua^*(Z, \theta^{*}_1), \nub^*(Z))]\|=O_{P_N}(\rho''_N). \label{eq:all_state2}
\end{align}
for any $k\in [1,\cdots,K]$. Then, the statement in \cref{thm:cf} is immediately concluded. For all $j,l\in [1,\cdots,d]\,(d=d_1+d_2)$,  \cref{eq:all_state2} follows once we have $ \mathcal{I}_{jl}=O_{P_N}(\rho''_N)$, where
\begin{align*}
    \mathcal{I}_{jl}&:=|\hP_{N,k}[ \psi_j(Z; \hat \theta, \hnua^{(k)}, \hnub^{(k)})  \psi_l(Z; \hat \theta,\hnua^{(k)},  \hnub^{(k)})]
    -\hP[\psi_j(Z;  \theta^*, \nua^*, \nub^*)\psi_l(Z;  \theta^*, \nua^*, \nub^*) ] |. 
\end{align*}
Here, to simplify the notation, we use $\hnua^{(k)}=\hnua^{(k)}(Z,\hthall^{(k)}), \nua^{*}=\nua^{*}(Z, \theta^{*}_1),\hnub^{(k)}=\hnub^{(k)}(Z,\hthall^{(k)}),\nub^{*}=\nub^{*}(Z, \theta^{*}_2)$.
Obviously we have $ \mathcal{I}_{jl}\leq\mathcal{I}_{jl,1}+\mathcal{I}_{jl,2},$ where 
\begin{align*}
    \mathcal{I}_{jl,1} &=| \hP_{N,k}[ \psi_j(Z; \hat \theta, \hnua^{(k)}, \hnub^{(k)})  \psi_l(Z; \hat \theta,\hnua^{(k)},  \hnub^{(k)})]-\hP_{N,k}[\psi_j(Z;  \theta^*, \nua^*, \nub^*)\psi_l(Z;  \theta^*, \nua^*, \nub^*) ]|,\\
    \mathcal{I}_{jl,2} &= |\hP_{N,k}[\psi_j(Z;  \theta^*, \nua^*, \nub^*)\psi_l(Z;  \theta^*, \nua^*, \nub^*) ]- \hP[\psi_j(Z;  \theta^*, \nua^*, \nub^*)\psi_l(Z;  \theta^*, \nua^*, \nub^*) ]|, 
\end{align*}
and we show that each term here is $O_p(\rho''_N)$. 

We first bound $  \mathcal{I}_{jl,2} $. This is upper bounded as 
\begin{align*}
    \hP[\mathcal{I}^2_{jl,2}] & \leq N^{-1}\hP[\psi^2_j(Z;  \theta^*, \nua^*, \nub^*)\psi^2_l(Z;  \theta^*, \nua^*, \nub^*) ]\\
& \leq N^{-1}\{\hP[\psi^4_j(Z;  \theta^*, \nua^*, \nub^*)]\hP[\psi^4_l(Z;  \theta^*, \nua^*, \nub^*) ]\}^{1/2} \\
& \leq N^{-1}\hP[\|\psi(Z;  \theta^*, \nua^*, \nub^*)\|^4]\leq N^{-1}C^4. 
\end{align*}
Here, we use the fourth moment assumption in \cref{assump:variance}. From conditional Markov inequality, we have $\mathcal{I}_{jl,2}=O_{P_N}(1/N^{-1/2})$. 

Next, we bound $\mathcal{I}_{jl,1}$. Following the proof of Theorem 3.2 \citep{ChernozhukovVictor2018Dmlf}, we have 
\begin{align*}
    &\mathcal{I}^2_{jl,1} \leq R_N\times \braces{\hP_{N,k}[\|\psi(Z; \theta^*, \nua^*, \nub^*  \|^2 ] +R_N }, \\
    & R_N=\hP_{N,k}[\|\psi(Z; \hat \theta, \hnua, \hnub )-\psi(Z; \theta^*, \nua^*, \nub^*) \|^2  ]. 
\end{align*}
In addition, from the fourth moment assumption in \cref{assump:variance}
\begin{align*}
  \hP[\hP_{N,k}[\|\psi(Z; \theta^*, \nua^*, \nub^*) \|^2 ]]= \hP[\|\psi(Z; \theta^*, \nua^*, \nub^*) \|^2]\leq C^2.
\end{align*}
It follows from Markov inequality that  
\begin{align*}
    \hP_{N,k}[\|\psi(Z; \theta^*, \nua^*, \nub^*)\|^2 ]=O_{P_N}(1).
\end{align*}
It remains to bound $R_N$. We have 
\begin{align}\nonumber 
    R_N &=\hP_{N,k}[\|\psi(Z; \hat \theta, \hnua, \hnub )-\psi(Z; \theta^*, \nua^*, \nub^*) \|^2  ] \\
     & \leq \hP_{N,k}[\|\psi(Z;  \hat \theta, \nua^*, \nub^*)-\psi(Z; \theta^*, \nua^*, \nub^*) \|^2  ]+ \hP_{N,k}[\|\psi(Z;   \hat \theta, \hnua, \hnub  )-\psi(Z;  \hat \theta, \nua^*, \nub^*) \|^2  ]. \label{eq:r_N2}
\end{align}
Then, the first term of \cref{eq:r_N2} is upper bounded with $P_N$-probability $1-o(1)$ as  
\begin{align*}
&\hP_{N,k}[\|\psi(Z;  \hat \theta, \nua^*, \nub^*)-\psi(Z; \theta^*, \nua^*, \nub^*) \|^2  ]\\
&=\frac{1}{\sqrt{N}}\hG_{N,k}[\|\psi(Z;  \hat \theta, \nua^*, \nub^*)-\psi(Z; \theta^*, \nua^*, \nub^*) \|^2  ]+ \hP[\|\psi(Z;  \hat \theta, \nua^*, \nub^*)-\psi(Z; \theta^*, \nua^*, \nub^*) \|^2  ]\\
&\leq \sup_{\theta \in \Theta }\frac{1}{\sqrt{N}}\hG_{N,k}[\|\psi(Z;  \theta, \nua^*, \nub^*)-\psi(Z; \theta^*, \nua^*, \nub^*) \|^2  ] + \hP[\|\psi(Z;  \hth, \nua^*, \nub^*)-\psi(Z; \theta^*, \nua^*, \nub^*)\|^2] \\
& \lesssim N^{-1/2}\log N\{1+N^{-1/2+2/q}\}+ \|\hat \theta-\theta^{*}\|^{\beta}_2= N^{-1/2}\log N\{1+N^{-1/2+2/q}\} +N^{-\beta/2}. 
\end{align*}
In the last inequality, we use Lemma 6.2 \citep{ChernozhukovVictor2018Dmlf}. Here, the envelops exists since 
\begin{align*}
    \|\psi(Z;  \theta, \nua^*, \nub^*)-\psi(Z; \theta^*, \nua^*, \nub^*) \|^2\leq CF^2_{\eta^{*},\theta^{*}}. 
\end{align*}
for some constant $C$ depending on $d_1,d_2$. According to condition \ref{assump: identification: metric entropy} in \cref{assump: identification}, it satisfies the moment condition $\|F^2_{\eta^{*},\theta^{*}}\|_{\hP,q/2}\leq c_1$. In addition, the metricy entropy assumption is satisfied since 
\begin{align*}
    &\sup_{\mathbb Q}\log N(\epsilon\|C\Fcal^2_{1,\eta^{*},\theta^{*}}\|_{\mathbb Q,2},\braces{\|\psi(Z;\theta, \nua^*, \nub^*)-\psi(Z; \theta^*, \nua^*, \nub^*)\|^2: \theta\in\Theta},\|\cdot\|_{\mathbb Q,2})\\
    &\lesssim \sup_{\mathbb Q}\log  \{N(\epsilon\|\Fcal_{1,\eta^{*},\theta^{*}}\|_{\mathbb Q,2},\Fcal_{1,\eta,\theta'_1},\|\cdot\|_{\mathbb Q,2})\}^2 \lesssim v\log (a/\epsilon).
\end{align*}

Similarly, with $P_N$-probability probability $1-o(1)$, the second term of \cref{eq:r_N2} can be  upper bounded as follows:
\begin{align*}
&\hP_{N,k}[\|\psi(Z;   \hat \theta, \hnua, \hnub  )-\psi(Z;  \hat \theta, \nua^*, \nub^*) \|^2  ]\\
 &= \frac{1}{\sqrt{N}}\hG_{N,k}[\|\psi(Z;   \hat \theta, \hnua, \hnub  )-\psi(Z;  \hat \theta, \nua^*, \nub^*) \|^2  ]+\hP[\|\psi(Z;   \hat \theta, \hnua, \hnub  )-\psi(Z;  \hat \theta, \nua^*, \nub^*) \|^2  ]\\
 &\leq  \sup_{\theta \in  \Theta}\frac{1}{\sqrt{N}}\hG_{N,k}[\|\psi(Z;   \theta, \hnua, \hnub  )-\psi(Z;  \theta, \nua^*, \nub^*) \|^2  ]+ \sup_{\theta \in  \mathcal{B}(\theta^{*};\tau_N )} \hP[\|\psi(Z;  \theta, \hnua, \hnub  )-\psi(Z;  \theta, \nua^*, \nub^*) \|^2  ]\\
 &\lesssim   N^{-1/2}\log N\{1+N^{-1/2+2/q}\} +\{r'_N\}^2. 
\end{align*}
In the last inequality, we use Lemma 6.2 \citep{ChernozhukovVictor2018Dmlf} and \cref{assump: error}. 
In the end, we have 
\begin{align*}
    R_N =O_{P_N}\prns{N^{-1/2+1/q}(\log N)^{1/2}  + N^{-1/4}(\log N)^{1/2}+r'_N}+N^{-\beta/4}.  
\end{align*}
This concludes the proof. 
\end{proof}

\subsection{Proofs for \texorpdfstring{\cref{sec: bandit}}{a}}
\begin{proof}[Proof for \cref{thm: causal}]
In this part, we prove the asymptotic distribution of our estimators 
corresponding to the general estimating equation \cref{eq: est-eq-causal}. We prove this by verifying all conditions in the assumptions for \cref{thm: general-split}.

\textbf{Verifying \cref{assump: jacob}. } 
\begin{align*}
    &J^*  = \partial_{\theta}\{\hP\left[\psi(Z; \theta, \tnua(Z; \theta_1), \tnub(Z))\right]\}\vert_{\theta = \tth} \\
    =& \partial_{\theta}\hP\bigg\{\frac{\ind(T = t)}{\tpib(t \mid X)}U(Y; \theta_1) - \frac{\ind(T = t) - \tpib(t \mid X)}{\tpib(t \mid X)} \tq(X, t; \theta_1) + V(\theta_2)\bigg\}\vert_{\theta = \theta^*} \\
    =& \partial_{\theta}\hP\bigg\{\frac{\ind(T = t)}{\tpib(t \mid X)}U(Y; \theta_1) + V(\theta_2)\bigg\}\vert_{\theta = \theta^*} \\
    =& \partial_{\theta}\hP\bigg\{\frac{\ind(T = t)}{\tpib(t \mid X)}U(Y; \theta_1) - \frac{\ind(T = t) - \tpib(t \mid X)}{\tpib(t \mid X)} \tq(X, t; \tth_1) + V(\theta_2)\bigg\}\vert_{\theta = \theta^*} \\
    =& \partial_{\theta}\{\hP\left[\psi(Z; \theta, \tnua(Z; \theta_1^*), \tnub(Z))\right]\}\vert_{\theta = \tth},
\end{align*}
where the second and third equality follow because $\hP\left[ \frac{\ind(T = t) - \tpib(t \mid X)}{\tpib(t \mid X)} \tq(X, t; \theta_1)\right] = 0$.

\textbf{Verifying \cref{assump: identification}. } We first verify conditions \ref{assump: identification: identification} and \ref{assump: identification: conditioning} in \cref{assump: identification}. We denote that $J_{jk}(\theta) = \partial_{\theta^{(k)}}\hP\left[U_j(Y(t); \theta_1) + V_j(\theta_2)\right]$ where $\theta^{(k)}$ is the $k^{\text{th}}$ component of $\theta = (\theta_1, \theta_2)$. By condition \ref{assump: causal: Jacobian}, $J_{jk}(\theta)$ is Lipschitz continuous at $\tth$ with Lipschitz constant $c'$. So for any $\varepsilon > 0$, if $\theta$ belongs to the open ball $\mathcal{B}(\tth; \epsilon/c')$, then 
\[
|J_{jk}(\theta) - J_{jk}(\tth)| = \left|\partial_{\theta^{(k)}}\hP\left[U_j(Y(t); \theta_1) + V_j(\theta_2)\right] - \partial_{\theta^{(k)}}\hP\left[U_j(Y(t); \tth_1) + V_j(\tth_2)\right]\right| \le \varepsilon.
\] 
By first-order Taylor expansion, for any $\theta \in \mathcal{B}(\tth; \delta)$, there exists $\overline{\theta} \in \mathcal{B}(\tth; \|\theta - \tth\|)$ such that 
\begin{align*}
    \|\hP\left[U(Y(t); \theta_1) + V(\theta_2)\right]\| 
        &= \|J(\overline{\theta})(\theta - \tth)\| \\
        &\ge \|J({\tth})(\theta - \tth)\| - \|(J(\overline{\theta}) - J({\tth}))(\theta - \tth)\| \\
        &\ge \|J({\tth})(\theta - \tth)\| - \varepsilon\sqrt{d}\|\theta - \tth\| \\
        &\ge \|J({\tth})(\theta - \tth)\| - \frac{1}{2}\|J({\tth})(\theta - \tth)\| \\
        &= \frac{1}{2}\|J({\tth})(\theta - \tth)\|,
\end{align*}
where the second last inequality holds if we choose $\varepsilon \le \frac{c_3}{2\sqrt{d}} \le \frac{1}{2\sqrt{d}}\sigma_{\min}(J({\tth}))$, where $\sigma_{\min}(J({\tth}))$ is the smallest singular value of $J({\tth})$. Thus 
\[
    \inf_{\theta \in \mathcal{B}(\tth; \varepsilon/c')}2\|\hP\left[U(Y(t); \theta_1) + V(\theta_2)\right]\|  \ge \|J({\tth})(\theta - \tth)\|.
\]
Moreover, for any $\theta \in \Theta \setminus \mathcal{B}(\tth; \frac{c_3}{2\sqrt{d}c'})$, $2\|\hP\left[U(Y(t); \theta_1) + V(\theta_2)\right]\| \ge c_2$ according to condition \ref{assump: causal: Jacobian}.

Therefore, $$2\|\hP\left[U(Y(t); \theta_1) + V(\theta_2)\right]\| \ge J^*(\theta - \tth) \wedge c_2$$
 where 
 $$J^* = J(\tth) = \partial_{\theta}\hP\left[U(Y(t); \theta_1) + V(\theta_2)\right]\vert_{\theta = \tth}.$$ 
 Moreover, the singular values $J^*$ are bounded between $c_3, c_4$ according to condition \ref{assump: causal: identification}.

We then verify condition \ref{assump: identification: orthogoanlity} in \cref{assump: identification}: for any $(\nua(\cdot; \theta_1'), \nub) \in \mathcal{T}_{N}$, 
\begin{align*}
      &\partial_{r}\big\{\hP\left[\psi(Z; \tth, \nua(Z; \tth_1) + r(\nua(\cdot; \theta_1') - \tnua(Z; \tth_1)), \tnub(Z))\right]\big\}\vert_{r = 0} \\
      =& \partial_{r}\hP\bigg\{\frac{\ind(T = t) - \tpib(t \mid X)}{\tpib(t \mid X)}r\big(\mu(X, T; \theta_1') - \tq(X, T; \tth_1) \big)\bigg\}\vert_{r = 0} = 0. \\
      & \partial_{r}\big\{\hP\left[\psi(Z; \tth, \tnua(Z; \tth_1)) , \tnub(Z) + r(\nub(Z) - \tnub(Z)))\right]\big\}\vert_{r = 0} \\
      =& \partial_r \hP\bigg\{\frac{\ind(T = t)}{\tpib(t \mid X) + r(\pi(t \mid X) - \tpib(t \mid X))}\big(U(Y; \tth_1) - \expect[U(Y; \tth_1) \mid X, T]\big)\bigg\}\vert_{r = 0} = 0.
\end{align*}

\textbf{Verifying \cref{assump: error}.}
We take $\mathcal{T}_N$ to be the set that contains all $(\mu(\cdot, \theta_1'), \pi(\cdot))$ that satisfies the following conditions: 
\begin{align*}
    &\left\|\bigg\{\hP \left[\mu\left(X, T; \theta_1'\right) - \tq\left(X, T; \theta_1'\right) \right]^2\bigg\}^{1/2}\right\|   \le \rho_{\mu, N},  \\
    &\bigg\{\hP \left[\pi(T \mid X) - \tpib(T \mid X)\right]^2\bigg\}^{1/2} \le \rho_{\pi, N}, ~~ \|\theta_1' - \theta^*_1\| \le  \rho_{\theta, N},
\end{align*}   
with $\rho_{\pi, N}(\rho_{\mu, N} + C\rho_{\theta, n}) \le \frac{\varepsilon_\pi^3}{3}\delta_N N^{-1/2}$, $\rho_{\pi, N} \le \frac{\delta_N^3}{\log N}$, and $\rho_{\mu, N} + C\rho_{\theta, N} \le \frac{\delta^2_N}{\log N}$.

Then \cref{assump: nuisance-rate} and condition \ref{assump: causal: rate} in  \cref{thm: causal} guarantee that the nuisance estimates $(\hq(, \hthinit), \hpib) \in \mathcal{T}_N$ with probability, namely, condition \ref{assump: error: nuisance-set} in \cref{assump: error} is satisfied. 

Before verifying other conditions, first note that the condition \ref{assump: causal: boundedness2} states that  
\[
  \big\{\hP \left[\tq(X,T; \theta_1) - \tq(X,T; \tth_1)\right]^2\big\}^{1/2} \le C\|\theta_1 - \tth_1 \|, ~~~ \forall \|\theta_1  - \tth_1 \| \le  \rho_{\theta, N},
\]
 which implies that for any $(\mu(\cdot, \theta_1'), \pi(\cdot)) \in \mathcal T_N$,
\begin{align*}
    &\qquad\qquad\qquad\qquad\qquad \left\|\bigg\{\hP \left[\mu(X, T; \theta_1') - \tq(X, T; \tth_1)\right]^2\bigg\}^{1/2}\right\| \\
    &\le \left\|\bigg\{\hP \left[\mu(X, T; \theta_1') - \tq(X, T; \theta_1')\right]^2\bigg\}^{1/2}\right\| + \left\|\bigg\{\hP \left[ \tq(X, T; \theta_1') - \tq(X, T; \tth_1)\right]^2\bigg\}^{1/2}\right\| \\
    &= \rho_{\mu, N} + C\rho_{\theta, N}.
\end{align*}

Now we verify the condition on $r_N$: for any $(\nua(\cdot; \theta_1'), \nub(\cdot)) = (\mu(\cdot, \theta_1'), \pi(\cdot)) \in \mathcal{T}_N$,  and $\theta \in \Theta$, 
\begin{align*}
   & \|\hP\left[\psi(Z; \theta, \nua(Z; \theta_1'), \nub(Z))\right] - \hP\left[\psi(Z; \theta, \tnua(Z; \tth_1), \tnub(Z))\right]\| \\
   \le & \|\hP\big(\frac{\ind(T  = t)}{\pi(t \mid X)} - \frac{\ind(T  = t)}{\tpib(t \mid X)}\big)\big(U(Y; \theta_1) - \tq(X, T; \tth_1)\big)\| \\
   &+ \|\hP\big(\frac{\ind(T  = t)}{\pi(t \mid X)} - \frac{\ind(T  = t)}{\tpib(t \mid X)}\big)\big( \tq(X, T; \tth_1) - \mu(X, T; \theta_1')\big)\| \\
   +& \|\hP\frac{\ind(T = t) - \tpib(t \mid X)}{\tpib(t \mid X)}[\tq(X, T; \tth_1) - \mu(X, T; \theta_1')]\| \\
   \le& \frac{1}{\varepsilon_\pi}\bigg\{\hP \left[\pi(t \mid X) - \tpib(t \mid X)\right]^2\bigg\}^{1/2}\left\|\bigg\{\hP \left[\tq(X, T; \theta_1) - \tq(X, T; \tth_1)\right]^2\bigg\}^{1/2}\right\| \\
   +&\frac{1}{\varepsilon_\pi} \bigg\{\hP \left[\pi(t \mid X) - \tpib(t \mid X)\right]^2\bigg\}^{1/2}\left\|\bigg\{\hP \left[\mu(X, T; \theta_1') - \tq(X, T; \tth_1)\right]^2\bigg\}^{1/2}\right\| \\
   \le& \frac{1}{\varepsilon_\pi}\rho_{\pi, N} \times (2C\sqrt{d} + \rho_{\mu, N} + C\rho_{\theta, N}) \le \frac{4C}{\varepsilon_\pi}\sqrt{d}\rho_{\pi, N}.
\end{align*} 
Thus, the condition on $r_N$ is satisfied with  $\tau_N$ such that $\tau_N = \frac{4C\sqrt{d}\rho_{\pi, N}}{\delta_N\varepsilon_\pi}$. 

Next, we verify the condition on $r_N'$: for any $\theta$ such that $\|\theta - \tth\| \le \frac{4C\sqrt{d}\rho_{\pi, N}}{\delta_N\varepsilon_\pi}$, and any $(\nua(\cdot; \theta_1'), \nub(\cdot)) = (\mu(\cdot, \theta_1'), \pi(\cdot)) \in \mathcal{T}_N$,
\begin{align*}
    &\left\|\left\{
    \hP\left[\psi(Z; \theta, \nua(Z; \theta_1'), \nub(Z)) - \psi(Z; \theta, \tnua(Z; \tth_1), \tnub(Z))\right]^2\right\}^{1/2}\right\| \\
    \le &
     \left\|\begin{bmatrix}
     \left\{\hP\left[\big(\frac{\ind(T  = t)}{\pi(t \mid X)} - \frac{\ind(T  = t)}{\tpib(t \mid X)}\big)\big(\tq_1(X, T; \theta_1) - \tq_1(X, T; \tth_1)\big)\right]^2\right\}^{1/2} \\ 
     \vdots \\
     \left\{\hP\left[\big(\frac{\ind(T  = t)}{\pi(t \mid X)} - \frac{\ind(T  = t)}{\tpib(t \mid X)}\big)\big(\tq_d(X, T; \theta_1) - \tq_d(X, T; \tth_1)\big)\right]^2\right\}^{1/2}
     \end{bmatrix}
     \right\| \\
     +& \left\|\begin{bmatrix}
     \left\{\hP\left[\big(\frac{\ind(T  = t)}{\pi(t \mid X)} - \frac{\ind(T  = t)}{\tpib(t \mid X)}\big)\big(\tq_1(X, T; \tth_1) - \mu_1(X, T; \theta_1')\big)\right]^2\right\}^{1/2} \\ 
     \vdots \\
     \left\{\hP\left[\big(\frac{\ind(T  = t)}{\pi(t \mid X)} - \frac{\ind(T  = t)}{\tpib(t \mid X)}\big)\big(\tq_d(X, T; \tth_1) - \mu_d(X, T; \theta_1')\big)\right]^2\right\}^{1/2}
     \end{bmatrix}
     \right\| \\
     +& \left\|\begin{bmatrix}
     \left\{\hP\left[\frac{\ind(T = t) - \tpib(t \mid X)}{\tpib(t \mid X)}\big(\tq_1(X, T; \tth_1) - \mu_1(X, T; \theta_1')\big)\right]^2\right\}^{1/2} \\ 
     \vdots \\
     \left\{\hP\left[\frac{\ind(T = t) - \tpib(t \mid X)}{\tpib(t \mid X)}\big(\tq_d(X, T; \tth_1) - \mu_d(X, T; \theta_1')\big)\right]^2\right\}^{1/2}
     \end{bmatrix}
     \right\| \\
     \le& \frac{4C^2\sqrt{d}\rho_{\pi, N}}{\delta_N\varepsilon^2_\pi} + \frac{1}{\varepsilon_{\pi}}\left(\rho_{\mu, N} + C\rho_{\theta, N}\right) + \frac{1}{\varepsilon_{\pi}}\left(\rho_{\mu, N} + C\rho_{\theta, N}\right)
\end{align*}
So when $\rho_{\pi, N} \le \frac{\delta_N^3}{\log N}$, and $\rho_{\mu, N} + C\rho_{\theta, N} \le \frac{\delta^2_N}{\log N}$, $r_N' = \frac{\delta_N^2}{\varepsilon^2_\pi\log N}\left(4C^2\sqrt{d} + 2\varepsilon_\pi\right) \le \frac{\delta_N}{\log N}$ if $\delta_N \le \frac{\varepsilon_\pi^2}{4C^2\sqrt{d} + 2\varepsilon_\pi}$. 

Finally, to verify the condition on $\lambda_N'$, we note that for any $\theta$ such that $\|\theta - \tth\| \le \frac{4C\sqrt{d}\rho_{\pi, N}}{\delta_N\varepsilon_\pi}$, and any $(\nua(\cdot; \theta_1'), \nub(\cdot)) = (\mu(\cdot, \theta_1'), \pi(\cdot)) \in \mathcal{T}_N$
\begin{align*}
    &\qquad\qquad\qquad\qquad\qquad\qquad f(r; \theta, \nua(Z; \theta_1'), \nub) \\
        &= \hP\bigg\{\frac{\ind(T = t)}{\tpib(T \mid X) + r(\pi(T \mid X) - \tpib(T \mid X))}\big[\tq(X, T; \tth_1 + r(\theta_1 - \tth_1))
    - \tq(X, T; \tth_1) \\
        &- r\big(\mu(X, T; \theta_1') - \tq(X, T; \tth_1)\big)\big] + \big
        [\tq(X, t; \tth_1) + r\big(\mu(X, t; \theta_1') - \tq(X, t; \tth_1)\big) \big] + V(\tth_2 + r({\theta}_2 - \tth_2))\bigg\}
\end{align*}
Thus the first-order derivative is 
\begin{align*}
    &\partial_{r}f(r; \theta, \nua(Z; \theta_1'), \nub) \\
    =& -\hP\bigg\{\frac{\ind(T = t)}{\big(\tpib(T \mid X) + r(\pi(T \mid X) - \tpib(T \mid X))\big)^2}\big(\pi(T \mid X) - \tpib(T \mid X)\big)\big[\tq(X, T; \tth_1 + r(\theta_1 - \tth_1)) \\
    -& \tq(X, T; \tth_1) - r\big(\mu(X, T; \theta_1') - \tq(X, T; \tth_1)\big)\big]\bigg\} + \hP\bigg\{\frac{\ind(T = t)}{\tpib(T \mid X) + r(\pi(T \mid X) - \tpib(T \mid X))} \\
    \times& \partial_{\overline\theta_1^\top}\tq(X, T; \overline\theta_1)\vert_{\overline\theta_1 = \tth_1 + r(\theta_1 - \tth_1)}(\theta_1 - \tth_1)\bigg\} - \hP\bigg\{\frac{\ind(T = t)}{\tpib(T \mid X) + r(\pi(T \mid X) - \tpib(T \mid X))}\\
    \times& \big[\mu(X, T; \theta_1') - \tq(X, T; \tth_1)\big]\bigg\} + \hP\bigg\{\big[\mu(X, t; \theta_1') - \tq(X, t; \tth_1)\big]\bigg\} +  \partial_{\overline\theta_2^\top}V(\overline\theta_2)\vert_{\overline\theta_2 = \tth_2 + r({\theta}_2 - \tth_2)}({\theta}_2 - \tth_2).
\end{align*}
The second order derivative is 
\begin{align*}
    &\partial^2_{r}f(r; \theta, \nua(Z; \theta_1'), \nub) \\
    =& \hP\bigg\{\frac{2\ind(T = t)}{\big(\tpib(T \mid X) + r(\pi(T \mid X) - \tpib(T \mid X))\big)^3}\big(\pi(T \mid X) - \tpib(T \mid X)\big)^2\big[\tq(X, T; \tth_1 + r(\theta_1 - \tth_1)) \\
    -& \tq(X, T; \tth_1) - r\big(\mu(X, T; \theta_1') - \tq(X, T; \tth_1)\big)\big]\bigg\} - \hP\bigg\{\frac{\ind(T = t)}{\big(\tpib(T \mid X) + r(\pi(T \mid X) - \tpib(T \mid X))\big)^2}\\
    \times & \big(\pi(T \mid X) - \tpib(T \mid X)\big)\partial_{\overline\theta_1^\top}\tq(X, T; \overline\theta_1)\vert_{\overline\theta_1 = \tth_1 + r(\theta_1 - \tth_1)}(\theta_1 - \tth_1)\bigg\}\\
    +&\hP\bigg\{\frac{\ind(T = t)}{\big(\tpib(T \mid X) + r(\pi(T \mid X) - \tpib(T \mid X))\big)^2}\big(\pi(T \mid X) - \tpib(T \mid X)\big)\big[\mu(X, T; \theta_1') - \tq(X, T; \tth_1)\big]
   \bigg\} \\
   +& \hP\bigg\{\frac{\ind(T = t)}{\tpib(T \mid X) + r(\pi(T \mid X) - \tpib(T \mid X))}\diag\big[(\theta_1 - \tth_1)^\top\big]\big[\partial^2_{\overline\theta_1, \overline\theta_1^\top}\tq(X, T;\overline\theta_1)\vert_{\overline\theta_1 = \tth_1 + r(\theta_1 - \tth_1)}\big](\theta_1 - \tth_1)\bigg\} \\
   -& \hP\bigg\{\frac{\ind(T = t)\big(\pi(T \mid X) - \tpib(T \mid X)\big)}{\big(\tpib(T \mid X) + r(\pi(T \mid X) - \tpib(T \mid X))\big)^2}\partial_{\overline\theta_1^\top}\tq(X, T; \overline\theta_1)\vert_{\overline\theta_1 = \tth_1 + r(\theta_1 - \tth_1)}(\theta_1 - \tth_1)\bigg\} \\
   +& \hP\bigg\{\frac{\ind(T = t)}{\big(\tpib(T \mid X) + r(\pi(T \mid X) - \tpib(T \mid X))\big)^2}\big(\pi(T \mid X) - \tpib(T \mid X)\big)\big[\mu(X, T; \theta_1') - \tq(X, T; \tth_1)\big] \\
   &+ \diag({\theta}_2 - \tth_2)^\top\partial^2_{\overline\theta_2, \overline\theta_2^\top}V(\overline\theta_2)\vert_{\overline\theta_2 = {\theta}_2 + r({\theta}_2 - \tth_2)}({\theta}_2 - \tth_2)
\end{align*}
Above, we use 
condition \ref{assump: causal: exchange-int-diff} in \cref{thm: causal} to ensure exchange of integration and differentiation so we can get terms $\partial_{\overline\theta_1^\top}\tq(X, T; \overline\theta_1)\vert_{\overline\theta_1 = \tth_1 + r(\theta_1 - \tth_1)}$ and $\partial^2_{\overline\theta_1, \overline\theta_1^\top}\tq(X, T; \overline\theta_1)\vert_{\overline\theta_1 = \tth_1 + r(\theta_1 - \tth_1)}$.

Note that  
\begin{align*}
    &\left\|\hP\left[\big(\pi(T \mid X) - \tpib(T \mid X)\big)\partial_{\overline\theta_1^\top}\tq(X, T; \overline\theta_1)\vert_{\overline\theta_1 = \tth_1 + r(\theta_1 - \tth_1)}(\theta_1 - \tth_1)\right]\right\|\\
    &= \left\| \begin{bmatrix}
    \hP\left[\big(\pi(T \mid X) - \tpib(T \mid X)\big)\partial_{\overline\theta_1^\top}\expect[U_1(Z; \overline\theta_1) \mid X, T]\vert_{\overline\theta_1 = \tth_1 + r(\theta_1 - \tth_1)}(\theta_1 - \tth_1)\right] \\
    \vdots \\
    \hP\left[\big(\pi(T \mid X) - \tpib(T \mid X)\big)\partial_{\overline\theta_1^\top}\expect[U_d(Z; \overline\theta_1) \mid X, T]\vert_{\overline\theta_1 = \tth_1 + r(\theta_1 - \tth_1)}(\theta_1 - \tth_1)\right]
    \end{bmatrix}\right\| \\
    &\le \big\{\hP\left[\big(\pi(T \mid X) - \tpib(T \mid X)\big)\right]^2\big\}^{1/2}\times \sqrt{d}\sup_{j, \|\theta_1 - \tth_1\| \le \frac{4C\sqrt{d}\rho_{\pi, N}}{\delta_N\varepsilon_\pi}}\left\|\hP\left\{\left[\partial_{\overline\theta_1}\tq_j(X, t; \overline\theta_1)\right]^2\right\}^{1/2}\right\|\times\|\theta_1 - \tth_1\| \\
    &\le C\sqrt{d}\rho_{\pi, N}\|\theta_1 - \tth_1\|
\end{align*}
\begin{align*}
    &\left\|\hP\left[\diag\big[(\theta_1 - \tth_1)^\top\big]\big[\partial^2_{\overline\theta, \overline\theta^\top}\tq(X, T; \overline\theta)\big]\vert_{\overline\theta_1 = \tth_1 + r(\theta_1 - \tth_1)}(\theta_1 - \tth_1) \right]\right\|\\
    =& \left\|\begin{bmatrix}(\theta_1 - \tth_1)^\top & 0 & \hdots & 0 \\
    0 & (\theta_1 - \tth_1)^\top & \hdots & 0 \\
    \vdots & \vdots & \ddots & \vdots \\
    0 & 0 &\hdots & (\theta_1 - \tth_1)^\top\end{bmatrix}\begin{bmatrix}
    \hP\left[\partial_{\overline\theta_1}\partial_{\overline\theta_1^\top}\tq_1(X, T; \overline\theta_1)\right]\vert_{\overline\theta_1 = \tth_1 + r(\theta_1 - \tth_1)} \\
    \vdots \\
    \hP\left[\partial_{\overline\theta_1}\partial_{\overline\theta_1^\top}\tq_d(X, T; \overline\theta_1)\right]\vert_{\overline\theta_1 = \tth_1 + r(\theta_1 - \tth_1)}
    \end{bmatrix}(\theta_1 - \tth_1)\right\|,  \\
    =& \left\|\begin{bmatrix}
    (\theta_1 - \tth_1)^\top\hP\left[\partial_{\overline\theta_1}\partial_{\overline\theta_1^\top}\tq_1(X, T; \overline\theta_1)\right]\vert_{\overline\theta_1 = \tth_1 + r(\theta_1 - \tth_1)}(\theta_1 - \tth_1) \\
    \vdots \\
    (\theta_1 - \tth_1)^\top\hP\left[\partial_{\overline\theta_1}\partial_{\overline\theta_1^\top}\tq_d(X, T; \overline\theta_1)\right]\vert_{\overline\theta_1 = \tth_1 + r(\theta_1 - \tth_1)}(\theta_1 - \tth_1)
    \end{bmatrix}\right\| \\
    \le& \sqrt{d}\|\theta_1 - \tth_1\|^2\sup_{j, \|\theta - \tth\| \le  \frac{4C\sqrt{d}\rho_{\pi, N}}{\delta_N\varepsilon_\pi}}\|\hP\left[\partial_{\overline\theta_1}\partial_{\overline\theta_1^\top}\tq_j(X, T; \overline\theta_1)\right]\| \le C\sqrt{d}\|\theta_1 - \tth_1\|^2,
\end{align*}
and   
\begin{align*}
    \sup_{r \in (0, 1)}\left\|\bigg\{\hP \left[\tq(X, T; \tth_1 + r(\theta_1 - \tth_1)) - \tq(X, T; \tth_1)\right]^2\bigg\}^{1/2}\right\| \le C\sqrt{d}\|\theta_1 - \tth_1\|.
\end{align*}

Thus for any $\theta$ such that $\|\theta - \tth\| \le  \frac{4C\sqrt{d}\rho_{\pi, N}}{\delta_N\varepsilon_\pi}$, 
\begin{align}
 &\|\partial^2_{r}f(r; \theta, \mu(X, T; \theta_1'), \pi)\| \\  
\le& \frac{\rho_{\pi, N}}{\varepsilon^3_\pi}\left[C\sqrt{d}\|\theta_1 - \tth_1\| + \rho_{\mu, N} + C\rho_{\theta, N}\right] \nonumber \\
+& \frac{C\sqrt{d}}{\varepsilon_\pi^2}\rho_{\pi, N}\|\theta_1 - \tth_1\| + \frac{1}{\varepsilon_\pi^2}\rho_{\pi, N}(\rho_{\mu, N} + C\rho_{\theta, N}) + \frac{C\sqrt{d}}{\varepsilon_\pi}\|\theta_1 - \tth_1\|^2 \nonumber \\
+& \frac{C\sqrt{d}}{\varepsilon_\pi^2}\rho_{\pi, N}\|\theta_1 - \tth_1\| + \frac{1}{\varepsilon_\pi^2}\rho_{\pi, N}(\rho_{\mu, N} + C\rho_{\theta, N}) + C\|\theta_2 - \tth_2\|^2 \nonumber \\
=& \frac{3}{\varepsilon_\pi^3}\rho_{\pi, N}(\rho_{\mu, N} + C\rho_{\theta, n}) + \frac{C\sqrt{d}}{\varepsilon_\pi}\|\theta - \tth\|^2 + \frac{C\sqrt{d}}{\varepsilon_\pi^3}\rho_{\pi, N}\|\theta_1 - \tth_1\| \nonumber \\
\le& \frac{3}{\varepsilon_\pi^3}\rho_{\pi, N}(\rho_{\mu, N} + C\rho_{\theta, n}) + \frac{4C^2{d}}{\varepsilon_\pi^2 \delta_N}\rho_{\pi, N}\|\theta - \tth\| + \frac{C\sqrt{d}}{\varepsilon_\pi^3}\rho_{\pi, N}\|\theta_1 - \tth_1\| \label{eq: thm2eq1}
\end{align} 

Given $\rho_{\pi, N} \le \frac{\delta_N^3}{\log N}$, when $\frac{\delta_N}{\log N} \le \frac{\varepsilon_\pi^2}{8C^2 d}$ and $\frac{\delta_N^2}{\log N} \le \frac{\varepsilon_\pi^3}{2C\sqrt{d}}$, $\frac{4C^2{d}}{\varepsilon_\pi^2 \delta_N}\rho_{\pi, N}\|\theta - \tth\| + \frac{C\sqrt{d}}{\varepsilon_\pi^3}\rho_{\pi, N}\|\theta_1 - \tth_1\| \le \delta_N\|\theta - \tth\|$. Moreover, when $\rho_{\pi, N}(\rho_{\mu, N} + C\rho_{\theta, n}) \le \frac{\varepsilon_\pi^3}{3}\delta_N N^{-1/2}$,  $\frac{3}{\varepsilon_\pi^3}\rho_{\pi, N}(\rho_{\mu, N} + C\rho_{\theta, n}) \le \delta_N N^{-1/2}$. Consequently, $\|\partial^2_{r}f(r; \theta, \mu(X, T; \theta_1'), \pi)\| \le \delta_N(\|\theta - \tth\| + N^{-1/2})$.
\end{proof}

\begin{proof}[Proof for \texorpdfstring{\cref{thm: quantile only}}{a}]
First note that 
\begin{align*}
\hP \left[U(Y(t); \theta_1) + V(\theta_2)\right] = 
\begin{bmatrix}
F_t(\theta_1) - \gamma \\
\theta_1 + \frac{1}{1- \gamma}\expect[Y(t) - \theta_1]^+ - \theta_2
\end{bmatrix}.
\end{align*}
When $F_t(\theta_1)$ is differentiable, $\hP \left[U(Y(t); \theta_1) + V(\theta_2)\right]$ is also differentiable by Leibnitz integral  rule, with derivative 
\[
    J(\theta) = \begin{bmatrix} f_{t}(\theta_1) & 0  \\ \frac{F_{t}(\theta_1) - \gamma}{1 - \gamma} & -1 
    \end{bmatrix}.
\]
Thus 
\[
J^* = \begin{bmatrix} f_{t}(\theta_1^*) & 0  \\ 0 & -1 
    \end{bmatrix}.
\]
Now we prove \Cref{thm: quantile only} by verifying the assumptions in \cref{thm: causal}.

\textbf{Verifying condition i in \cref{thm: causal}.}
We only need to verify  that condition 
  \ref{assump: identification: metric entropy} of \cref{assump: identification} hold.  
Since $\Theta$ is compact, $\{y \mapsto \indic{y \le \theta_1}, \theta \in \Theta\}$, $\{y \mapsto \max\{\theta_1, \frac{1}{1 - \gamma}(y  - \theta_1)\} - \theta_2, \theta \in \Theta\}$  are obviously Donsker classes, so condition \ref{assump: identification: metric entropy}  of \cref{assump: identification} is satisfied.

\textbf{Verifying conditions \ref{assump: causal: Jacobian} and \ref{assump: causal: identification} in \cref{thm: causal}.} 
It is straightforward to show that $J(\theta)$ is invertible with the following matrix as its inverse: 
\[
  J^{-1}(\theta) = \begin{bmatrix} \frac{1}{f_{t}(\theta_1)} & 0  \\ 
  - \frac{F_t(\theta_1) - \gamma}{f_t(\theta_1)(1 - \gamma)}
   & -1 
    \end{bmatrix}.
\]
Note that $\sigma_{\max}(J(\theta^*)) \le 2\max\{f_t(\theta_1^*), 1\} \le 2\max\{c_2', 1\}$ and $$\sigma_{\min}(J(\theta^*)) = 1/\sigma_{\max}(J^{-1}(\theta^*)) \ge \min\{\frac{f_t(\theta^*_1)}{2}, \frac{(1 - \gamma)f_t(\theta^*_1)}{2\gamma}, \frac{1}{2}\} \ge \frac{1}{2}\min\{1, \frac{1 - \gamma}{\gamma}c_1', c_1'\}.$$ Thus condition \ref{assump: causal: identification} in \cref{thm: causal} is satisfied with $c_3 =\frac{1}{2}\min\{1, \frac{1 - \gamma}{\gamma}c_1', c_1'\}$ and $c_4 = 2\max\{1, c_2'\}$.
When we estimate quantile only, then only $f_t\prns{\theta_1}$ in $J\prns{\theta}$ matters. Then condition \ref{assump: causal: identification} in \cref{thm: causal} is satisfied with $c_3 = c_1'$ and $c_4 = 2c_2'$.

Since $f_t(\theta_1) \le c_2'$ and $\dot{f}_t(\theta_1) \le c_3'$, it follows that each element in $J(\theta)$ is Lipschtiz continuous at $\theta^*$ with Lipschitz constant $c' = \max\{c_2', c_3'\}$. 
Moreover, for $\theta \in \Theta$ such that $\|\theta-\theta^*\| \ge \frac{c_3}{2\sqrt{2}c'}$, we have $2\|\hP \left[U(Y(t); \overline\theta_1) + V(\overline\theta_2)\right]\|  \ge c_5'$. This means that condition \ref{assump: causal: Jacobian} in \cref{thm: causal} is satisfied with $c' = \max\{c_2', c_3'\}$ and $c_2 = c_5'$.
When we estimate quantile only, we only require $|F(\theta^*_1) - F(\theta_1)| \ge c_4'$ for $|\theta_1 - \tth_1| \ge \frac{c_1'}{2c_3'}$. Then condition 
\ref{assump: causal: Jacobian} in \cref{thm: causal} is satisfied with $c' = c_3'$ and $c_2 = 2c_4'$.

\textbf{Verifying condition \ref{assump: causal: exchange-int-diff} in \cref{thm: causal}.} This condition can be verified by the following facts: for any $\theta_1$ such that $\left|\theta_1 - \tth_1\right|\le  \frac{4C\sqrt{d}\rho_{\pi, N}}{\delta_N\varepsilon_\pi}$, 
\begin{align*}
&\left|\partial_{r}\tq_1(X,t; \tth_1 + r(\theta_1 - \tth_1))\right|  = \left|\partial_{r}\{F_t(\tth_1+r(\theta_1 - \tth_1) \mid X) - \gamma\}\right|  \\
=& \left|f_t(\tth_1 + r(\theta_1 - \tth_1) \mid X)\right|\left|\theta_1 - \tth_1\right| \le C\left|\theta_1 - \tth_1\right|  \\
&\left|\partial^2_{r}\tq_1(X,t; \tth_1 + r(\theta_1 - \tth_1))\right|  = \left|\dot{f}_t(\tth_1 + r(\theta_1 - \tth_1) \mid X)\right|\left|\theta_1 - \tth_1\right|^2 \le C\left|\theta_1 - \tth_1\right|^2
\end{align*}
and 
\begin{align*}
\left|\partial_{r}\tq_2(X,t; \tth_1 + r(\theta_1 - \tth_1))\right| 
  &= \left|\partial_{r} \left\{\tth_1 + r(\theta_1 - \tth_1) + \frac{1}{1 - \gamma}\expect\left[\max(Y - \tth_1 - r(\theta_1 - \tth_1), 0) \mid X, T = t\right]\right\}\right| \\
  &= \left|\theta_1 - \tth_1\right|\left|1 - \frac{1}{1 - \gamma}\left(1 - F_t(\tth_1 + r(\theta_1 - \tth_1) \mid X)\right)\right| \le \left|\theta_1 - \tth_1\right| \\
\left|\partial^2_{r}\tq_2(X,t; \tth_1 + r(\theta_1 - \tth_1))\right| 
  &= \left|\theta_1 - \tth_1\right|^2\left|f_t(\tth_1 + r(\theta_1 - \tth_1) \mid X)\right| \le C\left|\theta_1 - \tth_1\right|^2.
\end{align*}  

\textbf{Verifying conditions \ref{assump: causal: boundedness} and \ref{assump: causal: boundedness2} in \cref{thm: causal}. } For any $(\theta_1, \theta_2) \in \Theta$, 
\begin{align*}
\left\{\hP \left[\tq_1(X,t; \theta_1)\right]^2\right\}^{1/2} 
    &= \left|F_t(\theta_1 \mid X) - \gamma\right| \le 1 \\
\left\{\hP \left[\tq_2(X,t; \theta_1)\right]^2\right\}^{1/2}
    &= \left\{\hP\left[\expect[\max(Y(t) - \theta_1, 0) \mid X]\right]^2\right\}^{1/2}\le C. 
\end{align*}

By first-order Taylor expansion, for any $\theta_1$ such that $\left|\theta_1 - \tth_1\right|\le  \max\{\frac{4C\sqrt{d}\rho_{\pi, N}}{\delta_N\varepsilon_\pi}, \rho_{\theta, N}\}$, there exists $\tilde{\theta}_1$ between $\theta_1$ and $\tth_1$ such that 
\begin{align*}
\left\{\hP \left[\tq_1(X,t; \theta_1) - \tq_1(X,t; \tth_1)\right]^2\right\}^{1/2} = \left\{\hP \left[(\theta_1 - \tth_1)f_t(\tilde{\theta}_1 \mid X) \right]^2\right\}^{1/2} \le C\left|\theta_1 - \tth_1\right| \\
\left\{\hP \left[\tq_2(X,t; \theta_1) - \tq_2(X,t; \tth_1)\right]^2\right\}^{1/2} = \left\{\hP \left[(\theta_1 - \tth_1)(F_{t}(\tilde{\theta}_1 \mid X) - 1) \right]^2\right\}^{1/2} \le \left|\theta_1 - \tth_1\right|.
\end{align*}
Moreover,  for any $\theta_1$ such that $\left|\theta_1 - \tth_1\right|\le  \max\{\frac{4C\sqrt{d}\rho_{\pi, N}}{\delta_N\varepsilon_\pi}, \rho_{\theta, N}\}$, 
\begin{align*}
    \left\{\hP\left[\partial_{\theta_{1}}\tq_1(X, t; \theta_1)\right]^2\right\}^{1/2}  = \left\{\hP\left[f_t(\theta_1 \mid X)\right]^2\right\}^{1/2} &\le C, \\
    \left\{\hP\left[\partial_{\theta_{1}}\tq_2(X, t; \theta_1)\right]^2\right\}^{1/2}  = \left\{\hP\left[F_t(\theta_1 \mid X) -1\right]^2\right\}^{1/2} &\le 1, \\
    \left|\hP\left[\frac{\partial^2}{\partial \theta_1^2}\tq_1(X, t; \theta_1)\right]\right| = \left|\hP \left[\dot{f}_t(\theta_1 \mid X)\right]\right| &\le C,  \\
    \left|\hP\left[\frac{\partial^2}{\partial \theta_1^2}\tq_2(X, t; \theta_1)\right]\right| = \left|\hP \left[{f}_t(\theta_1 \mid X)\right]\right| &\le C, \\
    \left(\partial_{\theta_2}\partial_{\theta_2^\top}V_j(\theta_2)\right) = 0 &\le C. 
\end{align*}
\end{proof}

\begin{proof}[Proof for \texorpdfstring{\cref{thm: IPW}}{a}.]
We follow the proof of \cref{thm: general-split} to consider any sequence of data generating process $P_N \in \mathcal P_N$ but we suppress it for ease of notation.
We prove the conclusion for a generic $k \in \{1, \dots, K\}$. For $l \in \mathcal H_{k, 1}$, we denote $\hP_{N, l}$ and $\hG_{N, l}$ as the empirical average operator and empirical process operator for data in the $\Dcal_l$. 
Throughout the proof, we condition on the event that the convergence rate of propensity score estimator $\hat\pi^{(k, l)}$ in mean squared error is $\rho_{\pi, N}$ and it is lower bounded by $\epsilon_\pi$, which holds with at least probability $1 - \Delta_N$ according to \cref{assump: nuisance-rate}.
In this proof, all notations $\lesssim$ only involve pre-specified constants and not any instance-dependent constants.  

We use the following decomposition analogous to that in Step I of proof for \cref{thm: general-split}.
\begin{align*}
&{\hP}\left[\psi^\text{IPW}(Z; \hat{\theta}\pk_{\text{init}}, \pi^*)\right] \\
  =&\frac{1}{K'}\sum_{l \in \mathcal H_{k, 1}}\left\{{\hP}\left[\psi^\text{IPW}(Z; \hat{\theta}\pk_{\text{init}}, \pi^*)\right] - {\hP}\left[\psi^\text{IPW}(Z; \hat{\theta}\pk_{\text{init}}, \hat\pi^{(k, l)})\right]\right\} \\
  +& \frac{1}{K'}\sum_{l \in \mathcal H_{k, 1}}\left\{{\hP}\left[\psi^\text{IPW}(Z; \hat{\theta}\pk_{\text{init}}, \hat\pi^{(k, l)})\right] - {\hP}_{N, l}\left[\psi^\text{IPW}(Z; \hat{\theta}\pk_{\text{init}}, \hat\pi^{(k, l)})\right]\right\} \\
  +& \frac{1}{K'}\sum_{l \in \mathcal H_{k, 1}}\left\{{\hP}_{N, l}\left[\psi^\text{IPW}(Z; \hat{\theta}\pk_{\text{init}}, \hat\pi^{(k, l)})\right] - {\hP}_{N, l}\left[\psi^\text{IPW}(Z; {\theta}^*, \hat\pi^{(k, l)})\right]\right\} \\
  +& \frac{1}{K'}\sum_{l \in \mathcal H_{k, 1}}\left\{ {\hP}_{N, l}\left[\psi^\text{IPW}(Z; {\theta}^*, \hat\pi^{(k, l)})\right] -  {\hP}\left[\psi^\text{IPW}(Z; {\theta}^*, \hat\pi^{(k, l)})\right]\right\} \\
  +& \frac{1}{K'}\sum_{l \in \mathcal H_{k, 1}}\left\{{\hP}\left[\psi^\text{IPW}(Z; {\theta}^*, \hat\pi^{(k, l)})\right] - {\hP}\left[\psi^\text{IPW}(Z; {\theta}^*, \pi^*)\right]\right\}
\end{align*}
By following the Step I of proof for \cref{thm: general-split}, we can also analogously show that 
\begin{align*}
\left\|{\hP}\left[\psi^\text{IPW}(Z; \hat{\theta}\pk_{\text{init}}, \pi^*)\right]\right\| \le \frac{4}{K'}\sum_{l \in \mathcal H_{k, 1}}\calI'_{1, l} +  \frac{4}{K'}\sum_{l \in \mathcal H_{k, 1}}\calI'_{2, l} + \epsilon_N
\end{align*}
where 
\begin{align*}
\calI'_{1, l}
  &= \sup_{\theta \in \Theta}\left\|{\hP}\left[\psi^\text{IPW}(Z; \theta, \pi^*)\right] - {\hP}\left[\psi^\text{IPW}(Z; \theta, \hat\pi^{(k, l)})\right]\right\| \\
\calI'_{2, l} 
  &= \sup_{\theta \in \Theta}\left\|{\hP}\left[\psi^\text{IPW}(Z; \theta, \hat\pi^{(k, l)})\right] - {\hP}_{N, l}\left[\psi^\text{IPW}(Z; \theta, \hat\pi^{(k, l)})\right]\right\|.
\end{align*}
\textbf{Bounding $\calI'_{1, l}$.} Note that by condition \ref{assump: causal: boundedness} of \cref{thm: causal}, 
\begin{align*}
\calI'_{1, l} =&\left\|\hP\left[\psi^\text{IPW}(Z; \theta, \pi^*) - \psi^\text{IPW}(Z; \theta, \hat\pi^{(k, l)})\right]\right\| \\
=& \sup_{\theta \in \Theta}\left\|{\hP}\left[\frac{\mu^*(X, t; \theta_1)}{\hat{\pi}^{(k, l)}(X)}\left(\hat{\pi}^{(k, l)}(X) - \pi^*(X)\right)\right]\right\|  \\
=& \frac{\sqrt{d}\rho_{\pi, N}}{\epsilon_{\pi}}\max_{j}\sup_{\theta \in \Theta}\left\{\hP \left[\mu^*_j(X, t; \theta_1)\right]^{2}\right\}^{1/2} \\
\le& \frac{C\sqrt{d}\rho_{\pi, N}}{\epsilon_{\pi}}. 
\end{align*}
\textbf{Bounding $\calI'_{2, l}$.} 
Note that 
\begin{align*}
\sqrt{\frac{N}{K'}}\calI'_{2, l} 
  &= \sqrt{\frac{N}{K'}}\sup_{\theta \in \Theta}\left\|{\hP}_{N, l}\left[\psi^\text{IPW}(Z; {\theta}, \hat\pi^{(k, l)}\right] -  {\hP}\left[\psi^\text{IPW}(Z; {\theta}, \hat\pi^{(k, l)}\right]\right\| \\
  &= \sup_{\theta \in \Theta}\left\|{\hG}_{N, l}\left[\psi^\text{IPW}(Z; {\theta}, \hat\pi^{(k, l)}\right]\right\| 
\end{align*}
Given that condition \ref{assump: identification: metric entropy} in \cref{assump: identification} is satisfied for the estimating equation $\psi^\text{IPW}$, we can follow the end of step I in the proof for \cref{thm: general-split} to prove that with $P_N$ probability $1 - c\prns{\log N}^{-1}$ for a constant $c$ that depends on only constants in the assumptions,
\[
  \sup_{\theta \in \Theta}\left\|{\hG}_{N, l}\left[\psi^\text{IPW}(Z; {\theta}, \hat\pi^{(k, l)}\right]\right\|  \lesssim \log\prns{\frac{N}{K'}} + \prns{\frac{N}{K'}}^{-1/2 + 1/q'}\log\prns{\frac{N}{K'}},
\]
so that $\calI'_{2, l} \lesssim \prns{\frac{K'}{N}}^{1/2}\log\prns{\frac{K'}{N}} +  \prns{\frac{K'}{N}}^{1 - \frac{1}{q'}}\log\prns{\frac{K'}{N}} \le\delta_N\rho_{\pi, N} < \rho_{\pi, N}$.

Therefore, with $P_N$-probability $1 - c\prns{\log N}^{-1}$,
\[
{\hP}\left[\psi^\text{IPW}(Z; \hat{\theta}\pk_{\text{init}}, \pi^*)\right] \le \prns{\frac{C\sqrt{d}}{\epsilon_\pi} + 1 }\rho_{\pi, N}.
\]
In the proof of \cref{thm: causal}, we have showed that conditions \ref{assump: causal: Jacobian} and \ref{assump: causal: identification} in \cref{thm: causal} imply that 
\[
  \|J^*(\hat{\theta}\pk_{\text{init}} - \theta^*)\| \wedge c_0  
  \le 
  2\left\|{\hP}\left[\psi^\text{IPW}(Z; \hat{\theta}\pk_{\text{init}}, \pi^*\right]\right\| \le 2\prns{\frac{C\sqrt{d}}{\epsilon_\pi} + 1 }\rho_{\pi, N}.
\]
Therefore, with probability $1 - c\prns{\log N}^{-1}$:
\[
  \rho_{\theta, N}  = \left\|\hat{\theta}\pk_{1, \text{init}} - \theta^*\right\| \le \left\|\hat{\theta}\pk_{\text{init}} - \theta^*\right\| \le \frac{2}{c_3}\prns{\frac{C\sqrt{d}}{\epsilon_\pi} + 1 }\rho_{\pi, N}.
\]
\end{proof}

\subsection{Proofs for Appendix}
\begin{proof}[Proof of \cref{thm: iv}]
 In this part, we prove the asymptotic distribution of our estimator $\hat\theta=\prns{\hat\theta_1, \hat\theta_2^{\op{aux}}} \in \Theta_1 \times \Theta_2 \subseteq \mathbb{R}_2$  corresponding to \cref{eq: IV-equation,eq: local-quantile-equation}. 
We denote $\theta =\prns{\theta_1, \theta_2^{\op{aux}}}$.
 We prove this by verifying all conditions in the assumptions in \cref{thm: general-split}.

\textbf{Verifying \cref{assump: jacob}.} Similar to the proof of \cref{thm: causal}, we can easily show that 
\begin{align*}
J^*  = \partial_{\theta}\{\hP\left[\psi(Z;\theta, \theta^{\op{aux}}_2,\eta^*_1(Z;\theta_1),\eta^*_2(Z))\right]\}\vert_{\theta = \tth}
\end{align*}
does not depend on $\tnua(Z; \theta_1)$ at all. Thus Assumption 1 holds trivially: 
\begin{align*}
J^*  = \partial_{\theta}\{\hP\left[\psi(Z; \theta, \tnua(Z; \theta_1), \tnub(Z))\right]\}\vert_{\theta = \tth} = \partial_{\theta}\{\hP\left[\psi(Z; \theta, \tnua(Z; \tth_1), \tnub(Z))\right]\}\vert_{\theta = \tth}.
\end{align*}

\textbf{Verifying \cref{assump: identification}. }
We first verify conditions \ref{assump: identification: identification} and \ref{assump: identification: conditioning} in \cref{assump: identification}. We can easily derive that
\begin{align*}
\hP\left[\psi(Z;\theta, \theta^{\op{aux}}_2,\eta^*_1(Z;\theta^*_1),\eta^*_2(Z))\right] = 
\begin{bmatrix}
\frac{\Prb{\mathcal{C}}F_1\prns{\theta_1 \mid \mathcal{C}}}{\theta_2^{\op{aux}}} - \gamma \\
\theta_2^{\op{aux}*} - \theta_2^{\op{aux}}
\end{bmatrix}
\end{align*}
and its Jacobian matrix is given by 
\begin{align*}
J(\theta) = \partial_{\theta}\{\hP\left[\psi(Z; \theta, \tnua(Z; \tth_1), \tnub(Z))\right]\} = 
\begin{bmatrix}
\frac{\Prb{\mathcal{C}}f_1\prns{\theta_1 \mid \mathcal{C}}}{\theta_2^{\op{aux}}} & -\frac{\Prb{\mathcal{C}}F_1\prns{\theta_1 \mid \mathcal{C}}}{\prns{\theta_2^{\op{aux}}}^2} \\
0 & - 1
\end{bmatrix}.
\end{align*}
This means that 
\begin{align*}
J(\theta^*) = 
\begin{bmatrix}
f_1\prns{\theta_1^* \mid \mathcal{C}} & -\frac{\gamma}{\theta^{\op{aux}*}_2} \\
0 & - 1
\end{bmatrix},
~~ 
\prns{J(\theta^*)}^{-1} = 
\begin{bmatrix}
\frac{1}{f_1\prns{\theta_1^* \mid \mathcal{C}}} & -\frac{\gamma}{\theta^{\op{aux}*}_2f_1\prns{\theta_1^* \mid \mathcal{C}}} \\
0 & - 1
\end{bmatrix}.
\end{align*}
Therefore,
\begin{align*}
\sigma_{\max}\prns{J(\theta^*)} \le 2\max\braces{f_1\prns{\theta_1^* \mid \mathcal{C}}, \frac{\gamma}{\theta^{\op{aux}*}_2}, 1} \le 2\max\braces{c_1', \frac{\gamma}{\epsilon}, 1} ,
\end{align*}
and 
\begin{align*}
\sigma_{\max}\prns{J^{-1}\prns{\tth}} \le 2\max\braces{\frac{1}{f_1\prns{\theta_1^* \mid \mathcal{C}}}, \frac{\gamma}{\theta^{\op{aux}*}_2f_1\prns{\theta_1^* \mid \mathcal{C}}}, 1} \le 2\max\braces{\frac{1}{c_3'}, \frac{\gamma}{c_3'\epsilon}, 1}.
\end{align*}
The latter implies that 
\begin{align*}
\sigma_{\min}\prns{J\prns{\tth}} = 1/\sigma_{\max}\prns{J^{-1}\prns{\tth}} \ge \frac{1}{2}\min\braces{c_3', c_3'\epsilon/\gamma, 1}.
\end{align*}
Therefore, condition \ref{assump: identification: conditioning} in \cref{assump: identification} is satisfied with $c_3 = \frac{1}{2}\min\braces{c_3', c_3'\epsilon/\gamma, 1}$, $c_4 =  2\max\braces{c_1', \frac{\gamma}{\epsilon}, 1} $.

Moreover, for any $\prns{\theta_1, \theta_2^{\op{aux}} } \in {\Theta_1 \times \Theta_2}$ and $t = 1$, $f_t(\theta_1 \mid \mathcal C) \le c_1'$, $\abs{\dot{f}_t(\theta_1 \mid \mathcal C)} \le c'_2$, so we have that entries in $J(\theta)$ are all Lipschtiz with $c_{\op{Lip}} \coloneqq \max\braces{\sqrt{\prns{\frac{c'_2}{\epsilon}}^2 + \prns{\frac{c_1'}{\epsilon^2}}^2}, \sqrt{\prns{\frac{2}{\epsilon^3}}^2 + \prns{\frac{c_1'}{\epsilon^2}}^2}}$ as a valid Lipschitz constant. Moreover, we have $2\|\hP\left[\psi(Z;\theta, \theta^{\op{aux}}_2,\eta^*_1(Z;\theta^*_1),\eta^*_2(Z))\right]\| \ge c_2$ for all $\theta = \prns{\theta_1, \theta_2^{\op{aux}}} \in\Theta$ such that $\|\theta - \theta^*\| \ge \frac{c_3}{2\sqrt{d}c_{\op{Lip}}}$.
By following the proof of \cref{thm: causal}, we can easily verify condition \ref{assump: identification: identification} in \cref{assump: identification}. 

Next, we verify condition 
\ref{assump: identification: metric entropy}
in \cref{assump: identification}. For any fixed $\eta_1\prns{Z;\theta_1'}$ and $\eta_2$, the class $\calF_{\eta, \theta_1'} = \{\psi_j(Z; \theta, \nua(Z; \theta_1'), \nub(Z)): j = 1, \dots, d, \theta \in \Theta\}$ depend on $\theta$ only through $\braces{\indic{Y \le \theta_1}: \theta_1\in\Theta_1}$ and $\braces{\theta_2^{\op{aux}}: \theta_2^{\op{aux}}\in\Theta_2}$. Since the latter two classes are Donsker class, \ref{assump: identification: metric entropy}
in \cref{assump: identification} for the function class $\calF_{\eta, \theta_1'} = \{\psi_j(Z; \theta, \nua(Z; \theta_1'), \nub(Z)): j = 1, \dots, d, \theta \in \Theta\}$ has to be satisfied as well. 

\textbf{Verifying \cref{assump: error}.} We take $\mathcal{T}_N$ to be the set that contains all $(\eta_1\prns{\cdot; \theta_1'} = \tilde\mu(\cdot, \theta_1'),\eta_2\prns{\cdot} =\prns{\nu_w\prns{\cdot}, \tilde\pi(\cdot))}$ that satisfies the following conditions:  for $w = 0, 1$, 
\begin{align*}
    &\left\|\bigg\{\hP \left[{\tilde \mu}_w\left(X; \hthinit\pk)\right) - \tilde{\mu}^*_w\left(X; \hthinit\pk)\right)  \right]^2\bigg\}^{1/2}\right\|   \le \tilde \rho_{\mu, N}, ~ \bigg\{\hP \left[{\nu}_w(X) - {\nu}^*_w(X)\right]^2\bigg\}^{1/2} \le \tilde\rho_{\nu, N}, \\
  &\qquad\qquad\qquad\qquad \bigg\{\hP \left[{\tilde \pi}^{(k)}(X) - \tilde{\pi}^*(X)\right]^2\bigg\}^{1/2} \le \tilde\rho_{{\pi}, N}, ~~ |\theta_1' - \theta^*_1| \le \tilde \rho_{\theta, N},
\end{align*}
and $\epsilon \le \hat{\tilde\pi}\pk( X) \le 1- \epsilon,  0 \le \hat{\tilde \mu}_w\pk\left(X; \hthinit\pk)\right) \le 1$, $0 \le \hat{\nu}^{(k)}_w(X) \le 1$ almost surely. Moreover, 
$\tilde\rho_{\pi, N} \le \frac{\delta_N^3}{\log N}$, $\tilde\rho_{\mu, N} + C\tilde\rho_{\theta, N} \le \frac{\delta^2_N}{\log N}$, 
$\tilde\rho_{\pi, N}\prns{\tilde\rho_{\mu, N} + C\tilde\rho_{\theta, N}} \le \frac{\epsilon^4\prns{1-\epsilon}^3}{4\prns{\epsilon^3 + \prns{1-\epsilon}^3}}\delta_N N^{-1/2}$, 
$\tilde\rho_{\pi, N}{\tilde\rho_{\nu, N}} \le  \frac{\epsilon^3\prns{1-\epsilon}^3}{8\prns{\epsilon^3 + \prns{1-\epsilon}^3}}\delta_N N^{-1/2}$
 with $\delta_N$ satisfying that $\delta_N \le \frac{\epsilon^3\prns{1-\epsilon}^2}{4C + 3\epsilon^2\prns{1-\epsilon}}$,
$\frac{\delta_N}{\log N} \le \frac{1}{C_{\epsilon}}$ for a positive constant $C_{\epsilon}$ given in \cref{eq: Cepsilon}.

Then \cref{assump: nuisance-rate-iv} and \cref{thm: iv} condition \ref{thm: iv: rate} ensure that  the nuisance estimates $(\hq(, \hthinit), \hpib) \in \mathcal{T}_N$ with probability, namely, condition \ref{assump: error: nuisance-set} in \cref{assump: error} is satisfied. 

Before verifying other conditions, we first note that 
\begin{align*}
\tilde\mu_w^*\prns{X;\theta_1} 
  &= \Prb{T=1, Y\le\theta_1\mid X, W= w}  \\
  &= \Prb{T(w)=1, Y(1)\le\theta_1\mid X} = F_{1, w}\prns{\theta_1 \mid X}v_w\prns{X}.
\end{align*}
It follows from \cref{thm: iv: 2} that for any $\theta_1 \in \mathcal{B}(\tth_1;\max\{\frac{4\tilde\rho_{\pi, N}}{\epsilon^2\prns{1-\epsilon}\delta_N}, \rho_{\theta, N}\}) \cap \Theta$, 
\begin{align*}
\bracks{\hP\bracks{\prns{\tilde\mu_w^*\prns{X;\theta_1}  - \tilde\mu_w^*\prns{X;\theta_1^*} }^2}}^{1/2} \le C\|\theta_1 - \theta^*_1\|. 
\end{align*}
This means that 
for any $(\mu(\cdot, \theta_1'), \pi(\cdot)) \in \mathcal T_N$,
\begin{align*}
    &\qquad\qquad\qquad\qquad\qquad \left\|\bigg\{\hP \left[\mu(X, T; \theta_1') - \tq(X, T; \tth_1)\right]^2\bigg\}^{1/2}\right\| \\
    &\le \left\|\bigg\{\hP \left[\mu(X, T; \theta_1') - \tq(X, T; \theta_1')\right]^2\bigg\}^{1/2}\right\| + \left\|\bigg\{\hP \left[ \tq(X, T; \theta_1') - \tq(X, T; \tth_1)\right]^2\bigg\}^{1/2}\right\| \\
    &= \tilde\rho_{\mu, N} + C\tilde\rho_{\theta, N}.
\end{align*}

Next, we verify  \cref{assump: error} condition \ref{assump: error: rate-condition}. 

We first verify the condition on $r_N$. By following the proof of \cref{thm: causal}, we can show that for any $(\eta_1(\cdot;\theta'_1), \eta_2(\cdot)) \in \mathcal{T}_N$,
\begin{align*}
&\|\hP\left[\psi(Z;\theta, \theta^{\op{aux}}_2,\eta_1(Z;\theta'_1),\eta_2(Z))\right] - \hP\left[\psi(Z;\theta, \theta^{\op{aux}}_2,\eta^*_1(Z;\theta^*_1),\eta^*_2(Z))\right]\| \\
\le & 
\left\|\frac{1}{\theta_2^{\op{aux}}}\hP\prns{\frac{W - \tilde\pi\prns{X}}{\tilde\pi\prns{X}\prns{1-\tilde\pi\prns{X}}} - \frac{W-\tilde\pi^*\prns{X}}{\tilde\pi^*\prns{X}\prns{1-\tilde\pi^*\prns{X}}}}
\prns{\tilde\mu^*_W\prns{X; \theta_1} - \tilde\mu^*_W\prns{X; \theta_1^*}}
\right\| 
\\
+& 
\left\|\frac{1}{\theta_2^{\op{aux}}}\hP\prns{\frac{W - \tilde\pi\prns{X}}{\tilde\pi\prns{X}\prns{1-\tilde\pi\prns{X}}} - \frac{W-\tilde\pi^*\prns{X}}{\tilde\pi^*\prns{X}\prns{1-\tilde\pi^*\prns{X}}}}
\prns{\tilde\mu^*_W\prns{X; \theta_1^*} - \tilde\mu_W\prns{X; \theta_1'}}
\right\| \\
\le & \frac{1}{\epsilon^2\prns{1-\epsilon}}\bracks{\hP\prns{\tilde\pi\prns{X} - \tilde\pi^*\prns{X}}^2}^{1/2}\braces{\bracks{\hP\prns{{\tilde\mu^*_1\prns{X; \theta_1} - \tilde\mu^*_1\prns{X; \theta_1^*}}}^2}^{1/2} + \bracks{\hP\prns{{\tilde\mu^*_0\prns{X; \theta_1} - \tilde\mu^*_0\prns{X; \theta_1^*}}}^2}^{1/2}} \\
+ & \frac{1}{\epsilon^2\prns{1-\epsilon}}\bracks{\hP\prns{\tilde\pi\prns{X} - \tilde\pi^*\prns{X}}^2}^{1/2}\braces{\bracks{\hP\prns{{\tilde\mu^*_1\prns{X; \theta_1^*} - \tilde\mu_1\prns{X; \theta_1'}}}^2}^{1/2} + \bracks{\hP\prns{{\tilde\mu^*_0\prns{X; \theta_1^*} - \tilde\mu_0\prns{X; \theta_1'}}}^2}^{1/2}} \\
\le & \frac{4}{\epsilon^2\prns{1-\epsilon}}\tilde\rho_{\pi, N}.
\end{align*}
The last inequality holds because 
\begin{align*}
\tilde\mu^*_w\prns{X; \theta_1} = \Prb{T\prns{w} = 1, Y\prns{1}\le\theta_1 \mid X} \in [0, 1], \text{ almost surely, }
\end{align*}
and so is $\tilde\mu_w\prns{X; \theta_1}$. 
This means that the condition on $r_N$ is satisfied with $\tau_N = \frac{4\tilde\rho_{\pi, N}}{\epsilon^2\prns{1-\epsilon}\delta_N}$.

Next, we verify the condition on $r_N'$. Again, by following the proof of \cref{thm: causal}, we have that for any $\|\theta-\theta^*\| \le \frac{4\tilde\rho_{\pi, N}}{\epsilon^2\prns{1-\epsilon}\delta_N}$ and any $(\eta_1(\cdot;\theta'_1), \eta_2(\cdot)) \in \mathcal{T}_N$, 
\begin{align*}
&\left\|\braces{\hP\left[\psi(Z;\theta, \theta^{\op{aux}}_2,\eta_1(Z;\theta'_1),\eta_2(Z)) - \psi(Z;\theta, \theta^{\op{aux}}_2,\eta^*_1(Z;\theta^*_1),\eta^*_2(Z))\right]^2}^{1/2}\right\| \\
\le&  \left\|\braces{\hP\prns{\frac{W - \tilde\pi\prns{X}}{\tilde\pi\prns{X}\prns{1-\tilde\pi\prns{X}}} - \frac{W-\tilde\pi^*\prns{X}}{\tilde\pi^*\prns{X}\prns{1-\tilde\pi^*\prns{X}}}}^2
\prns{\tilde\mu^*_W\prns{X; \theta_1} - \tilde\mu^*_W\prns{X; \theta_1^*}}^2}^{1/2}
\right\| 
\\
+& 
\left\|\braces{\hP\prns{\frac{W - \tilde\pi\prns{X}}{\tilde\pi\prns{X}\prns{1-\tilde\pi\prns{X}}} - \frac{W-\tilde\pi^*\prns{X}}{\tilde\pi^*\prns{X}\prns{1-\tilde\pi^*\prns{X}}}}^2
\prns{\tilde\mu^*_W\prns{X; \theta_1^*} - \tilde\mu_W\prns{X; \theta_1'}}^2}^{1/2}
\right\| \\
+& \left\|\braces{\hP\prns{\frac{W - \tilde\pi^*\prns{X}}{\tilde\pi^*\prns{X}\prns{1-\tilde\pi^*\prns{X}}}}^2
\prns{\tilde\mu^*_W\prns{X; \theta_1^*} - \tilde\mu_W\prns{X; \theta_1'}}^2}^{1/2}
\right\| \\
\le & \frac{4C}{\epsilon^3\prns{1-\epsilon}^2\delta_N}\tilde\rho_{\pi, N} + \frac{1}{\epsilon\prns{1-\epsilon}}\prns{\tilde\rho_{\mu, N} + C\tilde\rho_{\theta,N}} \\
&+ \frac{1}{\epsilon\prns{1-\epsilon}}\braces{\hP\bracks{\prns{\tilde\mu^*_1\prns{X; \theta_1^*} - \tilde\mu_1\prns{X; \theta_1'}}^2}}^{1/2} + \frac{1}{\epsilon\prns{1-\epsilon}}\braces{\hP\bracks{\prns{\tilde\mu^*_0\prns{X; \theta_1^*} - \tilde\mu_0\prns{X; \theta_1'}}^2}}^{1/2} \\
\le& \frac{4C}{\epsilon^3\prns{1-\epsilon}^2\delta_N}\tilde\rho_{\pi, N} + \frac{3}{\epsilon\prns{1-\epsilon}}\prns{\tilde\rho_{\mu, N} + C\tilde\rho_{\theta, N}}.
\end{align*}
Therefore, if $\tilde\rho_{\pi, N} \le \frac{\delta_N^3}{\log N}$ and $\tilde\rho_{\mu, N} + C\tilde\rho_{\theta, N} \le \frac{\delta_N^2}{\log N}$, then $r_N' = \frac{\delta_N^2}{\log N}\prns{\frac{4C}{\epsilon^3\prns{1-\epsilon}^2} + \frac{3}{\epsilon\prns{1-\epsilon}}} \le \frac{\delta_N}{\log N}$ given $\delta_N \le \frac{\epsilon^3\prns{1-\epsilon}^2}{4C + 3\epsilon^2\prns{1-\epsilon}}$.

Finally, we verify the condition on $\lambda_N'$. Note that in this case $V\prns{\theta_2} = 0$ and denote 
\begin{align*}
\tilde\psi_1\prns{Z;\theta, \eta_1\prns{Z;\theta_1}, \eta_2\prns{Z}} \coloneqq \theta_2^{\op{aux}}{\psi_1(Z;{\theta},{\eta_1(Z;\theta_1')},{\eta_2(Z)})}.
\end{align*}
Then for any $(\eta_1(\cdot;\theta'_1), \eta_2(\cdot)) \in \mathcal{T}_N$ and $\theta \in \mathcal{B}\prns{\theta^*;  \frac{4\tilde\rho_{\pi, N}}{\epsilon^2\prns{1-\epsilon}\delta_N}}$, we have 
\begin{align*}
&\abs{\partial_r^2 \hP\bracks{\psi_1(Z;\theta^* + r\prns{\theta - \theta^*},\eta_1^*(Z;\theta_1) + r\prns{\eta_1(Z;\theta_1') - \eta_1^*(Z;\theta_1)},\eta_2^*(Z) + r\prns{\eta_2(Z) - \eta_2^*(Z)})}} \\
=& \abs{\partial_r^2\hP\bracks{\tilde\psi_1(Z;\theta^* + r\prns{\theta - \theta^*},\eta_1^*(Z;\theta_1) + r\prns{\eta_1(Z;\theta_1') - \eta_1^*(Z;\theta_1)},\eta_2^*(Z) + r\prns{\eta_2(Z) - \eta_2^*(Z)})}} \\
& \qquad\qquad\qquad\qquad\qquad\qquad\qquad\qquad\qquad\qquad\qquad\qquad\qquad \times\frac{1}{\theta_2^{\op{aux}*}+r\prns{\theta_2^{\op{aux}} - \theta_2^{\op{aux}*}}} \\
+& 2\abs{\partial_r\hP\bracks{\tilde\psi_1(Z;\theta^* + r\prns{\theta - \theta^*},\eta_1^*(Z;\theta_1) + r\prns{\eta_1(Z;\theta_1') - \eta_1^*(Z;\theta_1)},\eta_2^*(Z) + r\prns{\eta_2(Z) - \eta_2^*(Z)})}} \\
& \qquad\qquad\qquad\qquad\qquad\qquad\qquad\qquad\qquad\qquad\qquad\qquad\qquad \times\frac{\theta_2^{\op{aux}} - \theta_2^{\op{aux}*}}{\prns{\theta_2^{\op{aux}*}+r\prns{\theta_2^{\op{aux}} - \theta_2^{\op{aux}*}}}^2} \\
+& \abs{\hP\bracks{\tilde\psi_1(Z;\theta^* + r\prns{\theta - \theta^*},\eta_1^*(Z;\theta_1) + r\prns{\eta_1(Z;\theta_1') - \eta_1^*(Z;\theta_1)},\eta_2^*(Z) + r\prns{\eta_2(Z) - \eta_2^*(Z)})}} \\
& \qquad\qquad\qquad\qquad\qquad\qquad\qquad\qquad\qquad\qquad\qquad\qquad\qquad \times\frac{2\prns{\theta_2^{\op{aux}} - \theta_2^{\op{aux}*}}^2}{\prns{\theta_2^{\op{aux}*}+r\prns{\theta_2^{\op{aux}} - \theta_2^{\op{aux}*}}}^3}.
\end{align*}
By following the proof of \cref{thm: causal}, we can prove that 
\begin{align*}
 &\abs{\partial_r^2\hP\bracks{\tilde\psi_1(Z;\theta^* + r\prns{\theta - \theta^*},\eta_1^*(Z;\theta_1) + r\prns{\eta_1(Z;\theta_1') - \eta_1^*(Z;\theta_1)},\eta_2^*(Z) + r\prns{\eta_2(Z) - \eta_2^*(Z)})}} \\
 \le & 4\prns{\frac{1}{\epsilon^3}+ \frac{1}{\prns{1- \epsilon}^3}}\tilde\rho_{\pi, N}\prns{\tilde\rho_{\mu, N} + C\tilde\rho_{\theta, N}} + 4C\prns{\frac{1}{\epsilon^3\prns{1-\epsilon}} + \frac{1}{\epsilon^2\prns{1-\epsilon}^2}}\tilde\rho_{\pi, N}\|\theta-\tth\|,
\end{align*}
and 
\begin{align*}
&2\abs{\partial_r\hP\bracks{\tilde\psi_1(Z;\theta^* + r\prns{\theta - \theta^*},\eta_1^*(Z;\theta_1) + r\prns{\eta_1(Z;\theta_1') - \eta_1^*(Z;\theta_1)},\eta_2^*(Z) + r\prns{\eta_2(Z) - \eta_2^*(Z)})}} \\
\le& 2\prns{1+\frac{1}{\epsilon}+\frac{1}{1-\epsilon}}\prns{\tilde\rho_{\mu, N} + C\tilde\rho_{\theta, N}} + \prns{\frac{2}{\epsilon^2} + \frac{2}{\prns{1-\epsilon}^2} + \frac{4C}{\epsilon^3\prns{1-\epsilon}\delta_N}+\frac{4C}{\epsilon^2\prns{1-\epsilon}^2\delta_N}}\tilde\rho_{\pi, N},
\end{align*}
and 
\begin{align*}
&\abs{\hP\bracks{\tilde\psi_1(Z;\theta^* + r\prns{\theta - \theta^*},\eta_1^*(Z;\theta_1) + r\prns{\eta_1(Z;\theta_1') - \eta_1^*(Z;\theta_1)},\eta_2^*(Z) + r\prns{\eta_2(Z) - \eta_2^*(Z)})}} \\
&\times \abs{\theta_2^{\op{aux}} - \theta_2^{\op{aux}*}} \le  8\prns{1+\frac{1}{\epsilon} + \frac{1}{1-\epsilon}}\frac{\tilde\rho_{\pi, N}}{\epsilon^2\prns{1-\epsilon}\delta_N}. 
\end{align*}
Therefore, 
\begin{align*}
&\abs{\partial_r^2 \hP\bracks{\psi_1(Z;\theta^* + r\prns{\theta - \theta^*},\eta_1^*(Z;\theta_1) + r\prns{\eta_1(Z;\theta_1') - \eta_1^*(Z;\theta_1)},\eta_2^*(Z) + r\prns{\eta_2(Z) - \eta_2^*(Z)})}} \\
\le & 4\prns{\frac{1}{\epsilon^4}+ \frac{1}{\prns{1- \epsilon}^3\epsilon}}\tilde\rho_{\pi, N}\prns{\tilde\rho_{\mu, N} + C\tilde\rho_{\theta, N}} + \prns{C_{\epsilon, 1}\frac{\tilde\rho_{\pi, N}}{\delta_N} + C_{\epsilon, 2}\prns{\tilde\rho_{\mu, N} + C\tilde\rho_{\theta, N}}}\|\theta - \tth\|,
\end{align*}
where 
\begin{align*}
C_{\epsilon, 1} 
  &= 4C\prns{\frac{1}{\epsilon^3\prns{1-\epsilon}} + \frac{1}{\epsilon^2\prns{1-\epsilon}^2}} + \frac{1}{\epsilon^2} \prns{\frac{2}{\epsilon^2} + \frac{2}{\prns{1-\epsilon}^2} + \frac{4C}{\epsilon^3\prns{1-\epsilon}}+\frac{4C}{\epsilon^2\prns{1-\epsilon}^2}}\nonumber \\
  &+  16\prns{1+\frac{1}{\epsilon} + \frac{1}{1-\epsilon}}\frac{1}{\epsilon^5\prns{1-\epsilon}},
\end{align*}
and 
\begin{align*}
C_{\epsilon, 2}  = \frac{2}{\epsilon^2}\prns{1+\frac{1}{\epsilon}+\frac{1}{1-\epsilon}}.
\end{align*}
Also define 
\begin{align}
C_{\epsilon} = C_{\epsilon, 1} + C_{\epsilon, 2}.\label{eq: Cepsilon}
\end{align}

Since $\tilde\rho_{\pi, N} \le \frac{\delta_N^3}{\log N}$ and $\tilde\rho_{\mu, N} + C\tilde\rho_{\theta, N} \le \frac{\delta_N^2}{\log N}$, if $\frac{\delta_N}{\log N} \le \frac{1}{C_{\epsilon}}$, then 
\begin{align*}
\prns{C_{\epsilon, 1}\frac{\tilde\rho_{\pi, N}}{\delta_N} + C_{\epsilon, 2}\prns{\tilde\rho_{\mu, N} + C\tilde\rho_{\theta, N}}} \le {C_{\epsilon}}\frac{\delta^2_N}{\log N}  \le \delta_N. 
\end{align*}
Morevoer, when $\tilde\rho_{\pi, N}\prns{\tilde\rho_{\mu, N} + C\tilde\rho_{\theta, N}} \le \frac{\epsilon^4\prns{1-\epsilon}^3}{8\prns{\epsilon^3 + \prns{1-\epsilon}^3}}\delta_N N^{-1/2}$, we have 
\begin{align*}
4\prns{\frac{1}{\epsilon^4}+ \frac{1}{\prns{1- \epsilon}^3\epsilon}}\tilde\rho_{\pi, N}\prns{\tilde\rho_{\mu, N} + C\tilde\rho_{\theta, N}} \le \frac{1}{2}\delta_N N^{-1/2}.
\end{align*}
Moreover, we can similarly show that 
\begin{align*}
\abs{\partial_{r}^2\hP\bracks{\psi_2(Z;\theta^{\op{aux}*}_2+r\prns{\theta^{\op{aux}}_2 - \theta^{\op{aux}*}_2},\eta^*_2(Z) + r\prns{\eta_2(Z)-\eta^*_2(Z)})}} 
  &\le 4\prns{\frac{1}{\epsilon^3} +  \frac{1}{\prns{1 - \epsilon}^3}}\tilde\rho_{\pi, N}{\tilde\rho_{\nu, N}} \\
  &\le \frac{1}{2}\delta_N N^{-1/2},
\end{align*}
provided that  $\tilde\rho_{\pi, N}{\tilde\rho_{\nu, N}} \le  \frac{\epsilon^3\prns{1-\epsilon}^3}{8\prns{\epsilon^3 + \prns{1-\epsilon}^3}}\delta_N N^{-1/2}$. 

If follows that 
\begin{align*}
&\left\|{\partial_r^2 \hP\bracks{\psi(Z;\theta^* + r\prns{\theta - \theta^*},\eta_1^*(Z;\theta_1) + r\prns{\eta_1(Z;\theta_1') - \eta_1^*(Z;\theta_1)},\eta_2^*(Z) + r\prns{\eta_2(Z) - \eta_2^*(Z)})}}\right\| \\
\le & \delta_N\|\theta-\tth\| + \delta_N N^{-1/2},
\end{align*}
which verifies \cref{assump: error} condition \ref{assump: error: rate-condition}. 

Therefore,  we have 
\begin{align*}
\sqrt{N}\begin{bmatrix}
\hat\theta_1  - \tth_1\\
\hat\theta_2^{\op{aux}} -\theta^{\op{aux}*}_2
\end{bmatrix}
= \frac{1}{\sqrt{N}}\sum_{i=1}^N J^{-1}\prns{\theta^*} \begin{bmatrix}
 \psi_1(Z_i;\theta^*,\eta^*_1(Z_i;\theta^*_1),\eta^*_2(Z_i)) \\
  \psi_2(Z_i; \theta^{\op{aux}*}_2,\eta^*_2(Z_i)) 
 \end{bmatrix}
 + O_{\hP}\prns{\rho_N}.
\end{align*}
This means that 
\begin{align*}
\sqrt{N}\prns{\hat\theta_1  - \tth_1} 
    &= \frac{1}{\sqrt{N}}\sum_{i=1}^N
        \bigg[\frac{1}{f_1\prns{\theta_1^* \mid \mathcal{C}}}\psi_1(Z_i;\theta^*,\eta^*_1(Z_i;\theta^*_1),\eta^*_2(Z_i)) \\
    &\qquad\qquad\qquad\qquad -\frac{\gamma}{\theta^{\op{aux}*}_2f_1\prns{\theta_1^* \mid \mathcal{C}}}  \psi_2(Z_i; \theta^{\op{aux}*}_2,\eta^*_2(Z_i))\bigg]+ O_{\hP}\prns{\rho_N}.
\end{align*}
\end{proof}

\begin{proof}[Proof for \texorpdfstring{\cref{thm: expectile}}{a}.]
We only need to verify the conditions in \cref{thm: causal}.

\textbf{Verifying condition \ref{assump: causal: regularity} in \cref{thm: causal}.}
We only need to verify  that condition 
  \ref{assump: identification: metric entropy} of \cref{assump: identification} hold.  
Since $\Theta$ is compact, $\{y \mapsto \prns{1-\gamma}{y - \theta_1}, \theta \in \Theta\}$, $\{y \mapsto \prns{1-2\gamma}\max\{y - \theta_1, 0\}, \theta \in \Theta\}$  are obviously Donsker classes, condition \ref{assump: identification: metric entropy}  of \cref{assump: identification} is satisfied. 

\textbf{Verifying condition \ref{assump: causal: Jacobian} and \ref{assump: causal: identification}  in \cref{thm: causal}.} According to \cref{eq: expectile-complete}, the estimating function for complete data is given by 
\[
  U(Y(1); \theta_1) = \left(1 - \gamma\right)\left(Y(1)  - \theta_1\right) - (1 - 2\gamma)\max\left(Y(1) - \theta_1 , 0\right).
\]
It follows that 
\begin{align*}
\frac{\partial}{\partial \theta_1}\hP\left[U(Y(t); \theta_1)\right] 
  &= -(1 - \gamma) - (1 - 2\gamma)\frac{\partial}{\partial \theta_1}\hP\left[\max\left(Y(t) - \theta_1 , 0\right)\right] \\ 
  &=  -(1 - \gamma) - (1 - 2\gamma)\frac{\partial}{\partial \theta_1}\int_{\theta_1}^{\infty}(y - \theta_1)f_t(y)dy \\
  &= -(1 - \gamma) + (1 - 2\gamma)\int_{\theta_1}^{\infty} f_t(y)dy \\
  &= -\gamma - (1 - 2\gamma)F_t(\theta_1).
\end{align*}
Here the differentiability of $\frac{\partial}{\partial \theta_1}\hP\left[U(Y(t); \theta_1)\right]$ is guaranteed by Leibniz integral rule, the continuity of its derivative at $\theta_1^*$ is guaranteed by the continuity of  $F_t(\theta_1)$ at $\theta_1^*$ , and $J(\theta_1^*) = \frac{\partial}{\partial \theta_1}\hP\left[U(Y(t); \theta_1)\right]\vert_{\theta_1 = \theta_1^*} = -\gamma - (1 - 2\gamma)F_t(\theta_1^*)$, whose singular value  $|-\gamma - (1 - 2\gamma)F_t(\theta_1^*)|$ is bounded between $c_4'$ and $\max\{\gamma, 1 - \gamma\}$. Moreover, $\frac{\partial}{\partial \theta_1}\hP\left[U(Y(t); \theta_1)\right] \le \max\{\gamma, 1 - \gamma\}$, which implies that  $\hP\left[U(Y(t); \theta_1)\right] $ is Lipschtiz continuous with Lipschitz constant $\max\{\gamma, 1 - \gamma\} \le 1$. Therefore, the constants $c'$ in condition \ref{assump: causal: Jacobian} and constant $c_3$ in \ref{assump: causal: identification} of \cref{thm: causal} can be set as $c_3 = c_1', c' = 1$. 
The assumption that $\|\theta - \theta^*\| \ge \frac{c_3}{2c'} = \frac{c_1'}{2}$,  $2\hP\left[U(Y(t); \theta_1)\right] \ge c_2'$ for any $\theta \in \Theta$ ensures the condition \ref{assump: causal: Jacobian} of \cref{thm: causal} with constant $c_2 = c_2'$. 

\textbf{Verifying condition \ref{assump: causal: exchange-int-diff} in \cref{thm: causal}.} Note that for any $\theta \in \mathcal{B}(\tth;\frac{4C\sqrt{d}\rho_{\pi, N}}{\delta_N\varepsilon_\pi}) \cap \Theta$, 
\begin{align*}
\mu^*(X, 1; \theta_1^* + r(\theta - \theta_1^*)) 
  &= \expect[U(Y; \theta_1^* + r(\theta - \theta_1^*)) \mid X, T = 1] \\
  &= (1- \gamma)\eta_{2, 1}^*(Z) - (1 - 2\gamma)\eta_{1}^*(Z; \theta_1^* + r(\theta_1 - \theta_1^*)).
\end{align*}
Thus 
\begin{align*}
\left|\partial_r\mu^*(X, 1; \theta_1^* + r(\theta_1 - \theta_1^*))\right| 
  &= \left|- \gamma(\theta_1 - \theta_1^*) - (1 - 2\gamma)(\theta_1 - \theta_1^*)F_t(\theta_1^* + r(\theta - \theta_1^*) \mid X)\right| \\
  &\le 2|\theta_1 - \theta_1^*|, \\
\left|\partial^2_r\mu^*(X, 1; \theta_1^* + r(\theta_1 - \theta_1^*))\right| &= |1 - 2\gamma||\theta_1 - \theta_1^*|f_t(\theta_1^* + r(\theta_1 - \theta_1^*) \mid X) \le C|1 - 2\gamma||\theta_1 - \theta_1^*|,
\end{align*}
which trivially imply condition \ref{assump: causal: exchange-int-diff} in \cref{thm: causal}.

\textbf{Verifying condition iv in  \cref{thm: causal}.} Again
\begin{align*}
\mu^*(X, 1; \theta_1) 
  &= (1- \gamma)\eta_{2, 1}^*(Z) - (1 - 2\gamma)\eta_{1}^*(Z; \theta_1).
\end{align*}
The the asserted assumpton iv means that $\left\{\hP[\eta_{2, 1}^*(Z)]^2\right\}^{1/2}\le C$ and $\left\{\hP[\eta_{1}^*(Z; \theta_1)]^2\right\}^{1/2} \le C$ for any $\theta \in \Theta$, thus $\left\{\hP\left[\mu^*(X, 1; \theta_1)\right]^2\right\}^{1/2}$ is upper bounded by $\left|1 - \gamma\right| + \left|1 - 2\gamma\right| C \le 2C$ for any $\theta \in \Theta$. 

Plus, for any $\theta_1 \in \mathcal{B}(\tth_1;\max\{\frac{4C\sqrt{d}\rho_{\pi, N}}{\delta_N\varepsilon_\pi}, \rho_{\pi, N}\}) \cap \Theta$
\begin{align*}
\left\{\hP\left[\frac{\partial}{\partial \theta_1}\mu^*(X, 1; \theta_1)\right]^2\right\}^{1/2}  
  &\le \sup_{x}|-\gamma - (1 - 2\gamma)F_t(\theta_1\mid X = x)| \le 2, \\
\hP\left[\frac{\partial^2}{\partial \theta_1^2}\mu^*(X, 1; \theta_1)\right] 
  &\le |1 - 2\gamma| \hP\left[f_t(\theta_1\mid X)\right] \le  C|1 - 2\gamma|,
\end{align*}
and there exists $\tilde{\theta}_1$ between $\theta_1$ and $\theta_1^*$ such that 
\begin{align*}
\left\{\hP \left[\tq(X,1; \theta_1) - \tq(X,1; \tth_1)\right]^2\right\}^{1/2} = |\theta_1 - \theta_1^*|\left\{\hP \left[\frac{\partial}{\partial \theta_1}\mu^*(X, 1; \tilde{\theta}_1)\right]^2\right\}^{1/2} \le 2|\theta_1 - \theta_1^*|.
\end{align*}
\end{proof}

\end{document}